\PassOptionsToPackage{table}{xcolor}
\documentclass[11pt,a4paper]{xera_paperstyle}

\usepackage{multirow}
\usepackage[bottom]{footmisc}
\usepackage{placeins}
\usepackage{booktabs}
\usepackage{graphicx}
\usepackage{colortbl}
\usepackage{subcaption}
\usepackage{tabularx}
\usepackage{pifont}
\usepackage{array}
\usepackage{algorithm}
\usepackage{algpseudocode}
\usepackage[capitalize]{cleveref}
\crefname{section}{Section}{Sections}
\Crefname{section}{Section}{Sections}
\crefname{subsection}{Section}{Sections}
\Crefname{subsection}{Section}{Sections}
\crefname{subsubsection}{Section}{Sections}
\Crefname{subsubsection}{Section}{Sections}
\crefname{appendix}{Appendix}{Appendices}
\Crefname{appendix}{Appendix}{Appendices}
\crefname{subappendix}{Appendix}{Appendices}
\Crefname{subappendix}{Appendix}{Appendices}
\crefname{subsubappendix}{Appendix}{Appendices}
\Crefname{subsubappendix}{Appendix}{Appendices}
\crefname{figure}{Figure}{Figures}
\Crefname{figure}{Figure}{Figures}
\crefname{table}{Table}{Tables}
\Crefname{table}{Table}{Tables}
\crefname{equation}{Equation}{Equations}
\Crefname{equation}{Equation}{Equations}
\crefname{algorithm}{Algorithm}{Algorithms}
\Crefname{algorithm}{Algorithm}{Algorithms}

\definecolor{oursbg}{HTML}{E8F5F2}
\definecolor{eOurs}{HTML}{E8F5F2}
\definecolor{eLibero}{HTML}{E7F2EE}
\definecolor{ePlus}{HTML}{E3EFF0}
\definecolor{ePro}{HTML}{E5EDF6}
\definecolor{eGroup}{HTML}{EEEEEE}

\providecommand{\ptYes}{\ding{51}}
\providecommand{\ptNo}{\ding{55}}
\providecommand{\ourscell}[1]{\cellcolor{oursbg}#1}

\renewcommand{\reportlogofile}{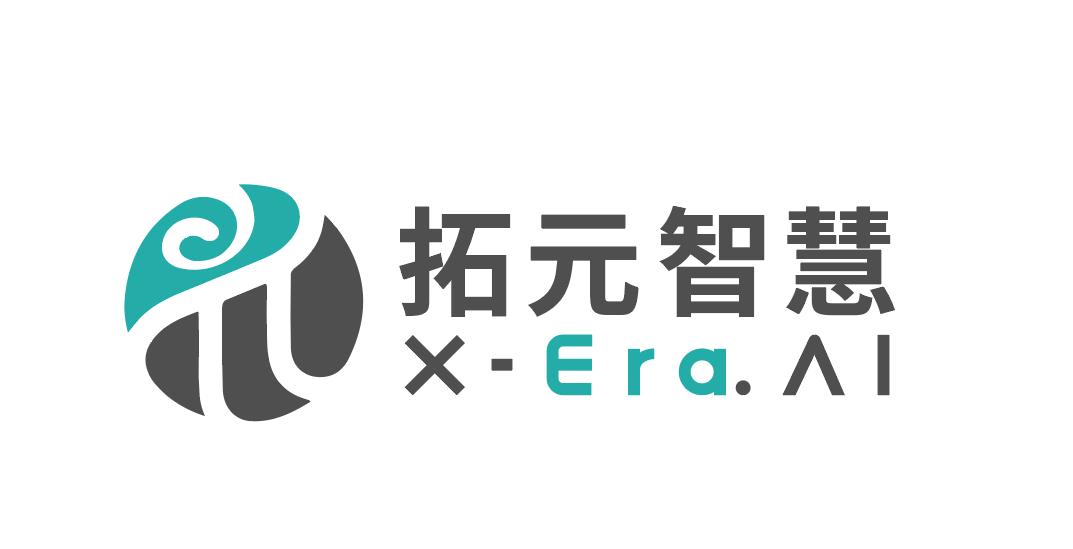}
\renewcommand{\reportlogoheight}{1.7cm}

\renewcommand{\reportresources}{%
  \reportlinkitem{GitHub}{https://github.com/X-EraAI/e05}%
  \reportlinkitem{Project}{https://www.x-era.com/e05}%
}

\renewcommand{\today}{}

\title{%
  \vspace*{-0.6cm}
  \teal{E0.5: }
  \graytext{Scalable World-Action Modeling on 3D Foundation Models through }
  \teal{Future Geometry Prediction and Backbone Mixtures}
  \vspace*{-0.8cm}
}

\author{\vspace{-0.6cm}
X-Era AI Lab Team \\ \vspace{-0.3cm}
\small X-Era.AI\\
}

\begin{abstract}
World-action models (WAMs) offer a promising framework for embodied intelligence, yet mainstream approaches remain centered on video foundation models or vision-language models. Even when geometric features extracted by 3D models such as VGGT are incorporated, they are typically passed into a video model or VLM that serves as the primary backbone, leaving 3D geometry as an auxiliary input or prediction target rather than the native foundation for world modeling. We present \textbf{E0.5}, a scalable world-action model built directly on 3D foundation models. E0.5 introduces three key innovations. First, it adopts a 3D foundation model as the primary backbone, making 3D geometry the native representation for both world modeling and action prediction. Second, it explicitly predicts future 3D states together with actions, enabling the model to capture geometric world dynamics beyond action-only modeling or future video prediction. Third, it introduces a Mixture-of-Geometry architecture that scales model capacity by combining multiple 3D foundation backbones. Experiments on widely used simulation benchmarks and real-robot tasks demonstrate the effectiveness of the proposed design. On XPolicyLab RoboTwin, E0.5 achieves a state-of-the-art average success rate of 84.60\% across the clean $\rightarrow$ clean and clean $\rightarrow$ random evaluation settings without embodied-data pretraining. Under the clean $\rightarrow$ random setting, E0.5 achieves 79.90\% average success, compared with 1.92\% for the video-based FastWAM. Ablations under this same setting show that future 3D prediction improves the success rate by 12.81 points over current-state reconstruction, while expanding from a single geometric backbone to Mixture-of-Geometry further improves average success by 2.73 points. E0.5 further achieves 98.90\% success on LIBERO, while maintaining generalization on LIBERO-Plus (88.78\%) and LIBERO-PRO (55.36\%). On real robots, without embodied-data pretraining, E0.5 achieves mean success rates of 73.3\% on Zerith H1 and 66.7\% on AgileX Piper under hard settings, outperforming embodied-pretrained $\pi_{0.5}$; we additionally show that pretraining the single-backbone variant on over 160k AgiBotWorld-Beta trajectories further transfers cross-embodiment manipulation knowledge to physical robots. E0.5 embodies this geometry-native approach as a member of the \textbf{\textcolor{brandTeal}{VWA}} family of world-action models.
\end{abstract}

\begin{document}
\maketitle

\section{\teal{Intro}\graytext{duction}}
\label{sec:introduction}

\begin{figure}[t]
    \centering
    \includegraphics[width=\textwidth]{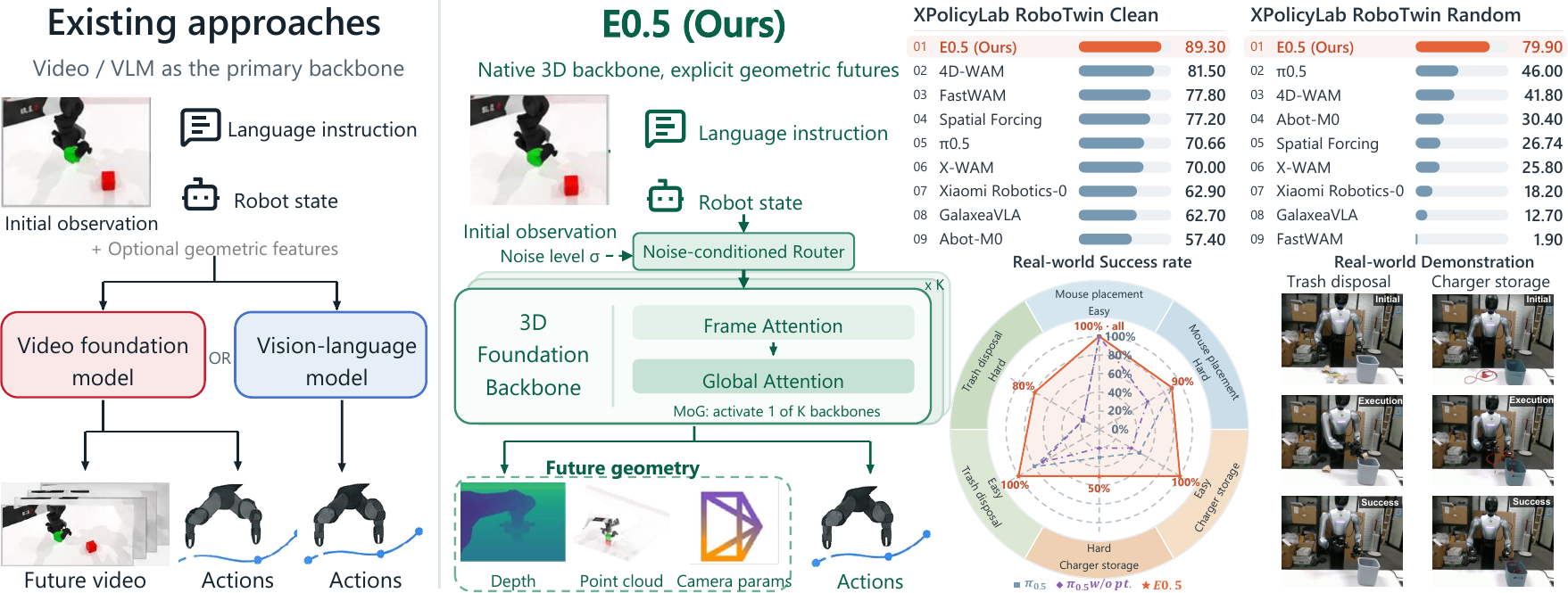}
    \caption{\textbf{Overview of E0.5 and representative results.} \textbf{Left:} Video- and VLM-centered backbones, optionally augmented with geometry. \textbf{Center:} E0.5 jointly predicts actions and action-horizon-aligned future geometry with trainable 3D backbones. Mixture-of-Geometry expands capacity while activating one backbone per action noise level. \textbf{Right:} XPolicyLab RoboTwin comparisons, Zerith H1 success rates under EASY/HARD settings, and execution examples.}
    \label{fig:e05_overview}
\end{figure}

Robotic manipulation requires understanding not only the current spatial structure of a scene, but also how that structure may change as the robot acts. World-action models (WAMs) provide a framework for learning this relationship by jointly modeling robot actions and future world states~\citep{song2025physical,ye2026world}. A central design question is therefore what kind of representation should underpin this joint modeling of actions and their physical consequences. Mainstream WAMs build on video foundation models or vision-language models, leveraging their capabilities in visual generation and semantic understanding. Although these models can incorporate geometric information, their primary computation remains organized around a video or vision-language backbone, rather than a native geometric foundation. This distinction motivates us to investigate an alternative: building WAMs directly on 3D foundation models, so that action generation and future-state prediction are learned within a shared geometric backbone.

Introducing 3D information into an existing model only partially addresses this requirement. Recent methods incorporate features extracted by geometric models into a vision-language or video backbone, or use these features as auxiliary supervision for the backbone's representations~\citep{yang2026understanding,guo2025glad}. Such approaches enrich spatial understanding, but retain a vision-language or video foundation model as the primary backbone, rather than building directly on a 3D foundation model. More generally, geometry can serve as an additional input or prediction target without becoming the representational foundation for world modeling and action generation. We investigate a complementary approach: making a 3D foundation model the core backbone, training it to predict geometric futures together with actions, and expanding its capacity within this geometric modeling framework (\cref{fig:e05_overview}).

First, E0.5 adopts a \emph{3D foundation model as its primary backbone}, placing geometric representations at the center of both world modeling and action prediction. The geometric backbone serves as the core model on which these capabilities are learned, extending the role of 3D models beyond front-end feature extraction or auxiliary supervision. Our implementation uses VGGT-Omega~\citep{wang2026vggtomega} as the pretrained geometric backbone. While such models recover the geometry of observed scenes, E0.5 develops this foundation into a \emph{predictive geometric backbone} that learns future world states jointly with robot actions.

Second, E0.5 \emph{explicitly predicts future 3D states together with actions}. Reconstructing current geometry provides information about the scene at the time of observation; predicting future geometry additionally requires modeling how its spatial configuration evolves during interaction. For each demonstrated action chunk, we supervise depth, pointmaps, and camera parameters at the frame aligned with its final step. The policy must predict this endpoint from the current observations, instruction, and robot state. It also offers an alternative to future video prediction, whose targets encompass both geometric changes and variations in visual appearance. Our hypothesis is that directly supervising future 3D states encourages the backbone to learn spatial structure and dynamics that remain useful across environments, supporting generalization when observations differ from those encountered during training.

This design also advances the use of 3D backbones beyond current-state reconstruction and restricted adaptation. VGA~\citep{song2026robotic} uses a native 3D backbone with current-scene geometric supervision and LoRA adaptation; its ablations report substantial degradation under full-parameter training. E0.5 instead unfreezes the geometric backbone and jointly optimizes it for future 3D prediction and action generation. Our experiments demonstrate that backbone optimization is effective in this formulation, allowing pretrained geometric representations to adapt to predictive world-action modeling. Thus, the distinction concerns both \emph{what the backbone predicts} and \emph{how its representations are learned}.

Third, we introduce \emph{Mixture-of-Geometry} to scale model capacity by combining multiple geometric backbones. Each constituent backbone participates in future 3D prediction, connecting the additional parameter capacity to the same predictive objective. This architecture expands capacity directly within the model's geometric foundation and supports joint learning of geometric futures and actions at larger model sizes. Future geometry prediction and backbone mixtures therefore work together: the prediction objective specifies what the geometric representations must learn, while the mixture architecture increases the capacity available for learning it.

On XPolicyLab RoboTwin~\citep{chen2025robotwin,community2026xpolicylab}, predicting future geometry instead of reconstructing the current scene improves task success by more than 10 percentage points under scene randomization. With the prediction target held fixed, expanding the geometric backbone through Mixture-of-Geometry further improves randomized success by 5.64 percentage points over a single backbone. In the main clean-to-random evaluation, E0.5 achieves 79.90\% randomized success, compared with 1.92\% for FastWAM and 25.84\% for X-WAM. On LIBERO~\citep{libero}, E0.5 achieves 98.90\% overall success across 40 tasks, while maintaining strong generalization on LIBERO-Plus (88.78\%)~\citep{libero_plus} and LIBERO-PRO (55.36\%)~\citep{libero_pro}. We further evaluate E0.5 on Zerith H1 and AgileX Piper under object and configuration shifts. Across three manipulation tasks, E0.5 achieves mean success rates of 73.3\% and 66.7\% under the HARD setting on the two platforms, respectively. Despite using no embodied-data pretraining, it outperforms embodied-pretrained $\pi_{0.5}$ on both platforms under both in-distribution and out-of-distribution settings. In addition, we investigate extending the single-backbone variant to large-scale embodied pretraining on AgiBotWorld-Beta~\citep{bu2025agibot}, confirming that joint future-geometry and action pretraining further improves downstream physical manipulation.

Our contributions are threefold:
\begin{itemize}
    \item \textbf{World-action models built on 3D foundations.} We introduce E0.5, which adopts 3D foundation models as its primary backbones and develops their geometric representations for joint world modeling and action prediction.
    \item \textbf{Explicit future 3D state prediction.} We jointly learn future geometry and actions while optimizing the geometric backbone, achieving more than a 10-percentage-point improvement over current-state reconstruction and strong OOD generalization relative to video-prediction-based WAMs.
    \item \textbf{Parameter scaling through Mixture-of-Geometry.} By routing across geometric backbones trained to predict the same future state, E0.5 expands the capacity devoted to geometric reasoning and improves manipulation success over a single predictive backbone.
\end{itemize}

\section{\teal{Related} \graytext{Work}}
\label{sec:related_work}

\paragraph{Vision-Language-Action and Video-Based World-Action Models.}
Vision-language-action (VLA) models transfer pretrained visual and linguistic representations to robot control~\citep{zhan2026e0,zhan2026stable,zhou2026tag,li2026vla}. OpenVLA~\citep{kim2024openvla} adapts a vision-language backbone using large-scale robot demonstrations, while $\pi_{0.5}$~\citep{physicalintelligence2025pi05} combines heterogeneous robot data, web data, and semantic supervision to improve open-world generalization. These approaches demonstrate the value of broad pretraining and multimodal knowledge transfer, with vision-language models providing the principal representational foundation for action generation. Despite these advances, strong benchmark performance does not necessarily translate into robust real-world generalization, with recent evaluations revealing sensitivity to physical perturbations and limitations in spatial reasoning~\citep{chen2026radar}.

A complementary line of research couples action learning with future visual prediction. Unified Video Action Model~\citep{li2025unified} learns a shared video-action latent representation with separate diffusion decoders, while Video Policy~\citep{liang2025video} jointly trains video and action generation and leverages action-free videos for policy generalization. PAR~\citep{song2025physical,song2026physics} and DreamZero~\citep{ye2026world} build world-action models on pretrained video diffusion backbones, jointly predicting future video and actions to transfer learned dynamics to new tasks and environments. These methods establish visual prediction as a useful source of supervision for embodied behavior. E0.5 shares the objective of learning actions together with future world states, but changes both the representational foundation and the prediction target: a 3D foundation model serves as the primary backbone, and future states are explicitly modeled through geometry rather than video appearance.

\paragraph{Geometric Foundations and Future 3D Prediction.}
Learning 3D structure from visual evidence is an important foundation for spatial intelligence~\citep{wang2023sparsenerf,shen2024geometry,he2026sceneexpander,zhao2026structsplat,wang2024perf,feng2025objfiller3d,chen2026style4d,xiang2025distilled}. More broadly, geometric foundation models learn transferable spatial representations. DUSt3R~\citep{wang2024dust3r} regresses point maps from image pairs to reconstruct scene geometry. VGGT~\citep{wang2025vggt} and VGGT-Omega~\citep{wang2026vggtomega} jointly estimate multiple geometric quantities from visual observations. Their reconstruction-oriented pretraining offers a geometric starting point for embodied learning. E0.5 builds on this class of models, extending their role from recovering observed geometry to learning geometric futures jointly with robot actions.

Several approaches introduce these geometric priors into existing VLA architectures. \citet{yang2026understanding} investigate geometric feature fusion and representation alignment within VLAs, while GLaD~\citep{guo2025glad} distills features from a frozen geometric teacher into the language backbone's visual-token representations. Such methods incorporate geometric knowledge while retaining a vision-language-centered policy architecture. Geometry also appears as a generative target: 3D-VLA~\citep{zhen2024vla} connects a 3D-aware language model with diffusion models that generate goal images and point clouds. Thus, using 3D information or generating geometric goals is distinct from making a geometric foundation model the primary, trainable backbone for world-action modeling.

VGA~\citep{song2026robotic} is more closely related in its use of a native geometric backbone for action generation. Its geometric supervision reconstructs the observed scene, and it uses LoRA adaptation to preserve pretrained backbone weights; its ablations report substantially lower performance with full-parameter training under the evaluated configuration. E0.5 instead explicitly predicts future 3D states and actions while unfreezing and jointly optimizing the geometric backbone. The distinction is therefore not simply the presence of a 3D backbone, but its transformation into a trainable predictive world-action model. Our comparison of future-state prediction with current-state reconstruction directly evaluates this distinction, while the OOD evaluations examine its implications for generalization.

\paragraph{Scaling through Mixture Architectures.}
Switch Transformers~\citep{fedus2022switch} expand language-model capacity through sparse expert layers. AdaMoE~\citep{shen2025expertise} expands a VLA action expert with feed-forward mixtures, while MoRE~\citep{gao2026more} introduces expert routing into a geometric foundation model for scene reconstruction. E0.5 instead mixes geometric backbones within a WAM, with each constituent trained to predict future geometry. This ties backbone-level capacity expansion to the same control-aligned predictive objective and enables direct evaluation of its effect on manipulation performance.

\section{\teal{Me}\graytext{thod}}
\label{sec:method}

\begin{figure}[t]
    \centering
    \includegraphics[width=0.95\linewidth]{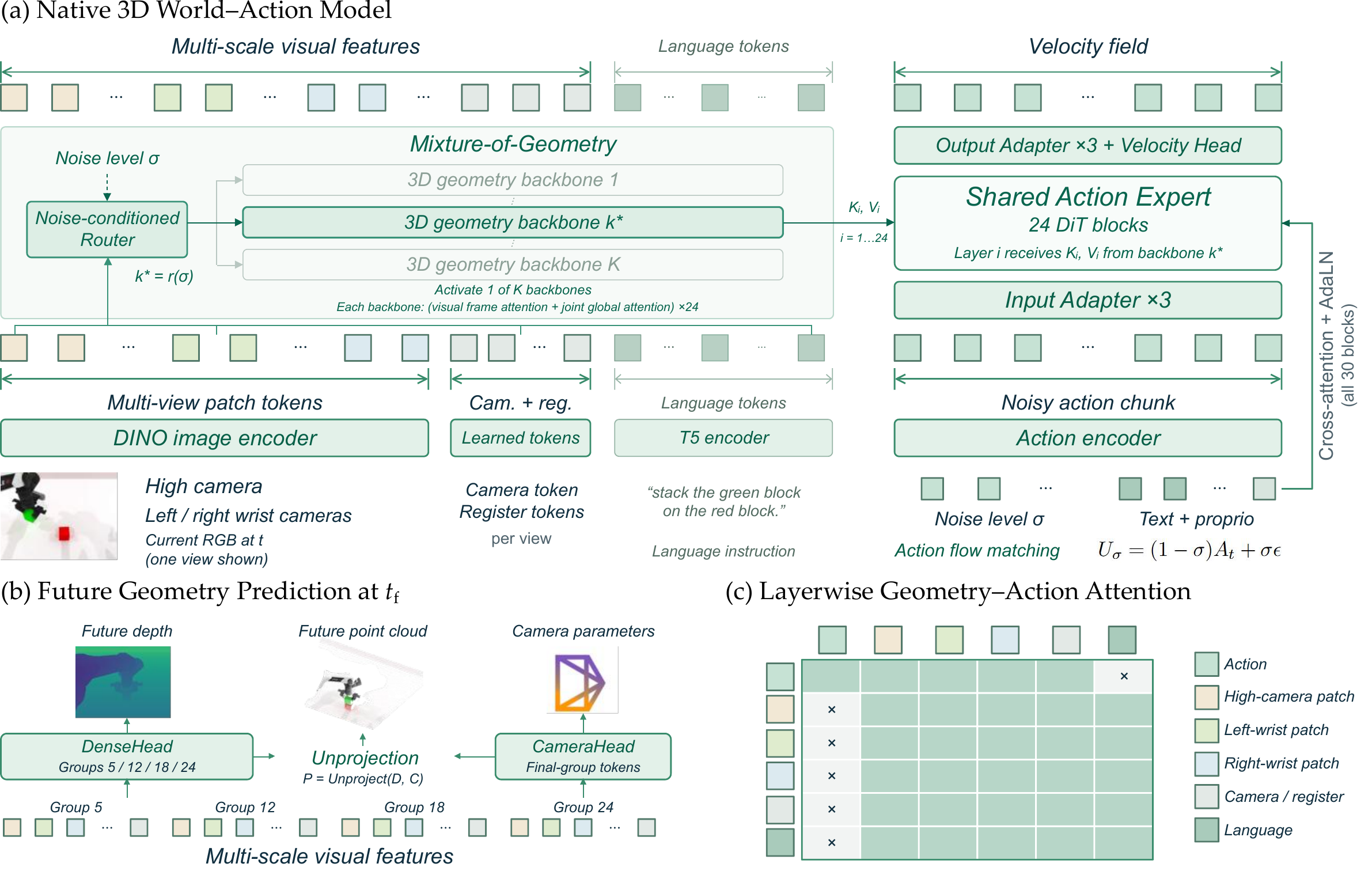}
    \caption{%
        \textbf{E0.5 architecture.}
        \textbf{(a)} At action noise level $\sigma$, the deterministic router $r(\sigma)$ selects one backbone to encode the current multi-view images and instruction. Its layer-wise visual keys and values condition a shared ActionDiT for flow-matching action generation.
        \textbf{(b)} Each backbone is trained to predict geometry at the same action-chunk endpoint $t_f$. The geometry heads predict depth and camera parameters, from which 3D points are obtained by unprojection.
        \textbf{(c)} The attention mask shows how action tokens attend to the selected backbone's visual tokens at each coupled layer.%
    }
    \label{fig:e05_architecture}
\end{figure}

E0.5 is designed around a single principle: \emph{the foundation used to generate actions should itself learn to anticipate geometric change}. This principle connects the three contributions introduced above. A 3D foundation model provides the primary representation; future geometry supervision adapts that representation to anticipating spatial change; and routed backbone mixtures increase the capacity trained under this objective. We therefore develop E0.5 as a \emph{future-predictive geometric foundation for world-action modeling}, rather than as a collection of independent perception and control modules. \Cref{fig:e05_architecture} gives an overview of E0.5.

\noindent{\textbf{World-action task.}}
At time $t$, the robot receives visual observations $O_t$, an instruction $\ell$, and its state $s_t$, collectively denoted by $x_t=(O_t,\ell,s_t)$. The target action chunk is $A_t=(a_t,\ldots,a_{t+H_a-1})$. Let $t_f>t$ denote the timestamp of the recorded frame corresponding to its final step. Alongside $A_t$, E0.5 predicts $G_t^{+}=(D_{t_f},P_{t_f},C_{t_f})$: the depth maps, point maps, and camera parameters of this future frame. Thus, the action and world-state targets describe the same interaction interval. Only $x_t$ is available to the policy; future observations and geometry supply training supervision, not model inputs.

\subsection{\teal{Building World-Action Models} \graytext{on a Native Geometric Foundation}}
\label{sec:geometric_backbone}

\paragraph{Geometry as the computational foundation.}
The first design choice concerns \emph{where geometric knowledge resides}. E0.5 makes a 3D foundation backbone $B_\theta$ the shared representation model for both action generation and world-state prediction:
\begin{equation}
\begin{aligned}
    h_t &= B_\theta(O_t,\ell),\\
    \widehat G_t^{+} &= Q_{\psi_g}(h_t),
    \qquad
    \widehat A_t \sim \pi_{\psi_a}(\,\cdot\mid h_t,\ell,s_t).
\end{aligned}
\label{eq:native_geometric_wam}
\end{equation}
Here, $Q_{\psi_g}$ is the geometric output computation and $\pi_{\psi_a}$ is the action decoder. Both operate on representations produced by the geometric foundation itself. Geometry is therefore not a front-end feature passed to a separate VLM or video foundation backbone. We instantiate $B_\theta$ with an existing 3D foundation architecture~\citep{wang2026vggtomega}, but the modeling principle is the use and adaptation of the geometric foundation, rather than a particular backbone implementation.

\paragraph{Task conditioning within geometric computation.}
To support instruction-following control without replacing this foundation, we introduce language into the backbone's global scene computation. Let $Z_i^{(j)}$ denote the visual and native geometric tokens for observed image $i$ at stage $j$, and let $L^{(j)}$ denote language tokens. The conditioning pattern is
\begin{equation}
\begin{aligned}
\widetilde Z_i^{(j)} &= \mathrm{FrameAttn}_j(Z_i^{(j)}), \\
X^{(j)} &= [\widetilde Z_1^{(j)}; \ldots; \widetilde Z_N^{(j)}; L^{(j)}], \\
(Z^{(j+1)}, L^{(j+1)}) &= \mathrm{GlobalAttn}_j(X^{(j)}).
\end{aligned}
\label{eq:language_attention}
\end{equation}
Language tokens are appended after image tokens only in global attention, where information is exchanged across observations. Frame attention retains its within-image token organization and excludes language tokens. This separates direct instruction fusion from frame-local processing, while allowing later frame-attention layers to process visual features already informed by global language interactions. \Cref{eq:language_attention} isolates the image--language attention pattern, ensuring that the resulting task-conditioned representation remains native to the geometric backbone.

\subsection{\teal{From Scene Reconstruction} \graytext{to Future-Predictive Adaptation}}
\label{sec:future_geometry_prediction}

A geometric foundation is not, by itself, a world-action model. Recovering the structure visible in $O_t$ can provide a strong spatial representation, but does not require predicting the scene associated with the forthcoming action chunk. E0.5 changes \emph{what the geometric foundation is trained to explain}: its output is the geometry at $t_f$, rather than the geometry of the input frame.

\paragraph{Action-horizon-aligned geometric prediction.}
We train the geometric output computation to produce
\begin{equation}
    Q_{\psi_g}(B_\theta(O_t,\ell))
       =\left(\widehat D_{t_f},\widehat P_{t_f},\widehat C_{t_f}\right).
    \label{eq:future_geometry_decoder}
\end{equation}
All three quantities refer to the same future endpoint. Depth describes the future visible scene layout, pointmaps describe its 3D spatial configuration, and camera parameters describe the associated imaging geometry.

Selecting the final frame of the action chunk gives world modeling a control-relevant horizon: the representation must support both a sequence of movements and the spatial state paired with that sequence in the demonstration. This alignment does not require rendering intermediate video frames. It extends the joint future-state and action modeling direction of PAR~\citep{song2025physical} into a native geometric foundation, with future 3D structure as the explicit prediction target.

\paragraph{Geometric supervision directed toward the future.}
For a predicted endpoint state, the three output-supervision terms are
\begin{equation}
\begin{aligned}
    \ell_{\mathrm{geo}}(\widehat G_t^{+},G_t^{+})
       ={}&\lambda_D\ell_D(\widehat D_{t_f},D_{t_f})\\
          &+\lambda_P\ell_P(\widehat P_{t_f},P_{t_f})\\
          &+\lambda_C\ell_C(\widehat C_{t_f},C_{t_f}).
\end{aligned}
\label{eq:future_geometry_loss}
\end{equation}
We adopt the depth, pointmap, and camera loss definitions of \citet{wang2026vggtomega}, with confidence terms and validity masks implicit. The essential change is temporal: the same geometric supervision is applied to the unobserved future frame. Replacing $G_t^{+}$ with the observed state $G_t$ yields the reconstruction counterpart examined in our ablation, not an intermediate stage of the proposed method.

This choice also clarifies the generalization motivation. A future video target supervises appearance as well as scene evolution; our explicit targets instead ask the model to explain spatial change through depth, 3D locations, and cameras. We hypothesize that this directs representation learning toward spatial information useful under visual distribution shifts. The OOD evaluations test whether this predictive geometric representation transfers beyond the training distribution.

\paragraph{Prediction-driven, full-parameter adaptation.}
Future prediction supervises the backbone, not only a decoder attached to fixed features. We jointly optimize the geometric backbone, input-conditioning modules, geometric prediction heads, and action decoder, while keeping the pretrained language encoder frozen. Action learning shapes the representation for control, while future geometry supervision provides an additional structural learning signal throughout the trainable geometric backbone.

This distinguishes E0.5 from using a native 3D backbone for current-scene reconstruction with restricted adaptation, as in VGA~\citep{song2026robotic}. E0.5 combines a future-state objective with full-parameter world-action learning: the geometric foundation is adapted to anticipation as well as perception. The design therefore changes both the temporal target and the trainable representation, rather than merely replacing a prediction head.

\subsection{\teal{Scaling Future-Predictive Capacity} \graytext{with Routed Backbone Mixtures}}
\label{sec:mixture_of_geometry}

Mixture-of-Geometry (MoG) scales the capacity of the future-predictive foundation. It expands the model into a bank of $K$ trainable geometric backbones, $\mathcal B_K=\{B_{\theta_1},\ldots,B_{\theta_K}\}$, with routing that selects which backbones participate in the computation. This places additional parameter capacity in the representations used to model geometric futures and actions, rather than only in an action decoder.

\paragraph{Routing over geometric backbones.}
MoG uses deterministic hard routing over action denoising time. The noise distribution is partitioned into $K$ intervals with equal probability under the training-time sampling distribution, and each training example activates one geometric expert. The router has no trainable parameters and depends only on the action noise level, rather than on the content of the observations or instruction.

Let $\sigma\in[0,1]$ denote the action noise level, with $\sigma=1$ corresponding to noise and $\sigma=0$ to the clean action chunk. During training, we sample $u\sim\mathcal U(0,1)$ and apply the shift transformation
\begin{equation}
\begin{aligned}
\sigma &= f_s(u) = \frac{su}{1+(s-1)u}, \\
b_i    &= f_s(i/K), \quad i=1,\ldots,K-1,
\end{aligned}
\label{eq:routing_boundaries}
\end{equation}
where $s>0$ is the training-time shift. The selected expert is
\begin{equation}
    r(\sigma)
      =K-\sum_{i=1}^{K-1}\mathbf 1[\sigma\ge b_i],
    \qquad
    \mathcal S(\sigma)=\{r(\sigma)\}.
    \label{eq:routing_selection}
\end{equation}
Experts are indexed from the highest-noise interval ($k=1$) to the lowest-noise interval ($k=K$). The intervals have equal probability under the continuous training distribution.

The geometric representation therefore depends on the current denoising stage:
\begin{equation}
    \begin{aligned}
        \mathcal H_t(\sigma)
            &= B_{\theta_{r(\sigma)}}(O_t,\ell), \\
        h_t^{\mathrm{act}}(\sigma)
            &= \operatorname{Readout}(\mathcal H_t(\sigma)).
    \end{aligned}
    \label{eq:geometry_mixture}
\end{equation}
Here, $\operatorname{Readout}$ extracts the layer-wise visual keys and values from the selected geometric backbone. All stages share one action decoder. During inference, denoising proceeds from high to low noise and can switch between geometric experts. These stages partition action denoising time rather than successive physical future timestamps: all experts are trained to predict geometry at the same action-horizon endpoint.

\paragraph{Future supervision accompanies capacity expansion.}
Each constituent backbone participates in future geometry prediction. Let $\mathcal E_t$ index the backbone-associated predictions evaluated for geometric supervision on an example. For $k\in\mathcal E_t$, let $h_t^{(k)}$ denote the corresponding representation and $Q_{\psi_g}^{(k)}$ its geometric output computation. The future objective is
\begin{equation}
\begin{aligned}
    \widehat G_t^{+,(k)} &= Q_{\psi_g}^{(k)}(h_t^{(k)}),\\
    \mathcal L_{\mathrm{future}}^{(K)}
      &=\mathbb E\!\left[
           \sum_{k\in\mathcal E_t}\,
           \ell_{\mathrm{geo}}(\widehat G_t^{+,(k)},G_t^{+})
         \right].
\end{aligned}
\label{eq:mixture_future_loss}
\end{equation}
\Cref{eq:mixture_future_loss} makes the relationship between prediction and scaling explicit: each constituent backbone is trained to support the same future-state task, coupling \emph{what is scaled}---the geometric foundation---with \emph{what supervises it}---future spatial structure.

\subsection{\teal{Joint Optimization} \graytext{and Action Generation}}
\label{sec:joint_optimization}

We train the full model with action generation and backbone-associated future geometry supervision:
\begin{equation}
    \min_{\Theta^{(K)}}\;
       \mathcal L_{\mathrm{act}}
          +\lambda_g\mathcal L_{\mathrm{future}}^{(K)}.
    \label{eq:joint_world_action_objective}
\end{equation}
We jointly optimize the geometric backbones, input-conditioning modules, geometric prediction heads, and shared action decoder while keeping the pretrained language encoder frozen. Future geometry supervision updates the geometric representations that also condition action generation, allowing both objectives to shape the backbone during adaptation.

For continuous action chunks, we use a flow-matching decoder~\citep{lipman2022flow}. Given a demonstrated chunk $A_t$ and Gaussian noise $\epsilon\sim\mathcal N(0,I)$, we construct
\begin{equation}
    U_\sigma=(1-\sigma)A_t+\sigma\epsilon,
    \qquad \sigma\in[0,1],
    \label{eq:action_flow_path}
\end{equation}
where $\sigma$ is sampled from the shifted distribution in \cref{eq:routing_boundaries}. The shared action decoder learns the velocity $\epsilon-A_t$, conditioned on the geometric expert selected at the current noise level:
\begin{equation}
\begin{aligned}
    \mathcal L_{\mathrm{act}}
      &=\mathbb E\!\left[
          \left\|\widehat v_\sigma-(\epsilon-A_t)\right\|^2
        \right],\\
    \widehat v_\sigma
      &=v_{\psi_a}\!\left(
          U_\sigma,\sigma;
          h_t^{\mathrm{act}}(\sigma),\ell,s_t
        \right).
\end{aligned}
    \label{eq:flow_matching_loss}
\end{equation}
At deployment, the policy requires only current visual observations, the instruction, and the robot state; future geometric ground truth is never required during inference.

\section{\teal{Exp}\graytext{eriments}}
\label{sec:experiment}

\subsection{\teal{Implementation} \graytext{Details}}
\label{sec:implementation_details}

We initialize two geometric backbones ($K=2$, yielding a 3.14B-parameter policy core) from pretrained VGGT-Omega~\citep{wang2026vggtomega}, including its depth and camera decoding heads. The shared action expert follows WAN2.2's DiT architecture~\citep{wan2025} with hidden dimension 1024 and 30 blocks. A mixture-of-transformers (MoT) couples the selected geometric backbone and the action expert across blocks 4--27, while the expert's first and last three blocks serve as uncoupled input and output adapters. A frozen T5 encoder provides language conditioning to both components; the geometric backbones remain fully trainable. We sample noise levels using the shifted uniform distribution $\sigma = f_5(u) = \frac{5u}{1+4u}$ during training and run 10 Euler denoising steps with classifier-free guidance (CFG) scale 1.0 at inference time. The predicted action horizon is $H_a = 32$ on XPolicyLab RoboTwin and real robots, and $H_a = 10$ on LIBERO. Across all training configurations, trainable modules are optimized with AdamW at peak learning rate $1 \times 10^{-4}$, weight decay 0.01, and cosine annealing. The outer future geometry loss weight is set to $\lambda_g = 0.01$.

\subsection{\teal{Simulation} \graytext{Experiments}}
\label{sec:sim_experiments}

\begin{table}[t]
\centering
\caption{\textbf{Performance (success rate, \%) on XPolicyLab RoboTwin.} All results are reported under clean-only post-training and evaluated on clean and randomized scenes. Policy-core sizes count active inference-time policy backbones and action modules (excluding frozen text encoders, VAEs, and optional auxiliary modules; $\sim$ denotes nominal architecture-based estimates, and \texttt{--} denotes unreported comparable counts).}
\label{tab:robotwin_c2r}
\setlength{\tabcolsep}{5.5pt}
\renewcommand{\arraystretch}{1.12}
\small
\resizebox{\textwidth}{!}{%
\begin{tabular}{lrcrrr}
\toprule
\textbf{Method} & \shortstack{\textbf{Policy core}\\\textbf{params. (B)} $\downarrow$} & \shortstack{\textbf{Embodied}\\\textbf{pretrain.}} & \textbf{Clean} $\uparrow$ & \textbf{Randomized} $\uparrow$ & \textbf{Average} $\uparrow$ \\
\midrule
\rowcolor{eGroup}
\multicolumn{6}{l}{\textit{VLA}} \\
StarVLA~\citep{community2026starvla} & -- & \ptNo & 46.52 & 3.16 & 24.84 \\
GalaxeaVLA~\citep{jiang2026galaxy} & -- & $\sim$2k h & 62.72 & 12.72 & 37.72 \\
XR-0~\citep{cai2026xiaomi} & 4.70 & $\sim$1.9k h & 62.86 & 18.22 & 40.54 \\
ABot-M0~\citep{yang2026abot} & $\sim$4.16 & $\sim$9.5k h & 57.40 & 30.36 & 43.88 \\
Spatial Forcing~\citep{li2026spatial} & $\sim$3.30 & $\sim$10k h & 77.18 & 26.74 & 51.96 \\
$\pi_{0.5}$~\citep{physicalintelligence2025pi05} & $\sim$3.30 & $\sim$10k h & 70.66 & 46.00 & 58.33 \\
GigaBrain-0.7~\citep{team2026gigabrain} & 3.50 & $\sim$37.3k h & 66.80 & 67.92 & 67.36 \\
\rowcolor{eGroup}
\multicolumn{6}{l}{\textit{WAM}} \\
FastWAM~\citep{yuan2026fast} & 6.00 & \ptNo & 77.76 & 1.92 & 39.84 \\
X-WAM~\citep{guo2026unified} & $\sim$5.04 & $\sim$5.9k h & 69.98 & 25.84 & 47.91 \\
4D-WAM~\citep{yang20264d} & 5.30 & $\sim$16k h & 81.50 & 41.80 & 61.65 \\
ImageWAM~\citep{zhang2026imagewam} & $\sim$4.64 & \ptNo & 84.40 & 18.30 & 51.35 \\
\midrule
\rowcolor{eOurs}
\textbf{E0.5 (Ours)} & \textbf{3.14} & \ptNo & \textbf{89.30} & \textbf{79.90} & \textbf{84.60} \\
\bottomrule
\end{tabular}%
}
\end{table}

\begin{figure}[t]
    \centering
    \includegraphics[width=\textwidth]{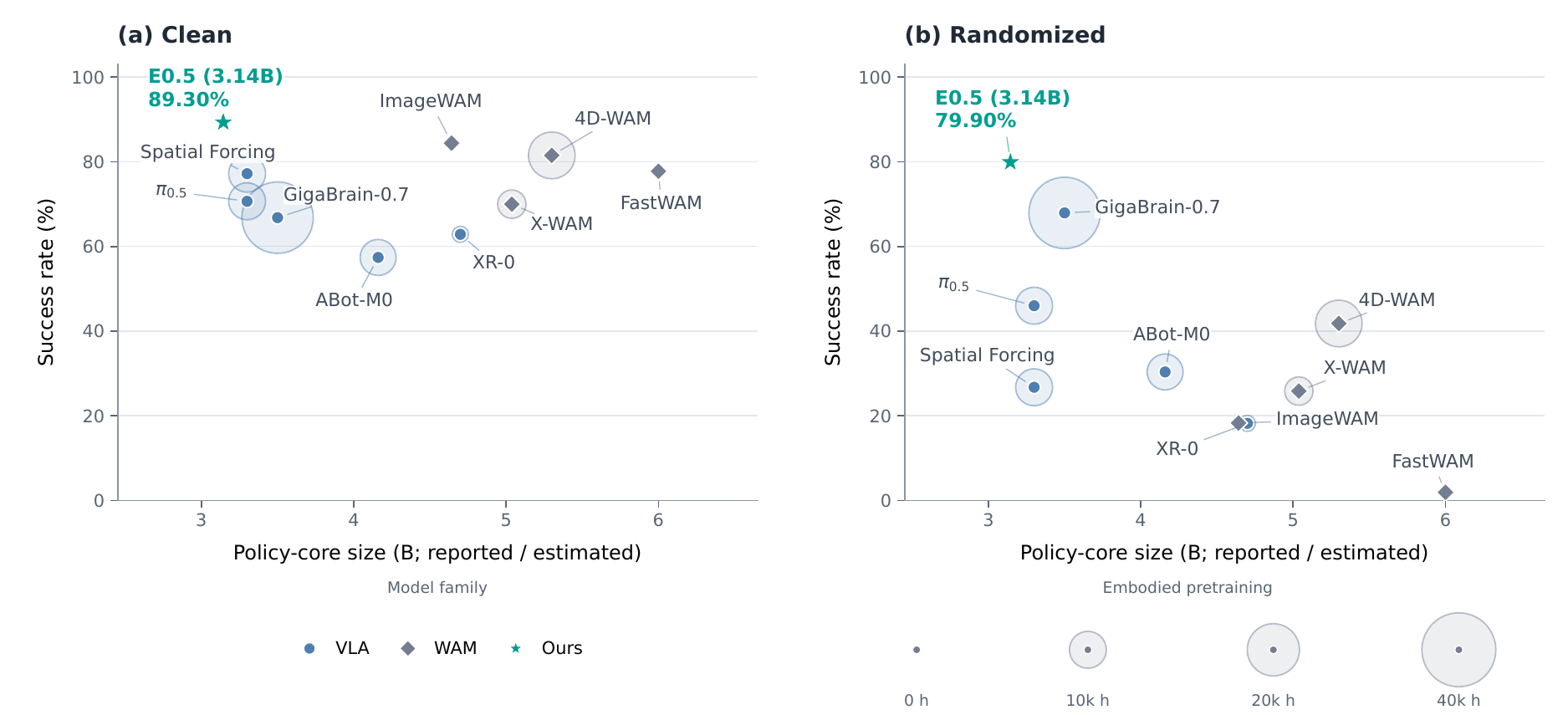}
    \caption{%
        \textbf{Model size, embodied pretraining, and generalization on XPolicyLab RoboTwin.}
        Success rates and embodied pretraining durations follow \cref{tab:robotwin_c2r}, under clean-only post-training and evaluation in (a) clean and (b) randomized scenes.
        Circles denote VLA baselines, diamonds WAM baselines, and the star E0.5.
        Translucent circle area is proportional to embodied pretraining duration, with a shared area scale across both panels; no background circle denotes no embodied pretraining.
        The horizontal axis uses reported nominal model sizes or policy-core estimates.%
    }
    \label{fig:robotwin_size_success}
\end{figure}

\begin{table}[t]
\centering
\caption{\textbf{Performance (success rate, \%) on LIBERO and robustness variants.} Results follow the respective benchmark evaluations. \texttt{--} denotes an unavailable or unverified result.}
\label{tab:libero_all}
\setlength{\tabcolsep}{3pt}
\renewcommand{\arraystretch}{1.12}
\resizebox{\textwidth}{!}{%
\begin{tabular}{lc*{19}{r}}
\toprule
& & \multicolumn{5}{c}{\cellcolor{eLibero}\textbf{\textit{LIBERO}}}
& \multicolumn{8}{c}{\cellcolor{ePlus}\textbf{\textit{LIBERO-Plus}}}
& \multicolumn{6}{c}{\cellcolor{ePro}\textbf{\textit{LIBERO-PRO}}} \\
\cmidrule(lr){3-7}\cmidrule(lr){8-15}\cmidrule(lr){16-21}
\textbf{Method} & \shortstack{Embodied\\pretrain.}
& Spatial & Object & Goal & Long & Overall
& Camera & Robot & Lang. & Light & Bg. & Noise & Layout & Overall
& Object & Pos. & Sem. & Task & Env. & Overall \\
\midrule
\rowcolor{eGroup}
\multicolumn{21}{l}{\textit{VLA}} \\
StarVLA~\citep{community2026starvla} & \ptNo & 97.80 & 98.60 & 96.20 & 93.80 & 96.60 & 64.04 & 57.16 & 85.75 & 93.78 & 96.00 & 80.95 & 77.25 & 77.83 & 76.65 & 2.80 & 94.85 & 17.80 & 45.05 & 47.43 \\
ABot-M0~\citep{yang2026abot} & $\sim$9.5k h & 98.80 & 99.80 & 99.00 & 96.60 & 98.60 & 60.40 & 67.90 & 86.40 & 96.20 & 91.60 & 86.40 & 82.60 & 80.50 & -- & -- & -- & -- & -- & -- \\
Spatial Forcing~\citep{li2026spatial} & $\sim$10k h & 99.40 & 99.60 & 98.80 & 96.00 & 98.50 & 56.85 & 37.94 & 84.39 & 95.36 & 91.82 & 65.90 & 79.02 & 71.10 & 73.95 & 2.05 & 94.45 & 9.80 & 31.85 & 42.42 \\
$\pi_{0.5}$~\citep{physicalintelligence2025pi05} & $\sim$10k h & 98.80 & 98.20 & 98.00 & 92.40 & 96.85 & 78.40 & 73.60 & 80.80 & 96.20 & 94.10 & 89.00 & 84.50 & 84.40 & 95.90 & 20.85 & 95.70 & 1.10 & 52.40 & 53.19 \\
\rowcolor{eGroup}
\multicolumn{21}{l}{\textit{WAM}} \\
FastWAM~\citep{yuan2026fast} & \ptNo & 98.20 & 100.00 & 97.00 & 95.20 & 97.60 & 16.01 & 46.26 & 72.93 & 81.96 & 55.48 & 39.41 & 66.49 & 52.56 & 73.00 & 0.05 & 82.90 & 14.90 & 31.70 & 40.51 \\
X-WAM~\citep{guo2026unified} & $\sim$5.9k h & 96.80 & 98.60 & 95.40 & 95.60 & 96.60 & 72.92 & 46.32 & 88.48 & 93.61 & 88.29 & 62.90 & 80.79 & 74.80 & 83.95 & 21.20 & 97.25 & 17.70 & 45.55 & 53.13 \\
4D-WAM~\citep{yang20264d} & $\sim$16k h & 98.80 & 99.50 & 98.60 & 97.40 & 98.60 & 45.15 & 64.26 & 90.63 & 94.29 & 57.71 & 69.08 & 79.21 & 71.01 & 80.75 & 3.90 & 95.85 & 21.10 & 40.10 & 48.34 \\
\midrule
\rowcolor{eOurs}
\textbf{E0.5 (Ours)} & \ptNo & 99.40 & 99.40 & 98.60 & 98.20 & \textbf{98.90} & 84.37 & 83.74 & 88.03 & 98.51 & 95.54 & 90.38 & 85.57 & \textbf{88.78} & 83.45 & 23.05 & 94.35 & 22.15 & 53.80 & \textbf{55.36} \\
\bottomrule
\end{tabular}%
}
\end{table}

We evaluate task success on XPolicyLab RoboTwin, LIBERO, LIBERO-Plus, and LIBERO-PRO~\citep{chen2025robotwin,community2026xpolicylab,libero,libero_plus,libero_pro}, comparing native geometric world-action learning with representative VLA and WAM baselines.

\paragraph{XPolicyLab RoboTwin.}
We follow the clean-to-random protocol on this bimanual manipulation benchmark~\citep{chen2025robotwin,community2026xpolicylab}, training with 50 clean demonstrations per task and evaluating 100 trials per task under both clean and randomized conditions. Without embodied-data pretraining, E0.5 achieves 89.30\% Clean and 79.90\% Randomized success, with an 84.60\% average (\cref{tab:robotwin_c2r}). This exceeds $\pi_{0.5}$ and FastWAM by 26.27 and 44.76 average percentage points, respectively. Under Randomized evaluation, E0.5 reaches 41.6 times FastWAM's success rate (79.90\% versus 1.92\%), supporting future geometry as an effective alternative to video-centered world-action learning under scene variations.

The comparison with geometry-enhanced policies further tests the backbone choice: E0.5 exceeds Spatial Forcing, X-WAM, and 4D-WAM by 53.16, 54.06, and 38.10 randomized-success points, respectively. As summarized in \cref{fig:robotwin_size_success}, E0.5 achieves the highest success rate in both conditions with a compact 3.14B-parameter policy core---using approximately 40.7\% fewer policy-core parameters than 4D-WAM (5.30B) while improving Randomized success by 38.10 percentage points, and surpassing GigaBrain-0.7 (67.92\%, pretrained on $\sim$37.3k hours of embodied data) by 11.98 points under randomization without any prior embodied-data pretraining.

\paragraph{LIBERO Performance and Robustness.}
We evaluate a single unified E0.5 policy across the Spatial, Object, Goal, and Long suites of LIBERO~\citep{libero} and its robustness extensions, LIBERO-Plus~\citep{libero_plus} and LIBERO-PRO~\citep{libero_pro}. On the standard LIBERO benchmark, E0.5 achieves \textbf{98.90\%} overall success across 40 tasks and 2,000 episodes (99.40\% Spatial, 99.40\% Object, 98.60\% Goal, and 98.20\% Long), outperforming FastWAM (97.60\%), X-WAM (96.60\%), Spatial Forcing (98.50\%), 4D-WAM (98.60\%), and $\pi_{0.5}$ (96.85\%) (\cref{tab:libero_all}).

To assess robustness, we directly evaluate the same LIBERO-trained checkpoint without further fine-tuning. On LIBERO-Plus (10,030 perturbed episodes across seven categories), E0.5 achieves \textbf{88.78\%} overall success, surpassing $\pi_{0.5}$ (84.40\%) by 4.38 percentage points and consistently outperforming the compared VLA and WAM baselines. On the more challenging LIBERO-PRO benchmark (10,000 episodes across five types of distribution shifts), E0.5 achieves \textbf{55.36\%} average success, exceeding $\pi_{0.5}$ (53.19\%) by 2.17 points and achieving the strongest overall result among all compared methods. These results demonstrate that predictive geometric representations support both high nominal accuracy and strong robustness to visual, spatial, and semantic variations.

\begin{figure}[t]
    \centering
    \includegraphics[width=0.72\linewidth]{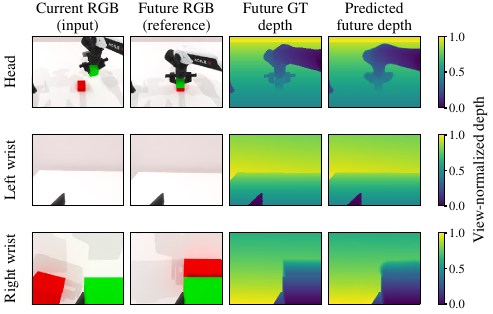}
    \caption{\textbf{Future geometry reconstruction example during closed-loop policy evaluation on \texttt{stack\_blocks\_two}.} Future RGB is shown solely as a visual reference and is never supplied as a model input.}
    \label{fig:future_reconstruction_best}
\end{figure}

\begin{table}[t]
\centering
\small
\setlength{\tabcolsep}{3.5pt}
\caption{\textbf{Training-set future geometry diagnostic on XPolicyLab RoboTwin.} All rows evaluate the same 200 samples (600 views across 50 tasks) at a prediction horizon of 32 frames (2.13\,s). Distances are in scene-normalized units; no post-hoc scale or pose alignment is applied.}
\label{tab:supp_geometry_offline}
\begin{tabular}{lrrrrrrrr}
\toprule
Expert & AbsRel $\downarrow$ & $\delta_1$ (\%) $\uparrow$ & MAE $\downarrow$ & RMSE $\downarrow$ & Point L2 $\downarrow$ & $e_R$ ($^\circ$) $\downarrow$ & $e_C$ $\downarrow$ & $e_f$ $\downarrow$ \\
\midrule
High-noise ($k=1$) & 0.020418 & 98.33 & 0.008123 & 0.030718 & 0.011464 & 0.630 & 0.004122 & 0.000099 \\
Low-noise ($k=2$)  & 0.020785 & 98.13 & 0.008350 & 0.030776 & 0.011826 & 0.701 & 0.004241 & 0.000083 \\
\rowcolor{eOurs}
Expert mean        & 0.020602 & 98.23 & 0.008236 & 0.030747 & 0.011645 & 0.666 & 0.004182 & 0.000091 \\
\bottomrule
\end{tabular}
\end{table}

\paragraph{Future geometry diagnostic.}
On 200 training-set samples from 50 XPolicyLab RoboTwin tasks (600 views), both routed geometric experts accurately predict scene geometry 32 frames (2.13\,s) ahead, achieving an expert-mean depth AbsRel of 0.0206, a $\delta_1$ accuracy of 98.23\%, a Point L2 error of 0.0116, and a mean camera rotation error of $0.666^\circ$ (\cref{tab:supp_geometry_offline}). Complementing this offline diagnostic, \cref{fig:future_reconstruction_best} visualizes future geometry predictions generated during actual closed-loop policy evaluation rather than on training demonstrations (see \cref{sec:supp_future_geometry} for quantitative closed-loop rollout diagnostics). E0.5 captures the overall spatial layout and future configuration across all three camera views. Although local errors remain around fast-moving robot arms and wrist-camera boundaries, the substantial manipulation gains over current-scene reconstruction at fixed backbone capacity (\cref{tab:ablation_future}) indicate that anticipating future 3D structure provides an effective representation-learning signal for control.

\subsection{\teal{Real-World} \graytext{Experiments}}
\label{sec:real_world_experiments}

\begin{figure}[t]
    \centering
    \includegraphics[width=0.95\linewidth]{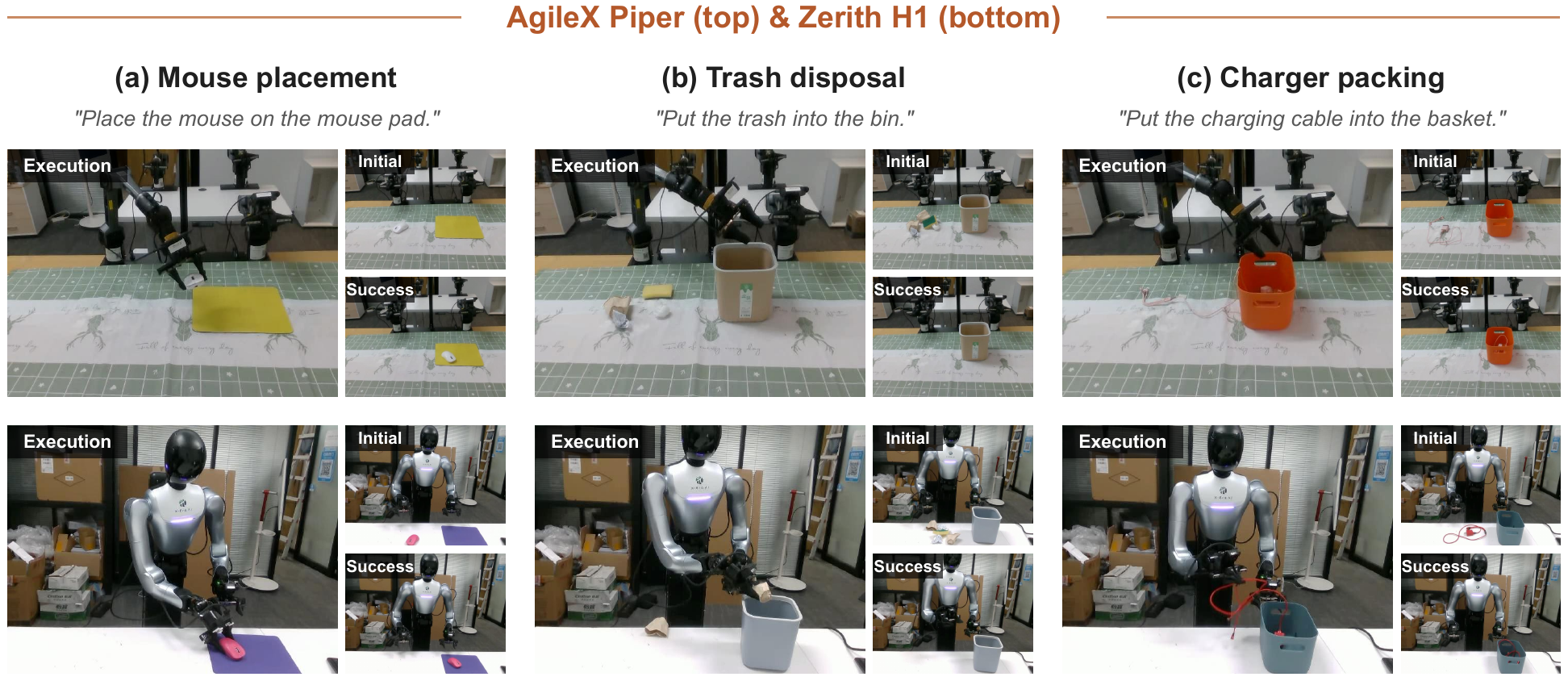}
    \caption{\textbf{Real-world tasks on AgileX Piper and Zerith H1.} Initial configurations, execution, and successful terminal states for mouse placement, trash disposal, and charger packing across both embodiments.}
    \label{fig:real_robot_setup}
\end{figure}

We evaluate E0.5 on the Zerith H1 humanoid robot and AgileX Piper dual-arm system across three manipulation tasks: \emph{mouse placement}, \emph{charger packing}, and \emph{trash disposal} (\cref{fig:real_robot_setup}). Baselines include official embodied-pretrained $\pi_{0.5}$ and Gemma-initialized $\pi_{0.5}$ without embodied-data pretraining ($\pi_{0.5}$ w/o pt.); E0.5 ($K=2$) uses no embodied-data pretraining. All policies undergo 5 epochs of multitask training with approximately 200 demonstrations per task on each platform, with DA3-estimated future depth supervision for E0.5. EASY (ID) uses training-distribution objects and configurations; HARD (OOD) introduces randomized mouse poses, unfamiliar chargers with thinner or different-material cables, and expanded trash collections containing six items (including unseen objects, compared with three items on Zerith H1 and four items on AgileX Piper under EASY). Each policy is evaluated over 10 trials per task and setting on each platform, totaling 120 trials per policy.

To distinguish complete task completion from partial progress, we report success rate (SR, \%) and mean normalized completion score (CS $\in [0,1]$). Mouse placement is binary and permits pushing onto the pad. Charger packing awards 0.25 for enclosing the rigid charger body and 0.75 for enclosing the entire flexible cable, requiring both for complete success. Trash disposal scores $k/N$ for depositing $k$ of $N$ designated objects into the bin, succeeding when $k=N$.

\begin{table}[t]
\centering
\caption{\textbf{Real-world manipulation performance on Zerith H1 and AgileX Piper.} We report per-task and three-task mean success rate (SR, \%) and normalized completion score (CS $\in [0,1]$) under EASY (ID) and HARD (OOD) settings (10 trials per task and setting). Here, $\pi_{0.5}$ uses official embodied-data pretraining, whereas $\pi_{0.5}$ w/o pt.\ and E0.5 ($K=2$) use no embodied-data pretraining. \textbf{Bold} indicates the best value within each platform and setting.}
\label{tab:h1_real_world}
\small
\setlength{\tabcolsep}{4.5pt}
\renewcommand{\arraystretch}{1.12}
\begin{tabular}{ll*{8}{r}}
\toprule
& & \multicolumn{2}{c}{\textbf{Mouse}} & \multicolumn{2}{c}{\textbf{Charger}} & \multicolumn{2}{c}{\textbf{Trash}} & \multicolumn{2}{c}{\textbf{Mean}} \\
\cmidrule(lr){3-4}\cmidrule(lr){5-6}\cmidrule(lr){7-8}\cmidrule(lr){9-10}
\textbf{Setting} & \textbf{Method} & SR $\uparrow$ & CS $\uparrow$ & SR $\uparrow$ & CS $\uparrow$ & SR $\uparrow$ & CS $\uparrow$ & SR $\uparrow$ & CS $\uparrow$ \\
\midrule
\multicolumn{10}{c}{\textbf{Zerith H1}} \\
\midrule
\multirow{3}{*}{EASY (ID)}
& $\pi_{0.5}$ & \textbf{100.0} & \textbf{1.000} & 50.0 & 0.600 & 80.0 & 0.867 & 76.7 & 0.822 \\
& $\pi_{0.5}$ w/o pt. & \textbf{100.0} & \textbf{1.000} & 40.0 & 0.450 & 70.0 & 0.867 & 70.0 & 0.772 \\
& \ourscell{\textbf{E0.5 (Ours)}} & \ourscell{\textbf{100.0}} & \ourscell{\textbf{1.000}} & \ourscell{\textbf{100.0}} & \ourscell{\textbf{1.000}} & \ourscell{\textbf{100.0}} & \ourscell{\textbf{1.000}} & \ourscell{\textbf{100.0}} & \ourscell{\textbf{1.000}} \\
\cmidrule(lr){2-10}
\multirow{3}{*}{HARD (OOD)}
& $\pi_{0.5}$ & \textbf{90.0} & \textbf{0.900} & 30.0 & 0.425 & 20.0 & 0.700 & 46.7 & 0.675 \\
& $\pi_{0.5}$ w/o pt. & 60.0 & 0.600 & 20.0 & 0.375 & 20.0 & 0.517 & 33.3 & 0.497 \\
& \ourscell{\textbf{E0.5 (Ours)}} & \ourscell{\textbf{90.0}} & \ourscell{\textbf{0.900}} & \ourscell{\textbf{50.0}} & \ourscell{\textbf{0.625}} & \ourscell{\textbf{80.0}} & \ourscell{\textbf{0.950}} & \ourscell{\textbf{73.3}} & \ourscell{\textbf{0.825}} \\
\midrule
\multicolumn{10}{c}{\textbf{AgileX Piper}} \\
\midrule
\multirow{3}{*}{EASY (ID)}
& $\pi_{0.5}$ & \textbf{100.0} & \textbf{1.000} & \textbf{80.0} & \textbf{0.850} & 60.0 & 0.850 & 80.0 & 0.900 \\
& $\pi_{0.5}$ w/o pt. & 50.0 & 0.500 & 50.0 & 0.625 & 40.0 & 0.550 & 46.7 & 0.558 \\
& \ourscell{\textbf{E0.5 (Ours)}} & \ourscell{\textbf{100.0}} & \ourscell{\textbf{1.000}} & \ourscell{70.0} & \ourscell{0.775} & \ourscell{\textbf{90.0}} & \ourscell{\textbf{0.975}} & \ourscell{\textbf{86.7}} & \ourscell{\textbf{0.917}} \\
\cmidrule(lr){2-10}
\multirow{3}{*}{HARD (OOD)}
& $\pi_{0.5}$ & \textbf{90.0} & \textbf{0.900} & 40.0 & 0.500 & \textbf{60.0} & 0.900 & 63.3 & 0.767 \\
& $\pi_{0.5}$ w/o pt. & 40.0 & 0.400 & 10.0 & 0.225 & 10.0 & 0.300 & 20.0 & 0.308 \\
& \ourscell{\textbf{E0.5 (Ours)}} & \ourscell{80.0} & \ourscell{0.800} & \ourscell{\textbf{60.0}} & \ourscell{\textbf{0.700}} & \ourscell{\textbf{60.0}} & \ourscell{\textbf{0.933}} & \ourscell{\textbf{66.7}} & \ourscell{\textbf{0.811}} \\
\bottomrule
\end{tabular}
\end{table}

As shown in \cref{tab:h1_real_world}, E0.5 achieves the highest three-task mean SR and CS on both platforms under both EASY and HARD settings. Under EASY, E0.5 attains 100.0\% SR across all 30 trials on Zerith H1 and 86.7\% SR on AgileX Piper. Under HARD, E0.5 reaches 73.3\% mean SR (0.825 CS) on Zerith H1 and 66.7\% mean SR (0.811 CS) on AgileX Piper, improving over $\pi_{0.5}$ w/o pt.\ by 40.0 and 46.7 percentage points, respectively, and outperforming embodied-pretrained $\pi_{0.5}$ (46.7\% on H1 and 63.3\% on Piper). On H1, HARD trash disposal improves from 20.0\% to 80.0\% SR despite unseen items and doubled object counts; on Piper, HARD charger packing improves from 40.0\% to 60.0\% SR. These results confirm that E0.5 transfers effectively to physical dual-arm and humanoid embodiments under object and configuration shifts, even without embodied-data pretraining.

\subsection{\teal{Ablation} \graytext{Studies}}
\label{sec:ablation}

\begin{table}[t]
\centering
\caption{\textbf{Future geometry and backbone capacity ablation study on XPolicyLab RoboTwin (success rate, \%).} All variants are trained under the controlled 15-epoch unaugmented protocol (effective batch size 128). \texttt{None} denotes training without geometry reconstruction supervision. \textbf{Bold} marks the best value in each column.}
\label{tab:ablation_future}
\small
\setlength{\tabcolsep}{10pt}
\begin{tabular}{clrrr}
\toprule
$K$ & \textbf{Target} & \textbf{Clean} $\uparrow$ & \textbf{Random} $\uparrow$ & \textbf{Avg.} $\uparrow$ \\
\midrule
1 & None       & 77.40 & 16.39 & 46.90 \\
1 & Current 3D & 82.98 & 16.58 & 49.78 \\
1 & Future 3D  & 86.18 & 29.39 & 57.79 \\
\midrule
2 & None       & 77.90 & 20.00 & 48.95 \\
2 & Current 3D & 84.63 & 19.59 & 52.11 \\
\rowcolor{eOurs}
2 & Future 3D  & 86.00 & \textbf{35.03} & \textbf{60.52} \\
\midrule
3 & Future 3D  & \textbf{86.88} & 32.76 & 59.82 \\
4 & Future 3D  & 81.82 & 15.52 & 48.67 \\
\bottomrule
\end{tabular}
\end{table}

We first isolate the core effects of future geometry supervision and backbone capacity in Mixture-of-Geometry on XPolicyLab RoboTwin under a controlled 15-epoch unaugmented training protocol with effective batch size 128 (\cref{tab:ablation_future}; see \cref{app:training_configurations} for comparisons under the augmented recipe).

\paragraph{Effect of future 3D prediction.}
Predicting future geometry consistently outperforms both current-scene reconstruction and training without geometric supervision across single- and dual-backbone configurations (\cref{tab:ablation_future}). The improvements are especially pronounced under scene randomization: compared with current 3D reconstruction, future 3D prediction increases Random success by 12.81 percentage points at $K=1$ (29.39\% vs.\ 16.58\%) and by 15.44 percentage points at $K=2$ (35.03\% vs.\ 19.59\%), while improving average success by 8.01 and 8.41 points, respectively. Supervising the action-chunk endpoint encourages the geometric backbone to capture control-relevant spatial transitions rather than static scene appearance, strengthening generalization under distribution shifts.

\paragraph{Effect of backbone capacity.}
Increasing the number of geometric backbones from $K=1$ to $K=2$ improves performance across all three supervision regimes, with the largest gains observed under future 3D prediction: Random success rises from 29.39\% to 35.03\% (+5.64 points), accompanied by a 2.73-point gain in average success (from 57.79\% to 60.52\%). This indicates that expanding geometric capacity via noise-routed experts complements predictive supervision. Further increasing $K$ under a fixed 15-epoch total training budget reduces per-expert sample exposure (\cref{sec:supp_architecture}): while $K=3$ achieves the highest Clean score (86.88\%) and still outperforms $K=1$ on average (59.82\%), $K=4$ degrades to 48.67\% average success. We therefore adopt $K=2$ as our default configuration.

\begin{table}[t]
\centering
\begin{minipage}[t]{0.50\linewidth}
    \vspace{0pt}
    \centering
    \caption{\textbf{Adapter placement ablation ($K=1$, 5 epochs).} $(n_{\mathrm{in}},n_{\mathrm{out}})$ counts uncoupled action blocks before/after the 24 coupled blocks on the action-only baseline.}
    \label{tab:supp_adapters}
    \small
    \setlength{\tabcolsep}{4.5pt}
    \begin{tabular}{ccrrr}
        \toprule
        $n_{\mathrm{in}}$ & $n_{\mathrm{out}}$ & Clean & Random & Avg. \\
        \midrule
        0 & 0 & 69.14 & 14.34 & 41.74 \\
        0 & 3 & 71.86 & 14.75 & 43.30 \\
        3 & 0 & 71.48 & 16.77 & 44.12 \\
        0 & 6 & 73.16 & 15.01 & 44.08 \\
        6 & 0 & \textbf{76.96} & 16.63 & \textbf{46.79} \\
        \rowcolor{eOurs}
        3 & 3 & 73.22 & \textbf{19.84} & 46.53 \\
        6 & 6 & 69.44 & 14.66 & 42.05 \\
        \bottomrule
    \end{tabular}
\end{minipage}%
\hfill
\begin{minipage}[t]{0.46\linewidth}
    \vspace{0pt}
    \centering
    \caption{\textbf{Future geometry loss weight sweep ($K=1$, 5 epochs).} Clean and Random are task-macro averages on RoboTwin without visual augmentation.}
    \label{tab:appendix_future_weight}
    \small
    \setlength{\tabcolsep}{5pt}
    \begin{tabular}{lrrr}
        \toprule
        $\lambda_g$ & Clean & Random & Avg. \\
        \midrule
        0.005 & 80.20 & 21.28 & 50.74 \\
        \rowcolor{eOurs}
        0.010 & 82.72 & \textbf{29.20} & \textbf{55.96} \\
        0.050 & \textbf{83.04} & 22.99 & 53.02 \\
        0.100 & 77.32 & 7.40  & 42.36 \\
        0.300 & 78.64 & 20.63 & 49.64 \\
        \bottomrule
    \end{tabular}
\end{minipage}
\end{table}

\begin{table}[t]
\centering
\small
\caption{\textbf{Comparison of future geometry prediction heads ($K=1$, 5 epochs) on XPolicyLab RoboTwin.} $\Delta$ denotes the change relative to the baseline without reconstruction in percentage points (pp).}
\label{tab:future_geometry_reconstruction}
\setlength{\tabcolsep}{8pt}
\begin{tabular}{lrrrr}
\toprule
\textbf{Reconstruction strategy} & \textbf{Clean} (\%) & $\Delta$ (pp) & \textbf{Random} (\%) & $\Delta$ (pp) \\
\midrule
No reconstruction & 73.22 & --- & 19.84 & --- \\
Auxiliary DepthDiT & 72.60 & $-0.62$ & 17.20 & $-2.64$ \\
\rowcolor{eOurs}
Native geometry heads & \textbf{82.72} & \textbf{+9.50} & \textbf{29.20} & \textbf{+9.36} \\
\bottomrule
\end{tabular}
\end{table}

\paragraph{Adapter placement, reconstruction loss weight, and native heads vs.\ auxiliary DepthDiT.}
We additionally summarize three architectural design ablations conducted with the single-backbone ($K=1$) model under a 5-epoch unaugmented screening budget (\cref{tab:supp_adapters,tab:appendix_future_weight,tab:future_geometry_reconstruction}):
\begin{itemize}
    \item \textbf{Input and output action adapters (\cref{tab:supp_adapters}).} Varying the number of uncoupled ActionDiT blocks $(n_{\mathrm{in}}, n_{\mathrm{out}})$ placed before and after the 24 geometry-coupled blocks on the action-only baseline shows that symmetric $(3,3)$ adapters achieve the highest Random success (19.84\%), improving over $(0,0)$ by 5.50 points and outperforming asymmetric $(6,0)$ and $(0,6)$ placements of equal total depth on randomized generalization.
    \item \textbf{Future geometry loss weight (\cref{tab:appendix_future_weight}).} Sweeping $\lambda_g \in \{0.005, 0.010, 0.050, 0.100, 0.300\}$ shows non-monotonic behavior: $\lambda_g = 0.010$ achieves the best Random (29.20\%) and average (55.96\%) success, balancing structural supervision with control adaptation.
    \item \textbf{Native geometry heads vs.\ auxiliary DepthDiT (\cref{tab:future_geometry_reconstruction}).} Motivated by WAMs that attach dedicated diffusion depth branches such as X-WAM~\citep{guo2026unified}, we compare supervising native VGGT-Omega heads against adding a flow-matching DepthDiT branch under the same 5-epoch $K=1$ setting. Adding an auxiliary DepthDiT slightly lowers performance relative to no reconstruction ($-0.62$ Clean and $-2.64$ Random points), whereas native geometry heads improve Clean and Random success by $+9.50$ and $+9.36$ points (improving 38/50 Clean and 44/50 Random tasks), confirming the advantage of supervising future geometry directly on the shared VGGT representations.
\end{itemize}

\subsection{\teal{Scaling Single-Backbone E0.5} \graytext{with Large-Scale Embodied Pretraining}}
\label{sec:embodied_pretraining}

While our primary $K=2$ Mixture-of-Geometry model already surpasses embodied-pretrained $\pi_{0.5}$ without prior robot-trajectory pretraining (\cref{tab:h1_real_world}), an important complementary question is whether E0.5's joint future-geometry and action formulation can further scale with large-scale cross-embodiment robot demonstrations. To investigate this dimension independently of backbone mixture scaling, we evaluate pretraining the single-backbone ($K=1$, 2.32B policy core) E0.5 architecture on AgiBotWorld-Beta~\citep{bu2025agibot}.

After modality filtering and preprocessing (detailed in \cref{appendix:pretrain-setting}), the pretraining corpus comprises 160,010 dual-arm trajectories spanning 177 tasks and 285.7M synchronized frames ($\approx$2,645.64 hours at 30\,Hz). We generate consistent multi-view metric depth pseudo-labels using Depth Anything 3 (\texttt{DA3METRIC-LARGE})~\citep{lin2025depth} and pretrain the $K=1$ model for two epochs (626,360 updates at global batch size 896) under the same joint flow-matching action and endpoint future-geometry objective, keeping the VGGT-Omega backbone fully trainable. We then apply the identical 5-epoch multitask post-training protocol on AgileX Piper and Zerith H1.

\begin{table}[t]
\centering
\caption{\textbf{Effect of large-scale AgiBotWorld-Beta embodied pretraining on the single-backbone ($K=1$, 2.32B) E0.5 variant.} We report success rate (SR, \%) and completion score (CS) under ID (EASY) and OOD (HARD) settings. For $\pi_{0.5}$~\citep{physicalintelligence2025pi05}, pretraining denotes the official pretrained checkpoint; for E0.5 ($K=1$), pretraining uses 160,010 trajectories from AgiBotWorld-Beta. \textbf{Bold} denotes the best value within each matched baseline--E0.5 pair.}
\label{tab:real_robot_pretrain_results}
\small
\setlength{\tabcolsep}{4.5pt}
\renewcommand{\arraystretch}{1.15}
\begin{tabular}{@{}ccc*{8}{c}@{}}
\toprule
\multirow{2}{*}{\textbf{Setting}} & \multirow{2}{*}{\textbf{Method}} & \multirow{2}{*}{\textbf{Pretrain}} & \multicolumn{2}{c}{\textbf{Mouse}} & \multicolumn{2}{c}{\textbf{Charger}} & \multicolumn{2}{c}{\textbf{Trash}} & \multicolumn{2}{c}{\textbf{Mean}} \\
\cmidrule(lr){4-5}\cmidrule(lr){6-7}\cmidrule(lr){8-9}\cmidrule(lr){10-11}
& & & SR & CS & SR & CS & SR & CS & SR & CS \\
\midrule
\multicolumn{11}{c}{\textbf{AgileX Piper}} \\
\midrule
\multirow{4}{*}{\textbf{ID}}
& $\pi_{0.5}$ & \ptNo & 50.0 & 0.500 & 50.0 & 0.625 & 40.0 & 0.550 & 46.7 & 0.558 \\
& \ourscell{\textbf{E0.5 ($K=1$)}} & \ourscell{\ptNo} & \ourscell{\textbf{90.0}} & \ourscell{\textbf{0.900}} & \ourscell{\textbf{70.0}} & \ourscell{\textbf{0.775}} & \ourscell{\textbf{100.0}} & \ourscell{\textbf{1.000}} & \ourscell{\textbf{86.7}} & \ourscell{\textbf{0.892}} \\
\cmidrule(lr){2-11}
& $\pi_{0.5}$ & \ptYes & \textbf{100.0} & \textbf{1.000} & 80.0 & 0.850 & 60.0 & 0.850 & 80.0 & 0.900 \\
& \ourscell{\textbf{E0.5 ($K=1$)}} & \ourscell{\ptYes} & \ourscell{\textbf{100.0}} & \ourscell{\textbf{1.000}} & \ourscell{\textbf{90.0}} & \ourscell{\textbf{0.925}} & \ourscell{\textbf{100.0}} & \ourscell{\textbf{1.000}} & \ourscell{\textbf{96.7}} & \ourscell{\textbf{0.975}} \\
\midrule
\multirow{4}{*}{\textbf{OOD}}
& $\pi_{0.5}$ & \ptNo & 40.0 & 0.400 & 10.0 & 0.225 & 10.0 & 0.300 & 20.0 & 0.308 \\
& \ourscell{\textbf{E0.5 ($K=1$)}} & \ourscell{\ptNo} & \ourscell{\textbf{70.0}} & \ourscell{\textbf{0.700}} & \ourscell{\textbf{40.0}} & \ourscell{\textbf{0.550}} & \ourscell{\textbf{50.0}} & \ourscell{\textbf{0.817}} & \ourscell{\textbf{53.3}} & \ourscell{\textbf{0.689}} \\
\cmidrule(lr){2-11}
& $\pi_{0.5}$ & \ptYes & \textbf{90.0} & \textbf{0.900} & 40.0 & 0.500 & \textbf{60.0} & 0.900 & 63.3 & 0.767 \\
& \ourscell{\textbf{E0.5 ($K=1$)}} & \ourscell{\ptYes} & \ourscell{80.0} & \ourscell{0.800} & \ourscell{\textbf{80.0}} & \ourscell{\textbf{0.850}} & \ourscell{\textbf{60.0}} & \ourscell{\textbf{0.917}} & \ourscell{\textbf{73.3}} & \ourscell{\textbf{0.856}} \\
\midrule
\multicolumn{11}{c}{\textbf{Zerith H1}} \\
\midrule
\multirow{4}{*}{\textbf{ID}}
& $\pi_{0.5}$ & \ptNo & \textbf{100.0} & \textbf{1.000} & 40.0 & 0.450 & 70.0 & 0.867 & 70.0 & 0.772 \\
& \ourscell{\textbf{E0.5 ($K=1$)}} & \ourscell{\ptNo} & \ourscell{\textbf{100.0}} & \ourscell{\textbf{1.000}} & \ourscell{\textbf{80.0}} & \ourscell{\textbf{0.825}} & \ourscell{\textbf{100.0}} & \ourscell{\textbf{1.000}} & \ourscell{\textbf{93.3}} & \ourscell{\textbf{0.942}} \\
\cmidrule(lr){2-11}
& $\pi_{0.5}$ & \ptYes & \textbf{100.0} & \textbf{1.000} & 50.0 & 0.600 & 80.0 & 0.867 & 76.7 & 0.822 \\
& \ourscell{\textbf{E0.5 ($K=1$)}} & \ourscell{\ptYes} & \ourscell{\textbf{100.0}} & \ourscell{\textbf{1.000}} & \ourscell{\textbf{100.0}} & \ourscell{\textbf{1.000}} & \ourscell{\textbf{100.0}} & \ourscell{\textbf{1.000}} & \ourscell{\textbf{100.0}} & \ourscell{\textbf{1.000}} \\
\midrule
\multirow{4}{*}{\textbf{OOD}}
& $\pi_{0.5}$ & \ptNo & 60.0 & 0.600 & 20.0 & 0.375 & 20.0 & 0.517 & 33.3 & 0.497 \\
& \ourscell{\textbf{E0.5 ($K=1$)}} & \ourscell{\ptNo} & \ourscell{\textbf{100.0}} & \ourscell{\textbf{1.000}} & \ourscell{\textbf{40.0}} & \ourscell{\textbf{0.550}} & \ourscell{\textbf{70.0}} & \ourscell{\textbf{0.917}} & \ourscell{\textbf{70.0}} & \ourscell{\textbf{0.822}} \\
\cmidrule(lr){2-11}
& $\pi_{0.5}$ & \ptYes & 90.0 & 0.900 & 30.0 & 0.425 & 20.0 & 0.700 & 46.7 & 0.675 \\
& \ourscell{\textbf{E0.5 ($K=1$)}} & \ourscell{\ptYes} & \ourscell{\textbf{100.0}} & \ourscell{\textbf{1.000}} & \ourscell{\textbf{50.0}} & \ourscell{\textbf{0.625}} & \ourscell{\textbf{80.0}} & \ourscell{\textbf{0.950}} & \ourscell{\textbf{76.7}} & \ourscell{\textbf{0.858}} \\
\bottomrule
\end{tabular}
\end{table}

\begin{figure}
    \centering
    \includegraphics[width=\textwidth]{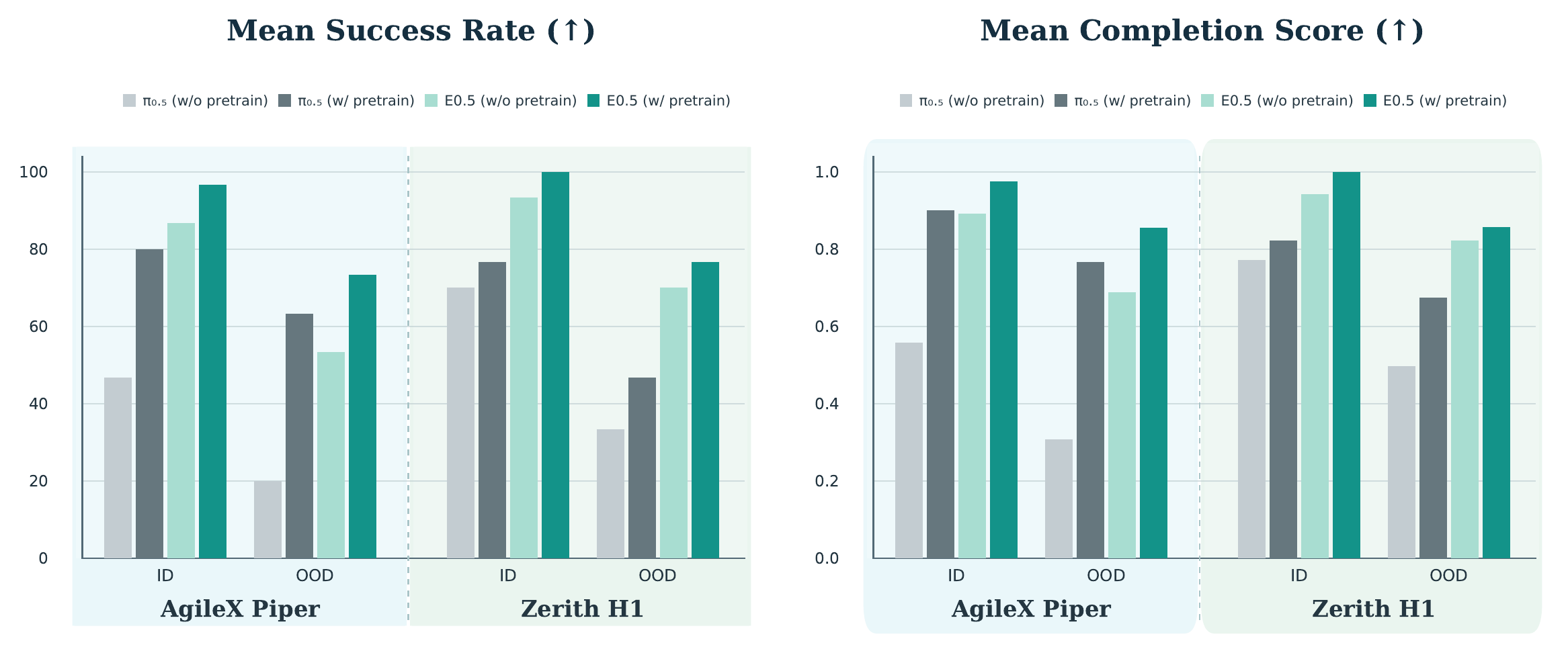}
    \caption{%
        \textbf{Effect of AgiBotWorld-Beta embodied pretraining on single-backbone ($K=1$) E0.5 across AgileX Piper and Zerith H1.}
        Mean success rate (SR, \%) and completion score (CS) are averaged across mouse placement, charger packing, and trash disposal under ID and OOD settings, comparing $\pi_{0.5}$ and E0.5 ($K=1$) with (w/) and without (w/o) large-scale robot-data pretraining.%
    }
    \label{fig:real_robot_performance}
\end{figure}

As shown in \cref{tab:real_robot_pretrain_results} and \cref{fig:real_robot_performance}, large-scale embodied pretraining consistently improves the single-backbone E0.5 policy across both robot embodiments: averaged over ID and OOD settings, mean SR increases from 70.0\% to 85.0\% (+15.0 points) on AgileX Piper and from 81.7\% to 88.3\% (+6.7 points) on Zerith H1. Gains are especially pronounced under OOD perturbations, rising from 53.3\% to 73.3\% on Piper (including doubling unfamiliar charger packing from 40.0\% to 80.0\%) and from 70.0\% to 76.7\% on H1. These results demonstrate that native geometric foundations benefit orthogonally from both architectural capacity scaling ($K=2$ MoG) and large-scale cross-embodiment robot pretraining.

\section{\teal{Con}\graytext{clusion}}
\label{sec:conclusion}

We presented E0.5, a world-action model that jointly learns action generation and future geometry prediction on a native 3D foundation backbone. By supervising geometry at the endpoint of each action chunk, E0.5 adapts pretrained geometric representations to anticipate spatial changes over the action horizon. Mixture-of-Geometry further expands this predictive capacity through noise-conditioned routing across geometric backbones, activating one backbone at each denoising stage. Experiments on XPolicyLab RoboTwin, LIBERO and its generalization benchmarks, and two real-world robot platforms demonstrate strong manipulation performance under changes in scene configurations, appearance, and object properties---achieving 89.30\% Clean and 79.90\% Randomized success on RoboTwin and outperforming embodied-pretrained $\pi_{0.5}$ on real robots without prior embodied-data pretraining. Controlled ablations confirm that future geometry prediction improves generalization over current-scene reconstruction, while a two-backbone mixture provides additional gains; complementary pretraining of the single-backbone variant on AgiBotWorld-Beta further demonstrates effective scaling with cross-embodiment robot trajectories. Together, these findings support future-predictive geometric representations as an effective foundation for world-action modeling and motivate further investigation into how geometric capacity and training resources should scale together.

\FloatBarrier
\section{\teal{Au}\graytext{thors}}
\label{sec:authors}

\textbf{Core Contributors:} Jianhao Lai$^*$, Zhihao Zhan$^*$, Zhiquan Liao$^*$, Qichang Li, Shubin Chen, Zhiyuan Liu, Hongsen Lei, Jianman Lin, Zhihua Xu, Xiaoxin Lin, Tianshui Chen$^{\dagger}$, Liang Lin$^{\dagger}$, Guangrun Wang$^{\dagger}$

\noindent
$^*$Co-First Authors (laijh39@mail2.sysu.edu.cn; zhanzhh6@mail2.sysu.edu.cn; liaozhiquan@x-era.ai).

\noindent
$^\dagger$Corresponding Authors (wanggrun@gmail.com; linliang@ieee.org; tianshuichen@gmail.com).

\noindent
\textbf{Contributors:} Ke Gong, Baiyue Gu, Shiming Chen, Junlin He, Longqi Yang, Junrong Xiao, Haolin Luo, Xiaojuan Su, Jianfeng Li, Yu Luo, Xu Sun, Likui Zhang, Jusheng Zhang, Qinhan Lyu

\bibliography{references}

\appendix

\section{\teal{Training Protocols} \graytext{and Augmentation Analyses}}
\label{app:training_configurations}

This appendix summarizes the complete training configurations across all simulation and real-world benchmarks (\cref{app:complete_training_configurations}), and then disentangles the role of visual data augmentation from our core architectural contributions through unaugmented benchmark evaluations (\cref{sec:unaug_results}), matched-augmentation comparisons with FastWAM (\cref{sec:matched_aug_fastwam}), and augmented component ablations on XPolicyLab RoboTwin (\cref{sec:aug_ablations}) to directly bridge \cref{tab:robotwin_c2r,tab:ablation_future}.

\subsection{\teal{Training} \graytext{Configurations}}
\label{app:complete_training_configurations}

\Cref{tab:appendix_training_configs} lists the training settings.
Steps denote optimizer updates, and the effective batch size equals
batch size per GPU $\times$ number of GPUs $\times$ gradient accumulation.
Simulation models initialize the geometric backbones from VGGT-Omega
and the action decoder randomly, jointly train the policy modules,
and keep the T5 language encoder frozen.

\begin{table}[t]
    \centering
    \caption{
    Training configurations. Image sizes are height $\times$ width
    per view. Real-robot recording durations are approximate.
    Common optimization settings apply to all five columns.
    }
    \label{tab:appendix_training_configs}
    \setlength{\tabcolsep}{4pt}
    \renewcommand{\arraystretch}{1.1}
    \resizebox{\textwidth}{!}{%
    \begin{tabular}{lccccc}
        \toprule
        Configuration
        & RoboTwin main & RoboTwin ablations & LIBERO
        & AgileX Piper & Zerith H1 \\
        \midrule
        Training tasks
        & 50 & 50 & 40 & 3 & 3 \\
        Demonstrations
        & 2,500 & 2,500 & 1,777 & 604 & 610 \\
        Frames
        & 549,787 & 549,787 & 289,017
        & 318,088 & 368,017 \\
        Recording duration
        & -- & -- & --
        & $\approx2.95$ h & $\approx3.41$ h \\
        Epochs
        & $15+15$ & 15 & 10 & 5 & 5 \\
        Optimizer updates
        & $64{,}440+64{,}440$ & 64,440 & 22,580
        & 12,427 & 14,377 \\
        GPUs
        & 8 & 8 & 8 & 8 & 8 \\
        Batch size per GPU
        & 16 & 16 & 16 & 8 & 8 \\
        Gradient accumulation
        & 1 & 1 & 1 & 2 & 2 \\
        Effective batch size
        & 128 & 128 & 128 & 128 & 128 \\
        \midrule
        Optimizer
        & \multicolumn{5}{c}{
        AdamW, $(\beta_1,\beta_2)=(0.9,0.95)$,
        weight decay $0.01$} \\
        Learning-rate schedule
        & \multicolumn{5}{c}{
        5\% linear warm-up to $10^{-4}$,
        followed by cosine decay to $10^{-6}$} \\
        Precision / gradient clipping
        & \multicolumn{5}{c}{
        BF16 / maximum gradient norm $1.0$} \\
        \midrule
        Camera views
        & Head + two wrists & Head + two wrists
        & External + wrist
        & Top + two wrists & Top + two wrists \\
        Backbone input size
        & $224\times288$ & $224\times288$ & $256\times256$
        & $224\times288$ & $224\times288$ \\
        Proprioceptive dimension
        & 14 & 14 & 8 & 14 & 23 \\
        Mixup / RandPhoto
        & On / On & Off / Off & Off / Off & Off / Off & Off / Off \\
        Predicted action length
        & 32 & 32 & 10 & 32 & 32 \\
        Action normalization
        & \multicolumn{5}{c}{Per-dimension z-score normalization} \\
        Geometry supervision
        & Future 3D & None / current / future 3D & Future 3D
        & Future depth & Future depth \\
        Future target offset
        & 32 & 32 when enabled & 10 & 32 & 32 \\
        Outer geometry weight $\lambda_g$
        & 0.01 & 0 or 0.01 & 0.01 & 0.01 & 0.01 \\
        Denoising steps
        & 10 & 10 & 10 & 10 & 10 \\
        \bottomrule
    \end{tabular}%
    }
\end{table}

\paragraph{Data and inputs.}
RoboTwin uses 50 clean demonstrations per task, without randomized
training scenes. LIBERO jointly trains on Spatial, Object, Goal,
and Long, retaining 450, 464, 459, and 404 successful simulator
replays, respectively, after filtering. All retained simulation
trajectories are used for training. Each policy receives only the
current RGB frame from each camera, together with proprioception
and the instruction. Image preprocessing preserves aspect ratio
with patch-aligned dimensions; the reference resolution of 256
therefore yields $224\times288$ inputs for RoboTwin.
Real-robot training covers three tasks: mouse, trash, and charging cable.
Across these three tasks, AgileX Piper uses 604 demonstrations
(318,088 frames), and Zerith H1 uses 610 demonstrations
(368,017 frames).
Real-robot images are recorded at $480\times640$ and 30 fps;
the predicted actions are absolute joint targets spanning
approximately 1.07 seconds.

\paragraph{Additional augmentation.}
The RoboTwin main recipe applies RandPhoto followed by RGB mixup.
RandPhoto uses \texttt{rand-m5-n2-p0.5}, with random choices shared
across views. Mixup forms
$\widetilde O_i=(1-\alpha_i)O_i+\alpha_iO_{\pi(i)}$,
where $\alpha_i\sim\mathcal U(0,0.4)$ is shared across cameras and
$\pi$ cyclically shifts samples within each GPU's microbatch.
The recipient's actions, instruction, proprioception, and geometry
targets remain unchanged. Neither augmentation is used during evaluation. No data augmentation was applied to LIBERO and real robots.

\paragraph{Geometry supervision and inference.}
Simulation uses the VGGT-Omega camera, depth, and point losses,
each with internal weight 5, followed by the outer coefficient
$\lambda_g$. For the two-backbone model, the two routed geometry
losses are averaged. Real-robot training instead supervises only future depth: its effective depth coefficient is $0.01 \times 5 = 0.05$, with camera and point losses disabled.
Future targets use offset 32 on RoboTwin and real robots, and
offset 10 on LIBERO, matching its predicted action length.
RoboTwin executes 24 predicted actions before replanning,
while standard LIBERO executes 10. Future geometry targets
are used only for training.

\paragraph{Controlled ablation protocol.}
The ablations in \cref{sec:ablation} assess the effects of geometric supervision and backbone count under a common training protocol. All ablation variants use the same single 15-epoch schedule (64,440 updates), with mixup and RandPhoto disabled. Within each ablation comparison, only the factor under study is varied, while the remaining training and evaluation settings are held fixed. This shared protocol allows component effects to be assessed without confounding them with differences in training duration or augmentation across variants. The main RoboTwin experiment uses a longer, augmented training recipe: two consecutive 15-epoch stages with both mixup and RandPhoto, totaling 128,880 updates. The second stage loads the first stage's final model weights and restarts the optimizer and learning-rate schedule. Training data, effective batch size, optimizer settings, visual inputs, action representation, and inference settings are otherwise shared with the ablations. Component-level conclusions are therefore based on comparisons within the controlled ablation protocol, rather than on performance differences between the main and ablation recipes. To directly bridge these two training regimes and ensure fair comparisons under identical visual augmentation, \cref{sec:matched_aug_fastwam,sec:aug_ablations} further evaluate a representative video-based WAM baseline and our component ablations under the augmented training recipe.

\subsection{\teal{Strong Performance} \graytext{and Generalization without Data Augmentation}}
\label{sec:unaug_results}

This section examines E0.5's performance and generalization without data augmentation, complementing the augmented RoboTwin results in \cref{sec:sim_experiments} (\cref{tab:robotwin_c2r}). We bring together unaugmented results across all evaluated benchmarks to assess the model's capabilities beyond the main RoboTwin training recipe. The results demonstrate strong manipulation performance and generalization across simulation and real-world settings, supporting the paper's conclusions under unaugmented training.

\begin{table}[t]
    \centering
    \caption{\textbf{Performance and generalization of E0.5 without data augmentation.} Panel (a) compares simulation success rates with VLA and WAM baselines, including geometry-enhanced models. Panel (b) reports mean real-world success across three tasks under EASY and HARD settings. All E0.5 results use no data augmentation or embodied-data pretraining. Baseline results are reproduced from \cref{tab:robotwin_c2r,tab:libero_all,tab:h1_real_world}. The unaugmented RoboTwin model is trained for 70 epochs; all other E0.5 results are identical to those reported in \cref{sec:experiment}. All values are success rates (\%), and bold indicates the best result in each column.}
    \label{tab:performance_without_augmentation}
    \begingroup
    \small
    \setlength{\tabcolsep}{6pt}
    \renewcommand{\arraystretch}{1.12}
    \begin{tabular}{lrrrrrr}
        \toprule
        \multicolumn{7}{c}{\textbf{(a) Simulation benchmarks}} \\
        \midrule
        & & & & \multicolumn{3}{c}{RoboTwin} \\
        \cmidrule(lr){5-7}
        Method & LIBERO & LIBERO-Plus & LIBERO-PRO & Clean & Random & Avg. \\
        \midrule
        $\pi_{0.5}$      & 96.85 & 84.40 & 53.19 & 70.66 & 46.00 & 58.33 \\
        Spatial Forcing & 98.50 & 71.10 & 42.42 & 77.18 & 26.74 & 51.96 \\
        FastWAM         & 97.60 & 52.56 & 40.51 & 77.76 &  1.92 & 39.84 \\
        X-WAM           & 96.60 & 74.80 & 53.13 & 69.98 & 25.84 & 47.91 \\
        4D-WAM          & 98.60 & 71.01 & 48.34 & 81.50 & 41.80 & 61.65 \\
        \midrule
        \textbf{E0.5} & \textbf{98.90} & \textbf{88.78} & \textbf{55.36} & \textbf{89.52} & \textbf{46.58} & \textbf{68.05} \\
        \bottomrule
    \end{tabular}

    \vspace{0.7em}

    \begin{tabular}{lrrrr}
        \toprule
        \multicolumn{5}{c}{\textbf{(b) Real-world benchmarks}} \\
        \midrule
        & \multicolumn{2}{c}{Zerith H1} & \multicolumn{2}{c}{AgileX Piper} \\
        \cmidrule(lr){2-3} \cmidrule(lr){4-5}
        Method & EASY & HARD & EASY & HARD \\
        \midrule
        $\pi_{0.5}$          & 76.7 & 46.7 & 80.0 & 63.3 \\
        $\pi_{0.5}$ w/o pt.  & 70.0 & 33.3 & 46.7 & 20.0 \\
        \midrule
        \textbf{E0.5} & \textbf{100.0} & \textbf{73.3} & \textbf{86.7} & \textbf{66.7} \\
        \bottomrule
    \end{tabular}
    \endgroup
\end{table}

\paragraph{Strong performance across benchmarks.}
\Cref{tab:performance_without_augmentation} summarizes E0.5's performance without data augmentation across RoboTwin, LIBERO, LIBERO-Plus, LIBERO-PRO, and two real-world robot platforms. E0.5 uses pretrained geometric and language models without embodied-data pretraining. The comparisons cover VLA and WAM baselines, including geometry-enhanced models such as Spatial Forcing, X-WAM, and 4D-WAM. E0.5 achieves the highest success rates in all aggregate metrics reported in the table.

\paragraph{Generalization from unaugmented demonstrations.}
On LIBERO, E0.5 achieves 98.90\% success. Directly evaluating the same checkpoint without further fine-tuning yields 88.78\% on LIBERO-Plus and 55.36\% on LIBERO-PRO. These results exceed $\pi_{0.5}$ by 4.38 and 2.17 percentage points on the two robustness benchmarks, respectively. Relative to the geometry-enhanced WAM 4D-WAM, the corresponding improvements are 17.77 and 7.02 points. This consistent performance across standard tasks and robustness evaluations demonstrates generalization from unaugmented task demonstrations.

\paragraph{Bimanual and real-world manipulation.}
On RoboTwin, unaugmented E0.5 achieves 89.52\% Clean and 46.58\% Random success, yielding a 68.05\% average and exceeding all five compared baselines. On real robots, E0.5 outperforms both $\pi_{0.5}$ and $\pi_{0.5}$ without embodied-data pretraining under EASY and HARD settings on both platforms. Under HARD, it achieves 73.3\% success on Zerith H1 and 66.7\% on AgileX Piper. These results extend the benefits of the proposed model to bimanual manipulation and real-world object and configuration shifts.

\paragraph{Training details.}
The unaugmented RoboTwin result is obtained from a separate 70-epoch run with mixup and RandPhoto disabled. All other E0.5 results in \cref{tab:performance_without_augmentation} match those reported in \cref{sec:experiment}. The controlled component ablations use a separate 15-epoch protocol with both augmentations disabled.

\subsection{\teal{Matched-Augmentation Comparison} \graytext{with Video-Based WAM}}
\label{sec:matched_aug_fastwam}

In \cref{tab:robotwin_c2r}, video-centered WAM baselines without embodied pretraining, such as FastWAM, exhibit severe degradation on RoboTwin Randomized scenes (1.92\% success), where background textures, lighting, and tabletop clutter differ substantially from the clean training demonstrations. Because the main RoboTwin recipe for E0.5 incorporates photometric augmentation (RandPhoto) and cross-sample RGB mixup (\cref{app:complete_training_configurations}), a natural question is how much of the clean-to-random performance gap stems from visual augmentation versus the underlying world-action modeling backbone.

\begin{table}[t]
\centering
\caption{\textbf{Controlled comparison with FastWAM on RoboTwin across unaugmented and matched-augmentation settings (success rate, \%).} Both methods use clean-only demonstrations without embodied-data pretraining. Under both regimes, E0.5 substantially outperforms the video-based WAM baseline.}
\label{tab:fastwam_matched_aug}
\small
\setlength{\tabcolsep}{5.2pt}
\begin{tabular}{llccc}
\toprule
Augmentation & Method & Clean & Random & Avg. \\
\midrule
\multirow{2}{*}{Off}
 & FastWAM & 77.76 & 1.92 & 39.84 \\
 & E0.5 (Ours) & \textbf{89.52} & \textbf{46.58} & \textbf{68.05} \\
\midrule
\multirow{2}{*}{RandPhoto + Mixup}
 & FastWAM & 80.58 & 48.38 & 64.48 \\
 & E0.5 (Ours) & \textbf{89.30} & \textbf{79.90} & \textbf{84.60} \\
\bottomrule
\end{tabular}
\end{table}

To provide a strictly matched comparison, we train FastWAM on the same 50 clean demonstrations per task with the exact same RandPhoto (\texttt{rand-m5-n2-p0.5}) and RGB mixup ($\alpha \sim \mathcal{U}(0, 0.4)$) augmentations applied during training, and evaluate it under the identical 50-task protocol. \Cref{tab:fastwam_matched_aug} compares both methods with and without augmentation:
\begin{itemize}
    \item \textbf{Effect of augmentation on video-based WAM.} Adding RandPhoto and mixup substantially mitigates appearance overfitting in FastWAM, lifting its Randomized success from 1.92\% to 48.38\% (+46.46 percentage points) and its average success from 39.84\% to 64.48\%.
    \item \textbf{Consistent advantage of the native 3D foundation.} Under identical data augmentation, E0.5 still outperforms FastWAM by \textbf{8.72} points on Clean (89.30\% vs.\ 80.58\%), \textbf{31.52} percentage points on Randomized (79.90\% vs.\ 48.38\%), and \textbf{20.12} points in average success (84.60\% vs.\ 64.48\%). Combined with E0.5's +44.66-point Randomized advantage in the unaugmented regime (46.58\% vs.\ 1.92\%), these results confirm that building world-action models on a predictive 3D geometric backbone provides substantial spatial generalization gains beyond image-level appearance augmentation.
\end{itemize}

\subsection{\teal{Component Ablations} \graytext{under Data Augmentation}}
\label{sec:aug_ablations}

\Cref{tab:ablation_future} evaluates geometric supervision targets and backbone capacity under an unaugmented 15-epoch schedule to isolate architectural effects without external regularizers. To verify whether future 3D geometry prediction and Mixture-of-Geometry (MoG) remain effective when strong visual augmentation is enabled---and to bridge the numerical gap between \cref{tab:ablation_future,tab:robotwin_c2r}---we repeat the core 15-epoch ablations (64,440 optimizer updates, effective batch size 128) with RandPhoto and RGB mixup enabled.

\begin{table}[t]
\centering
\caption{\textbf{Component ablations on RoboTwin under data augmentation (success rate, \%).} The upper block evaluates 15-epoch variants with RandPhoto and RGB mixup enabled, matching all other settings of \cref{tab:ablation_future}. The bottom row shows the two-stage ($15+15$ epochs) main recipe from \cref{tab:robotwin_c2r} for reference.}
\label{tab:aug_ablation}
\small
\setlength{\tabcolsep}{5.0pt}
\begin{tabular}{clcccc}
\toprule
$K$ & Target & Epochs & Clean & Random & Avg. \\
\midrule
1 & Current 3D & 15 & 83.04 & 68.08 & 75.56 \\
1 & Future 3D & 15 & 85.82 & 71.14 & 78.48 \\
2 & Future 3D & 15 & \textbf{86.34} & \textbf{74.52} & \textbf{80.43} \\
\midrule
\multicolumn{6}{l}{\emph{Full two-stage training recipe (\cref{tab:robotwin_c2r})}} \\
2 & Future 3D & $15+15$ & 89.30 & 79.90 & 84.60 \\
\bottomrule
\end{tabular}
\end{table}

\Cref{tab:aug_ablation} yields three takeaways:
\begin{itemize}
    \item \textbf{Orthogonality of predictive 3D modeling and visual augmentation.} Even atop a strong augmented baseline, replacing current-scene reconstruction with action-horizon-aligned future 3D prediction at $K=1$ consistently improves Clean (+2.78 points), Randomized (+3.06 points), and average success (+2.92 points, from 75.56\% to 78.48\%). Expanding capacity from $K=1$ to $K=2$ with MoG further raises Clean to 86.34\%, Randomized to 74.52\% (+3.38 points over $K=1$ Future 3D), and average success to 80.43\% (+4.87 points over $K=1$ Current 3D). Notably, even the 15-epoch single-backbone Current 3D variant (68.08\% Random) already exceeds the matched-augmentation FastWAM baseline (48.38\% Random) by 19.70 percentage points.
    \item \textbf{Interaction between appearance invariance and geometric targets.} Comparing unaugmented ablations (\cref{tab:ablation_future}) with \cref{tab:aug_ablation} clarifies how the two mechanisms interact. Without augmentation, Current 3D reconstruction remains susceptible to clean-background appearance overfitting (16.58\% Random), whereas predicting Future 3D states forces the backbone to attend to state transitions and acts partly as an implicit regularizer (+12.81 Random points). Once RandPhoto and mixup explicitly suppress low-level appearance overfitting, the pretrained 3D backbone's spatial invariance is unlocked even under Current 3D supervision (68.08\% Random). Across 50 bimanual tasks (5,000 evaluation trials per condition), future geometry supervision and MoG continue to deliver monotonic improvements across all metrics.
    \item \textbf{Seamless progression from ablation to main results.} Together, \cref{tab:ablation_future,tab:aug_ablation} fully account for the progression from the unaugmented 15-epoch ablation ($K=2$, Future 3D: 35.03\% Random, 60.52\% Avg.) to the main result: enabling RandPhoto and mixup under the identical 15-epoch budget increases performance to 74.52\% Random and 80.43\% Avg., and completing the second 15-epoch stage accounts for the remaining convergence gain to 79.90\% Random and 84.60\% Avg.
\end{itemize}

\section{\teal{Comparison} \graytext{with VGA}}
\label{sec:supp_compare_with_vga}

\paragraph{Comparison with VGA.}
VGA is a closely related approach that also builds
robotic policies upon a native 3D foundation backbone, but focuses
on current-scene geometric reconstruction with LoRA adaptation.
In contrast, E0.5 jointly optimizes its geometric backbone for
future 3D prediction and action generation.
Unlike VGA, which trains separate policies for the four LIBERO
task suites, E0.5 jointly trains a single unified policy across
all 40 tasks.
Despite this unified training setting, E0.5 achieves a higher
overall success rate (98.90\% vs.\ 98.10\%), with a particularly
notable improvement on LIBERO-Long (98.20\% vs.\ 95.00\%).

On LIBERO-Plus, the official VGA repository additionally reports
results across all seven perturbation categories, enabling a more
comprehensive comparison, as shown in \cref{tab:vga_comparison}.
E0.5 reports higher success rates in six of the seven categories,
with particularly substantial improvements under robot initial-state
variations (83.74\% vs.\ 40.10\%), background texture changes
(95.54\% vs.\ 68.70\%), and object layout perturbations
(85.57\% vs.\ 73.20\%).
E0.5 also achieves higher reported success rates under language
variations (88.03\% vs.\ 80.00\%) and lighting changes
(98.51\% vs.\ 90.90\%), while performing slightly below VGA
under camera viewpoint shifts (84.37\% vs.\ 86.30\%).
Notably, VGA employs separate suite-specific checkpoints and
a dedicated LIBERO-Plus training configuration, whereas E0.5
directly evaluates the same unified LIBERO-trained checkpoint
without additional fine-tuning.
Although the methods differ in their architectures and evaluation
aggregation, these results highlight the effectiveness of E0.5
as a unified, future-predictive geometric world-action model,
demonstrating strong cross-task performance and robustness
to diverse distribution shifts.

\begin{table}[t]
\centering
\caption{
Comparison with VGA on LIBERO and LIBERO-Plus.
(a) Standard LIBERO success rates across four task suites.
VGA trains separate suite-specific policies, whereas E0.5
jointly trains a single policy across all 40 tasks.
(b) LIBERO-Plus success rates across seven perturbation
categories. VGA results are obtained from its official
repository, while E0.5 directly evaluates its LIBERO-trained
checkpoint without additional fine-tuning. All values are percentages.
}
\label{tab:vga_comparison}

\begin{minipage}[t]{0.38\textwidth}
\centering
\textbf{(a) Standard LIBERO}
\vspace{4pt}

\begin{tabular}{lcc}
\toprule
Task Suite & VGA & E0.5 \\
\midrule
Spatial & 99.00 & \textbf{99.40} \\
Object  & \textbf{99.60} & 99.40 \\
Goal    & 98.60 & 98.60 \\
Long    & 95.00 & \textbf{98.20} \\
\midrule
\textbf{Overall}
& 98.10 & \textbf{98.90} \\
\bottomrule
\end{tabular}
\end{minipage}
\hfill
\begin{minipage}[t]{0.59\textwidth}
\centering
\textbf{(b) LIBERO-Plus}
\vspace{4pt}

\begin{tabular}{lcc}
\toprule
Perturbation & VGA & E0.5 \\
\midrule
Camera Viewpoints
& \textbf{86.30} & 84.37 \\
Robot Initial States
& 40.10 & \textbf{83.74} \\
Language Instructions
& 80.00 & \textbf{88.03} \\
Light Conditions
& 90.90 & \textbf{98.51} \\
Background Textures
& 68.70 & \textbf{95.54} \\
Sensor Noise
& 89.80 & \textbf{90.38} \\
Object Layout
& 73.20 & \textbf{85.57} \\
\bottomrule
\end{tabular}
\end{minipage}

\vspace{-2mm}
\end{table}

\section{\teal{Architecture and Training/Inference} \graytext{Procedures}}
\label{sec:supp_architecture}

As shown in \cref{tab:supp_adapters} (\cref{sec:ablation}), placing three uncoupled ActionDiT blocks before and after the 24 geometry-coupled blocks provides our default geometry--action interface across all configurations. Below we detail the branch-level connectivity, training loss reduction, per-expert sample exposure, and K/V-cached inference procedure.

\subsection{\teal{Information Flow} \graytext{and Attention Connectivity}}
\label{sec:supp_connectivity}

\paragraph{Branch inputs.}
The geometric branch receives current multi-view RGB images $O_t$ and
projected frozen-T5 features of instruction $\ell$. Each view contains
image patches, a camera token, and register tokens. The action branch
receives the noisy chunk $U_\sigma$, a noise embedding of $1000\sigma$,
and T5 features concatenated with an encoded robot-state token $s_t$.
Language/state cross-attention and noise modulation operate in all
30 action blocks. State and noisy actions do not enter the geometric
branch; $\sigma$ only selects its expert. Future geometry is used only
as supervision.

\paragraph{Attention connections and masks.}
Frame attention is visual-only and restricted to individual views.
Inter-frame attention is bidirectional over geometric tokens from all
views and one appended language sequence, with text padding masked.
At register-only groups 3, 7, 10, 15, and 21 (one-based), only
camera/register tokens interact with language; patch tokens retain
their frame-attention outputs.
At each coupled action block, action queries attend jointly to the
whole action chunk and the selected backbone's projected geometric
K/V, without a causal mask. The geometric K/V are captured before the
corresponding inter-frame update and exclude language tokens; register-only
groups expose only camera/register K/V. Direct language/state access
uses a separate cross-attention operation. These connections specify
the masks in \cref{fig:e05_architecture}(c): geometric and backbone-language tokens never read
action tokens. Action information does not flow back into the backbone
during the forward pass, but action-loss gradients propagate through
the undetached K/V (see \cref{sec:policy_aligned_wam} for the probabilistic formulation and gradient analysis).

\paragraph{Expert parameters and geometric readout.}
In the capacity ablation, geometric transformers, image encoders,
camera/register tokens, geometry heads, and bridge projections are
independent across experts. The action decoder and geometry-language
projector are shared. The dense head reads groups 5, 12, 18, and 24;
the camera head reads the final group. Point maps are obtained by
unprojecting predicted depth with predicted cameras.

\subsection{\teal{Training} \graytext{Procedure}}
\label{sec:supp_training}

\paragraph{Routing and supervision.}
Following the notation in \cref{sec:method}, we independently sample
$u\sim\mathcal{U}(0,1)$ and $\epsilon\sim\mathcal{N}(0,I)$ per example.
The shift is $s=5$, with $f_5(u)=5u/(1+4u)$ and $\sigma=f_5(u)$.
Experts are indexed from high to low noise:
\begin{equation}
  r(\sigma)=K-\sum_{i=1}^{K-1}\mathbf{1}[\sigma\geq f_5(i/K)],
  \qquad \mathcal{E}_t=\{r(\sigma)\}.
  \label{eq:supp_router}
\end{equation}
Thus, each example trains one expert, with boundary ties assigned to
higher noise. All experts receive language and predict geometry at the
same action-chunk endpoint. The single- and two-backbone configurations
are the $K=1$ and $K=2$ cases. \Cref{tab:supp_exposure} lists the
boundaries and ideal selection probabilities; finite-precision
timesteps can introduce small deviations.

\paragraph{Loss and optimization.}
We construct $U_\sigma=(1-\sigma)A_t+\sigma\epsilon$ and
predict $\epsilon-A_t$.
The joint objective follows \cref{eq:joint_world_action_objective}.
We specify here the timestep weighting and minibatch reduction
used in training: the action term applies a scheduler weight
to the velocity regression objective in \cref{eq:flow_matching_loss},
and the geometry term uses a per-expert reduction of the
future supervision in \cref{eq:mixture_future_loss}.
For a local minibatch $\mathcal{B}$, the implemented loss is
\begin{equation}
  \mathcal{L}=
  \frac{1}{|\mathcal{B}|}\sum_{n\in\mathcal{B}}
       w(\sigma_n)\ell_{\mathrm{FM},n}
  +\frac{0.01}{K}\sum_{k=1}^{K}
       \mathcal{G}(\mathcal{B}_k^+).
  \label{eq:supp_batch_loss}
\end{equation}
where $\ell_{\mathrm{FM},n}$ is the masked velocity MSE,
$\mathcal{B}_k^+$ contains examples routed to expert $k$
with valid future targets, and $\mathcal{G}$ is the official
VGGT-Omega depth/point/camera criterion evaluated on that
subset, with the internal weights specified in \cref{app:complete_training_configurations}.

The scheduler weight, not explicitly shown in \cref{eq:flow_matching_loss}, is
$w(\sigma)\propto e^{-2(\sigma-1/2)^2}-e^{-1/2}$,
normalized to unit mean on the shifted 1000-step reference
grid, without additional route-specific normalization.
For geometry, each example supervises only its selected
expert, i.e., $\mathcal{E}_t=\{r(\sigma)\}$ in \cref{eq:mixture_future_loss}.
Under equal-probability routing, that population objective
decomposes into an equally weighted average of the
expert-conditional expected losses.
Our minibatch implementation averages the subset criteria
with a fixed factor $1/K$, rather than weighting them by
the realized subset sizes.
Empty subsets contribute zero while the denominator remains
$K$.
Padded action positions and episode-end future targets are
excluded from their respective losses.

Geometric backbones and heads use pretrained VGGT-Omega weights;
the action expert and added modules are randomly initialized.
Action loss trains the shared decoder and conditioning modules, selected
bridge/backbone, and geometry-language projector. Geometry loss also
trains the selected heads. T5 is frozen and routing is parameter-free.
We use AdamW with learning rate $10^{-4}$, weight decay $0.01$, and
cosine decay. \Cref{alg:supp_training} summarizes training.

\begin{algorithm}[t]
  \caption{Joint action and future-geometry training}
  \label{alg:supp_training}
  \small
  \begin{algorithmic}[1]
    \Require Experts $\{B_k,Q_k,\Pi_k\}_{k=1}^{K}$; shared decoder $v$
    \For{each training iteration}
      \State Sample minibatch $(O_n,\ell_n,s_n,A_n,G_n^+)$
      \State Obtain frozen T5 features $L_n$ and padding masks
      \State Sample $u_n,\epsilon_n$; set $\sigma_n=f_5(u_n)$,
        $k_n=r(\sigma_n)$
      \State $U_n\gets(1-\sigma_n)A_n+\sigma_n\epsilon_n$
      \For{each nonempty routed subset $\mathcal{B}_k$}
        \State $h_n,M_n\gets B_k(O_n,L_n)$
          \Comment{Features and K/V}
        \State $\widehat{G}_n^+\gets Q_k(h_n)$
        \State $\widehat{v}_n\gets v(U_n,\sigma_n;\Pi_k M_n,L_n,s_n)$
      \EndFor
      \State Compute \cref{eq:supp_batch_loss}, masking invalid targets
      \State Backpropagate; average gradients across workers
      \State Apply AdamW and update the learning-rate schedule
    \EndFor
  \end{algorithmic}
\end{algorithm}

\paragraph{Per-expert budget and capacity scaling.}
The capacity ablation uses 15 epochs, batch size 16 per GPU on eight
GPUs, and no gradient accumulation: global batch $B=128$ and
$S=64{,}440$ updates. The nominal exposure is $N=BS=8{,}248{,}320$,
including distributed-loader repeats. Each expert receives $N/K$
examples in expectation; valid geometric targets are fewer after
padding exclusion. These are exposure counts, not distinct trajectories
or measured cumulative route counts.

\begin{table}[t]
  \centering
  \small
  \setlength{\tabcolsep}{3pt}
  \caption{\textbf{Routing and expected per-expert exposure.}
  Boundaries are ascending; experts are indexed from high to low noise.
  All configurations use the same 15-epoch budget.}
  \label{tab:supp_exposure}
  \begin{tabular}{c c c r r}
    \toprule
    $K$ & Boundaries & Prob. & Samples/expert & Epoch-equiv. \\
    \midrule
    1 & --- & $1$ & 8,248,320 & 15.00 \\
    2 & $5/6$ & $1/2$ & 4,124,160 & 7.50 \\
    3 & $5/7,10/11$ & $1/3$ & 2,749,440 & 5.00 \\
    4 & $5/8,5/6,15/16$ & $1/4$ & 2,062,080 & 3.75 \\
    \bottomrule
  \end{tabular}
\end{table}

The expected number of updates containing examples for each expert is
$S[1-(1-1/K)^{128}]\approx S$ for $K\leq4$. Increasing $K$ therefore
reduces samples per expert, rather than dividing optimizer updates by $K$.
In \cref{tab:ablation_future}, $K=3$ still improves average success over $K=1$
(59.82\% vs.\ 57.79\%), although it trails $K=2$ (60.52\%).
At $K=4$, average success falls to 48.67\%, with only half the
per-expert exposure of $K=2$. These results characterize the trade-off
between geometric capacity and per-expert training exposure under a fixed
total budget. We adopt $K=2$, which achieves the best Random and average
success in this setting.

\paragraph{Rationale and scope of MoG.}
We adopt MoG as a practical way to expand a geometry-native world-action model using multiple pretrained geometric backbones, each trained with the same future-geometry objective.
Our ablations show that expanding from one backbone to either two or three improves average manipulation success under the evaluated training budget, supporting MoG as an effective capacity-expansion strategy in this setting.
Our contribution is to demonstrate that geometry-native world-action modeling can benefit from this form of capacity expansion.
These results do not establish MoG's superiority over alternative scaling strategies, such as increasing backbone depth or width, or introducing layer-wise experts; such comparisons require dedicated controls for pretraining, training compute, and inference compute.

\subsection{\teal{Inference} \graytext{Procedure}}
\label{sec:supp_inference}

Starting from Gaussian action noise, we use Euler updates along
$\sigma_j=f_5(1-j/J)$, $j=0,\ldots,J$, and select $r(\sigma_j)$ at
each step. The action latent is preserved across expert switches.
On entering a new interval, we run its geometric backbone once and
replace the cached layer-wise K/V; subsequent steps in that interval
reuse the cache. Action-token K/V are recomputed at every step.
This requires one geometric forward per visited expert, at most
$\min(K,J)$ per planning cycle. Routing boundaries remain fixed to the
training partition, including when the inference shift is overridden.

In the RoboTwin ablations, we use ten denoising steps, predict 32 actions,
and execute the first 24 sequentially before observing and replanning
with a fresh cache. Future geometric supervision uses the frame at
offset 32 with stride one. Camera and dense heads are skipped during
ordinary deployment and enabled only for optional geometric diagnostics.

\begin{algorithm}[t]
  \caption{Inference with geometric K/V caching}
  \label{alg:supp_inference}
  \small
  \begin{algorithmic}[1]
    \Require $(O_t,\ell,s_t)$; horizon $H_a$; denoising steps $J$
    \State $L\gets\operatorname{T5}(\ell)$;
      $U\sim\mathcal{N}(0,I)$ in $\mathbb{R}^{H_a\times d_a}$
    \State $k_{\mathrm{cache}}\gets\varnothing$
    \State Set $\sigma_j=f_5(1-j/J)$ for $j=0,\ldots,J$
    \For{$j=0,\ldots,J-1$}
      \State $k\gets r(\sigma_j)$
      \If{$k\neq k_{\mathrm{cache}}$}
        \State Replace $M$ with geometric K/V from $B_k(O_t,L)$
        \State $k_{\mathrm{cache}}\gets k$ \Comment{Skip geometry heads}
      \EndIf
      \State $\widehat{v}\gets v(U,\sigma_j;\Pi_k M,L,s_t)$
      \State $U\gets U+(\sigma_{j+1}-\sigma_j)\widehat{v}$
    \EndFor
    \State Denormalize $U$ and execute the configured action prefix
    \State Observe and replan with an empty cache
  \end{algorithmic}
\end{algorithm}

\section{\teal{Future Geometry Targets} \graytext{and Supervision}}
\label{app:future_geometry}

We detail the temporal alignment, sources, coordinate conventions, and
scale normalization of the geometric targets in \cref{sec:method}. For simulation,
each target comprises multi-view depth, camera parameters, and pointmaps
derived by unprojection. All future targets are training labels; the geometric
backbone receives the current RGB observations and instruction.

\subsection{\teal{Action Intervals} \graytext{and Future-Frame Alignment}}
\label{app:geometry_alignment}

Let $t$ index a recorded observation, and let $a_t$ denote the action
associated with the transition from observation $t$ to $t+1$.
The action chunk is $A_t=(a_t,\ldots,a_{t+H_a-1})$.
The future target corresponds to the observation after completing this
chunk. Writing $\tau(j)$ for the recorded timestamp of frame $j$,
\begin{equation}
  j_f=t+H_a,\qquad t_f=\tau(t+H_a).
  \label{eq:supp_geometry_time}
\end{equation}
Thus, the last action is $a_{t+H_a-1}$, while its resulting observation
is at $t+H_a$. With sampling stride $s$, the corresponding source
frame is $t+sH_a$; the configurations considered here use $s=1$.

\paragraph{Sampling and synchronization.}
RoboTwin uses $H_a=32$. Its converted clean training data have a
15-fps timestamp grid, giving a recorded interval of $32/15\approx2.13$
seconds. This is the dataset time base, not a statement about the
simulator's internal integration frequency. RGB, depth, and calibration
are associated by episode, frame index, and camera identity. In this
dataset, the proprioceptive state is the joint configuration at frame $t$,
while the action stores the joint target at frame $t+1$.
LIBERO replays are recorded at 20 Hz, with RGB, depth, calibration, and
proprioception captured before executing each retained action. No-op
actions are removed before replay, so timestamps refer to the regenerated
trajectory. LIBERO uses $H_a=10$, corresponding to a 0.5-second interval.

\paragraph{Episode boundaries.}
For an episode containing $T$ recorded frames, define
$m_t^+=\mathbf{1}[t+sH_a<T]$. The loader clips an out-of-range index to
$T-1$ for tensor construction, but excludes that sample from the future
geometry loss using $m_t^+$. It never takes a target from the next episode
or treats a repeated terminal frame as a valid future observation.
Action padding has a separate mask: available actions may still
contribute to flow matching when the geometric endpoint is unavailable.
The executed action prefix at deployment does not redefine the training
target offset.

\subsection{\teal{Sources of Simulated} \graytext{and Real-World Geometry}}
\label{app:geometry_sources}

\paragraph{RoboTwin.}
We use paired RGB-D observations from the RoboTwin training demonstrations,
with the head, left-wrist, and right-wrist cameras in that order.
The source depth is simulator-generated optical-axis depth, stored as
16-bit PNGs in millimeters and converted to meters by dividing by 1000.
Zero denotes invalid depth. Camera intrinsics and OpenCV
world-to-camera extrinsics are read from the calibration associated
with the same frame. Thus, future wrist-camera labels use the recorded
future pose, not the pose at the input observation.

\paragraph{LIBERO.}
For the success-filtered replay dataset used by the archived training
configurations, we restore each demonstration's scene and initial state
in MuJoCo and replay its retained actions. At every recorded step,
the external \texttt{agentview} camera and the wrist-mounted
\texttt{robot0\_eye\_in\_hand} camera provide paired RGB and depth.
The simulator depth buffer is converted to metric optical-axis depth.
Intrinsics are computed from the image dimensions and camera field of
view; extrinsics are computed from the camera position and orientation
at that step and converted to the OpenCV convention.
Pixels with nonfinite or nonpositive depth, or depth-buffer values
at the far plane ($b\geq1-10^{-6}$), are invalid.
RGB, geometry, and proprioception come from the same replay state.
The retained dataset contains 1,777 successful replays across the four
training suites.

\paragraph{Pointmaps and image resizing.}
Pointmaps are computed from depth and calibration, rather than supplied
by an independent point teacher. Before supervision, depth and validity
masks are resized to the geometry output resolution by nearest-exact
sampling. Intrinsics are updated consistently: resizing from
$(H_0,W_0)$ to $(H,W)$ multiplies the first and second rows of the
intrinsic matrix by $W/W_0$ and $H/H_0$, respectively. Extrinsics are
unchanged by resizing. The corresponding input dimensions are
$224\times288$ for RoboTwin and $256\times256$ for LIBERO.

\begin{figure}[t]
  \centering
    \includegraphics[width=0.88\textwidth]{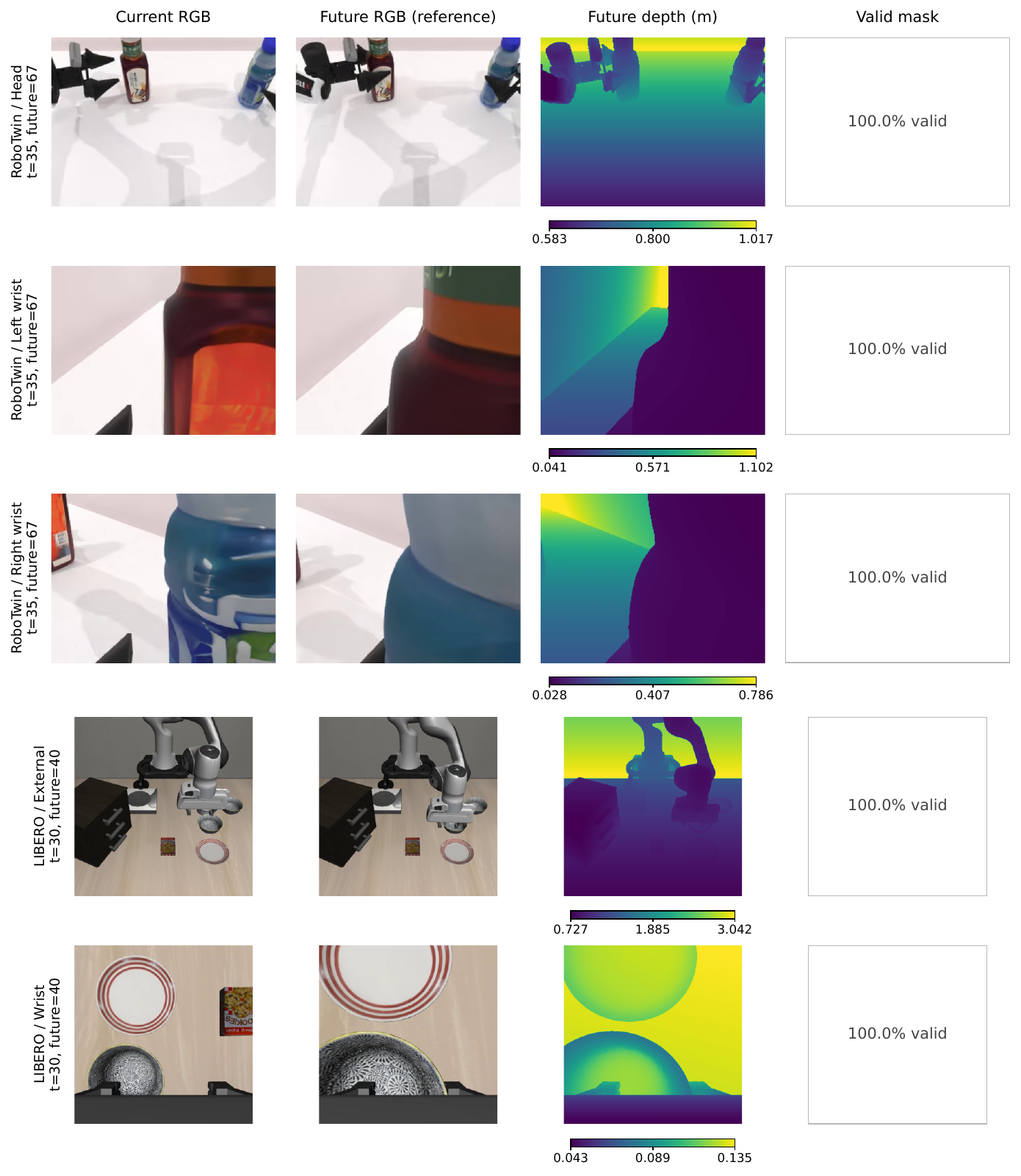}
  \caption{\textbf{Examples of future geometry supervision.}
  RoboTwin (top three rows) uses a future offset of 32 frames;
  LIBERO (bottom two rows) uses 10 frames.
  Columns show current RGB, future RGB for reference, future metric depth,
  and valid-pixel masks. Each depth panel has an independent color scale
  in meters, bounded by the 1st and 99th percentiles of its valid depths;
  colors are therefore not directly comparable across views.
  This reveals depth variation in the close-range wrist views.
  White mask pixels are valid and black pixels are invalid; all pixels
  in these examples are valid. Future RGB is shown only for illustration
  and is not a policy input. These are cached labels, not predictions.}
  \label{fig:supp_geometry_targets}
\end{figure}

\Cref{fig:supp_geometry_targets} illustrates the cached supervision
for both simulation datasets. Each depth panel uses its own metric color
scale to reveal local geometry, including close-range wrist-camera views.

\paragraph{Real robots.}
We obtain depth supervision for real-world demonstrations using
the metric-depth variant of Depth Anything 3 (DA3),
\texttt{DA3METRIC-LARGE}.
Each RGB frame is processed independently in single-frame mode.
The raw depth output is converted to metric depth as
\[
D(u) = D_{\mathrm{raw}}(u)\,\frac{f}{300},
\qquad
f = \frac{f_x + f_y}{2},
\]
where $D(u)$ is in meters and $f_x,f_y$ are the camera focal
lengths in pixels.
We use factory-calibrated camera intrinsics retrieved directly
from each camera through the manufacturer's SDK, without fitting
the intrinsics.
A pixel is valid if and only if $D(u)>0$.
We apply no additional depth-range clipping or confidence-based
filtering.

\subsection{\teal{Coordinates, Scale,} \graytext{and Camera Motion}}
\label{app:geometry_coordinates}

\paragraph{Unprojection.}
Let $K_{v,f}$ and $E_{v,f}=[R_{v,f}\mid\mathbf b_{v,f}]$ denote the
intrinsics and world-to-camera extrinsics of view $v$ at the future
time. OpenCV camera coordinates have $x$ pointing right, $y$ down,
and $z$ forward. For pixel $(u,v')$, the implementation uses the
pixel-center vector $\widetilde{\mathbf u}=(u+0.5,v'+0.5,1)^\top$.
Metric world points are
\begin{equation}
  P^w_{v,f}(\mathbf u)=R_{v,f}^{\top}
  \left[D_{v,f}(\mathbf u)K_{v,f}^{-1}\widetilde{\mathbf u}
  -\mathbf b_{v,f}\right].
  \label{eq:supp_geometry_unproject}
\end{equation}
Thus, depth is the camera-axis $z$ coordinate, not Euclidean distance
along the viewing ray.

\paragraph{Reference camera and scene scale.}
The simulation loss first expresses all target views in the first
camera's coordinate system at the \emph{target} time. In homogeneous
notation, $E'_{v,f}=E_{v,f}E_{1,f}^{-1}$, and
$P'_{v,f}=R_{1,f}P^w_{v,f}+\mathbf b_{1,f}$.
Let $M_{v,f}$ denote the valid-depth mask for view $v$ at the target time.
One scale is computed per sample over all valid pixels and views:
\begin{equation}
  s_f=\frac{\sum_{v,\mathbf u}M_{v,f}(\mathbf u)
            \|P'_{v,f}(\mathbf u)\|_2}
           {\sum_{v,\mathbf u}M_{v,f}(\mathbf u)}.
  \label{eq:supp_geometry_scale}
\end{equation}
Targets are $\bar D=D/s_f$, $\bar P=P'/s_f$, and
$\bar E_v=[R'_v\mid\mathbf b'_v/s_f]$; rotations and intrinsics
are unchanged. The first target camera has exactly $[I\mid0]$
extrinsics. Predictions directly learn these normalized targets;
they are not divided by the ground-truth scale before computing the
loss, nor aligned by a fitted per-image scale and shift.
Although the caches store metric geometry, the supervised outputs
therefore use scene-normalized units.

\paragraph{Moving wrist cameras.}
Each future depth image is indexed by its future camera's image plane,
and its pointmap is expressed in the shared future reference-camera
frame. Changing wrist poses affect both the visible surface and the
camera label. Future points are not unprojected using current wrist
extrinsics, and the objective does not impose temporal pixel
correspondence. The head/external view is the reference camera for
RoboTwin/LIBERO, respectively. Current-reconstruction ablations apply
the same transformation at $t$, with an independently computed $s_t$.
Even when the reference camera is stationary, independently normalized
current and future pointmaps need not have identical numerical scales.

\section{\teal{Future Geometry Diagnostics} \graytext{and World-Action Formulation}}
\label{sec:supp_future_geometry}

We complement the manipulation results with quantitative and qualitative
diagnostics of future geometry prediction. We evaluate the same
two-backbone checkpoint on training demonstrations and on observations
collected during closed-loop policy execution in XPolicyLab RoboTwin.
These settings assess, respectively, how well the model fits demonstrated
geometric futures and how its predictions agree with the futures actually
reached by the deployed policy. Both geometric experts receive current
multi-view RGB and the instruction and predict the same future endpoint.
The high-/low-noise labels identify their action-denoising intervals,
not different prediction horizons. Future images and geometry are used
only as references or evaluation targets.

\subsection{\teal{Evaluation} \graytext{Metrics}}
\label{sec:supp_geometry_metrics}

\paragraph{Coordinate convention and valid pixels.}
We use the same geometric target convention as in training.
For each target, ground-truth
points and cameras are expressed relative to the first (head) camera
at the future timestep. Depths, points, and camera translations are
divided by the mean distance of valid ground-truth points from that
camera, pooled across the three views. Predictions are evaluated directly
in this scene-normalized coordinate system, without fitting a scale,
rigid transform, or similarity transform to the target. Consequently,
distance errors below are in scene-normalized units, not meters.
We use valid ground-truth depth pixels without confidence-based
selection. Target depth and validity maps are resized to the prediction
resolution with nearest-neighbor sampling, with intrinsics scaled
accordingly.

\paragraph{Depth.}
We evaluate depth predictions using absolute relative error (AbsRel),
mean absolute error (MAE), root mean squared error (RMSE), and threshold
accuracy.
Let $\Omega$ be the valid pixels of one view, $N=|\Omega|$, and
$\hat d_p,d_p>0$ the predicted and target depths. We compute
\begin{align}
\mathrm{AbsRel} &= \frac{1}{N}\sum_{p\in\Omega}
                    \frac{|\hat d_p-d_p|}{d_p}, \\
\mathrm{MAE} &= \frac{1}{N}\sum_{p\in\Omega}|\hat d_p-d_p|, \\
\mathrm{RMSE} &= \sqrt{\frac{1}{N}\sum_{p\in\Omega}(\hat d_p-d_p)^2}, \\
\delta_j &= \frac{1}{N}\sum_{p\in\Omega}\mathbf{1}\!\left[
        \max\!\left(\frac{\hat d_p}{d_p},\frac{d_p}{\hat d_p}\right)<1.25^j
        \right].
\end{align}
Our tables report $\delta_1$ as a percentage; $\delta_2$ and $\delta_3$
are also recorded by the diagnostic implementation.

\paragraph{Points and cameras.}
For the offline diagnostic, Point L2 is the mean Euclidean distance
between corresponding predicted and ground-truth 3D points:
$N^{-1}\sum_{p\in\Omega}\|\hat{\mathbf P}_p-\mathbf P_p\|_2$.
Predicted points are unprojected from predicted depth, focal lengths,
and extrinsics, using the ground-truth principal point to match the
training loss convention. This is a pixel-corresponded point error,
rather than a nearest-neighbor point-cloud distance.
For world-to-camera extrinsics $[R\mid\mathbf t]$, we evaluate
\begin{align}
e_R &= \frac{180}{\pi}\arccos\!\left[
 \operatorname{clip}\!\left(\frac{\operatorname{tr}(\hat R R^\top)-1}{2},-1,1\right)
 \right], \\
e_C &= \|\hat{\mathbf C}-\mathbf C\|_2,
\qquad \mathbf C=-R^\top\mathbf t, \\
e_f &= \frac{1}{2}\sum_{a\in\{x,y\}}
                 \frac{|\hat f_a-f_a|}{f_a}.
\end{align}
Here, $e_R$ is in degrees, $e_C$ is in scene-normalized units, and
$e_f$ is dimensionless. Camera errors describe the relative multi-camera
configuration in the future head-camera frame, not absolute world-frame
localization.

\paragraph{Aggregation.}
We first compute each metric per view, average the three cameras within
each prediction, average predictions within each task, and finally average
tasks equally. Thus, longer rollouts and views with more valid pixels do
not receive greater weight in the reported task average. In particular,
RMSE is averaged after taking the per-view square root. We report each
expert separately. An expert-mean row averages their scalar metrics;
it does not evaluate an ensemble of their predicted geometry.

\subsection{\teal{Training-Set} \graytext{Diagnostic}}
\label{sec:supp_geometry_offline}

\paragraph{Protocol.}
We select two demonstration episodes per task using seed 42, and two
starting frames per episode at 20\% and 60\% of the valid starting-frame
range. This yields 200 samples from 50 tasks and 600 camera views per
expert. Every target is a recorded, non-padded future frame at offset
32, corresponding to approximately 2.13\,s at the dataset's 15\,Hz rate.
Both experts are evaluated on exactly the same samples, without
training-time image augmentation. The model sees only the current RGB
views and instruction when predicting geometry.

\begin{figure}[t]
\centering
\includegraphics[width=\textwidth]{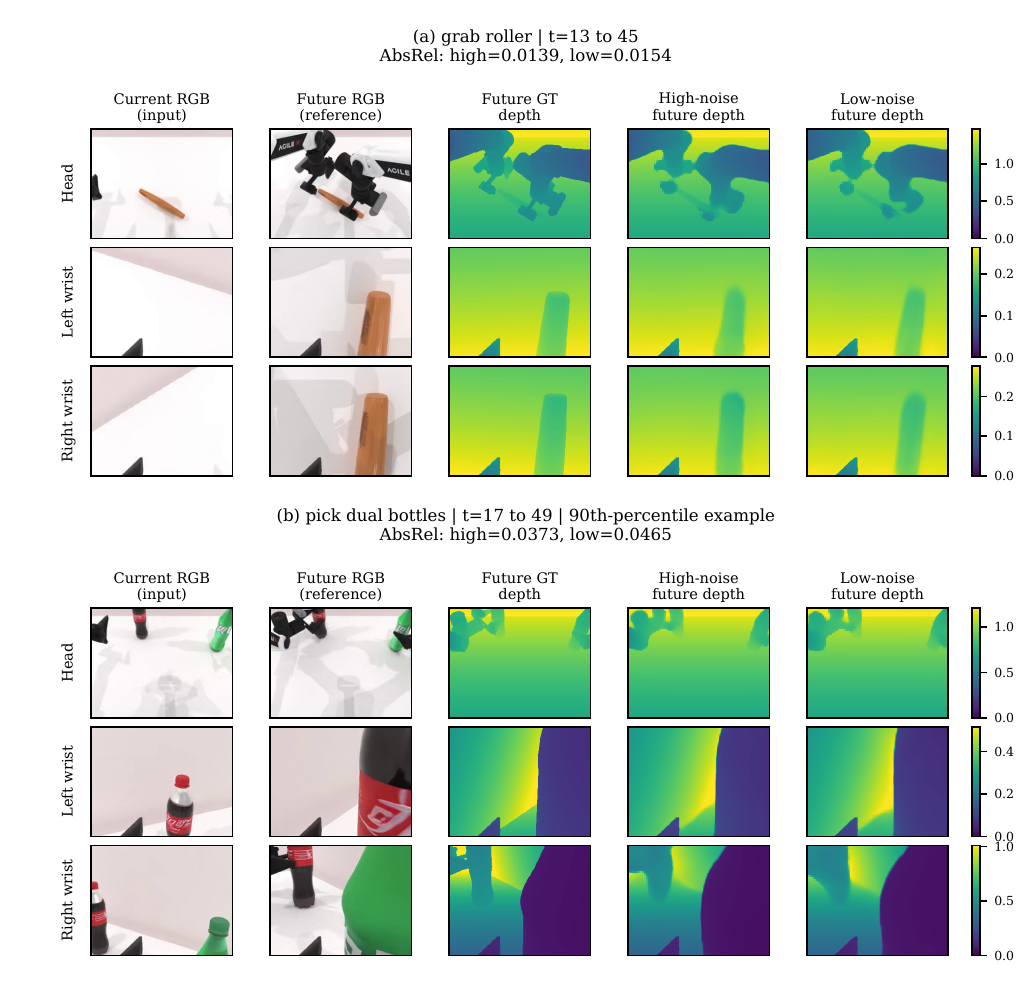}
\caption{\textbf{Future geometry prediction on training demonstrations.}
Each example shows the head and two wrist views. Columns contain the
current RGB input, future RGB reference, future ground-truth depth,
and predictions from the high- and low-noise experts. Panel (a) is
selected for visible arm motion between the current and future frames;
panel (b) is near the 90th percentile of paired mean AbsRel over all
200 samples. Quantitative results use all samples. For each view, the
GT and both predictions share a depth color range from zero to that
view's 99th percentile of valid GT depth, in scene-normalized units;
larger values saturate for display only.
All quantitative metrics use the original values. Future RGB is not
a model input.}
\label{fig:supp_geometry_offline}
\end{figure}

\paragraph{Results.}
\Cref{tab:supp_geometry_offline} shows close agreement between
predicted and demonstrated future geometry for both experts. Their
mean AbsRel is 0.0206, $\delta_1$ is 98.23\%, and mean camera rotation
error is $0.666^\circ$, matching \cref{sec:sim_experiments}. The low point
and camera errors further show that this fit extends beyond depth alone
to the relative multi-view scene configuration.
\Cref{fig:supp_geometry_offline} illustrates the predicted future
depth across all three cameras. Panel (a) shows the arms approaching
the roller, with clear changes in both the head and wrist views.
Panel (b) retains a more difficult example near the 90th percentile
of the two experts' mean AbsRel. Together, these quantitative and
qualitative results demonstrate our model's strong ability to learn
future geometry from demonstrations, capturing the spatial structure
and geometric changes associated with the demonstrated interactions.

\subsection{\teal{Closed-Loop Rollout} \graytext{Diagnostic}}
\label{sec:supp_geometry_online}

\paragraph{Protocol.}
We run one policy rollout for each of the 50 RoboTwin tasks under
\texttt{demo\_clean}, using 10 action-denoising steps and replanning
every 24 executed action steps. At each planning step $t$, both
geometric experts predict the endpoint at $t+32$ from the same current
observation. We retain these predictions and compare them with the
simulator observation actually reached 32 actions later. This interval
includes the next replanning event at $t+24$, so the target reflects
closed-loop execution rather than uninterrupted execution of the
initially predicted chunk. Predictions whose targets fall beyond
episode termination are recorded as unpaired and are not padded.
Future RGB, depth, and calibration are used only for diagnosis.

We use the run that records both depth and camera parameters, yielding
544 paired predictions (1,632 views) per expert over 50 rollouts and
66 unpaired terminal predictions per expert. All 50 tasks contribute
equally to the reported averages. This is a diagnostic sample of executions, not an additional
success-rate benchmark or a multi-seed significance study.

\begin{table}[t]
\centering
\small
\caption{\textbf{Future depth during closed-loop policy execution.}
Task-macro averages over all 50 tasks / 544 paired predictions per expert.
The upper block uses all valid
future GT pixels. The lower block evaluates every method on the same
intersection of current and future valid masks. Copy-current uses
simulator GT depth at $t$, normalized at $t$, without temporal warping.
It is a privileged persistence reference, not an RGB-only learned model.}
\label{tab:supp_geometry_online}
\begin{tabular}{llrrrr}
\toprule
Evaluation mask & Prediction & AbsRel $\downarrow$ & $\delta_1$ (\%) $\uparrow$ & MAE $\downarrow$ & RMSE $\downarrow$ \\
\midrule
Future valid & High-noise & 0.088463 & 90.33 & 0.038555 & 0.074471 \\
             & Low-noise  & 0.089807 & 89.88 & 0.039324 & 0.074119 \\
\midrule
Common valid & Copy-current GT & 0.464132 & 66.11 & 0.122977 & 0.169056 \\
             & High-noise & 0.088400 & 90.34 & 0.038375 & 0.074150 \\
             & Low-noise  & 0.089722 & 89.88 & 0.039178 & 0.073865 \\
\bottomrule
\end{tabular}
\end{table}

\begin{table}[t]
\centering
\small
\caption{\textbf{Future camera prediction during closed-loop execution.}
The same 50 tasks and 544 pairs as \cref{tab:supp_geometry_online}. Errors use the future head-camera
frame and retain all measured outliers.}
\label{tab:supp_geometry_online_camera}
\begin{tabular}{lrrr}
\toprule
Expert & $e_R$ ($^\circ$) $\downarrow$ & $e_C$ $\downarrow$ & $e_f$ $\downarrow$ \\
\midrule
High-noise & 3.122 & 0.018778 & 0.032269 \\
Low-noise  & 3.475 & 0.019462 & 0.000148 \\
\bottomrule
\end{tabular}
\end{table}

\paragraph{Prediction beyond persistence.}
\Cref{tab:supp_geometry_online} reports higher errors than the
training-set diagnostic, but both experts substantially outperform
copying the current scene. On the common valid mask, AbsRel decreases
from 0.4641 for copy-current to 0.0884 and 0.0897, while $\delta_1$
increases from 66.11\% to 90.34\% and 89.88\%. This comparison indicates
that the predictions capture changes in future view geometry beyond
simple persistence. It does not isolate object motion: the depth
changes also reflect wrist-camera motion and the per-scene normalization.

\begin{figure}[t]
\centering
\includegraphics[width=\textwidth]{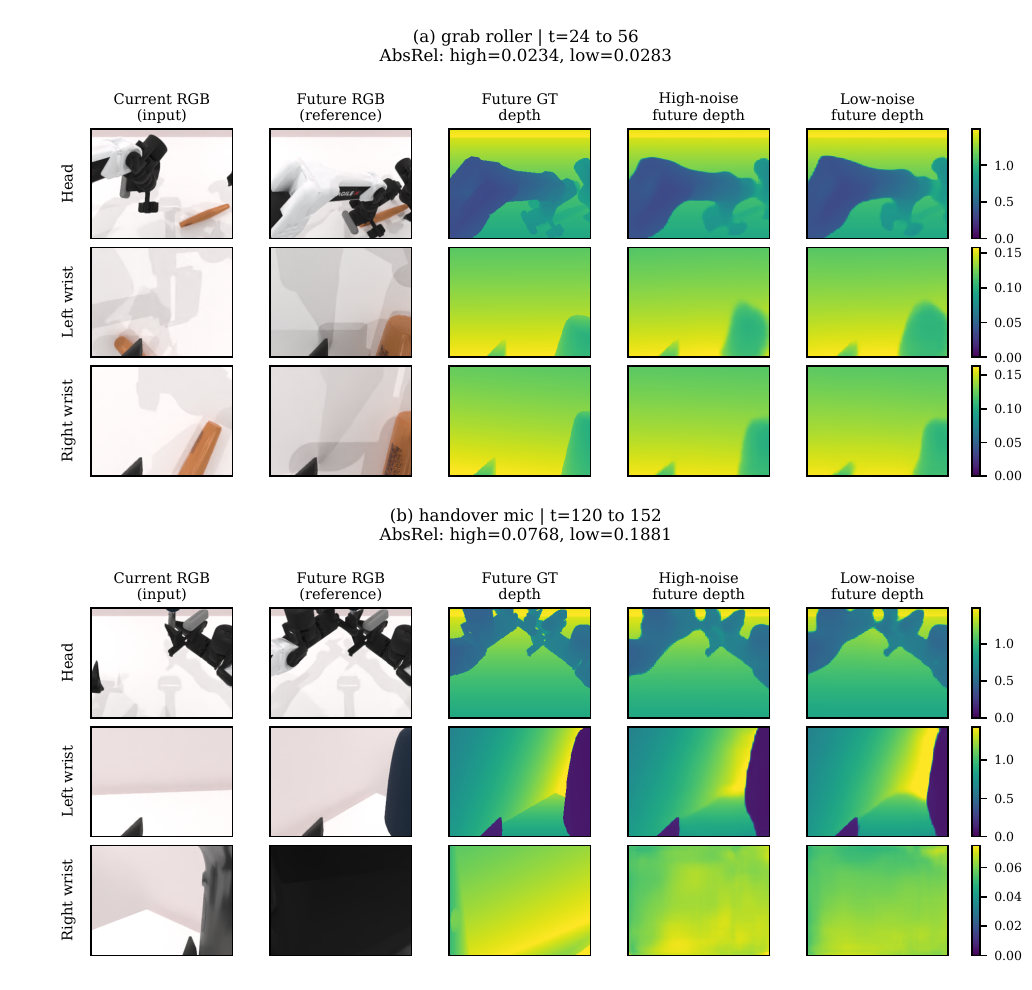}
\caption{\textbf{Future geometry during actual closed-loop rollouts.}
The columns match \cref{fig:supp_geometry_offline}. Examples are
selected for visible current-to-future changes in arm configuration and
wrist-camera observations, rather than by prediction-error percentile.
For each view, the GT and both predictions share a depth color range
from zero to that view's 99th percentile of valid GT depth, in
scene-normalized units; upper-tail saturation affects display only.
Both experts are compared with the same realized future at $t+32$,
including replanning at $t+24$. Quantitative results use all 544 paired
predictions from all 50 tasks, not just these illustrative examples.}
\label{fig:supp_geometry_online}
\end{figure}

\paragraph{Qualitative behavior and remaining errors.}
\Cref{fig:supp_geometry_online} shows actual rollout predictions
selected to illustrate visible changes between current and future observations.
In \texttt{grab\_roller}, the arms approach the roller; in
\texttt{handover\_mic}, the arms change configuration and the wrist views
undergo substantial changes in visibility. The predictions capture much
of the resulting future structure, with remaining errors around arm
boundaries and in the wrist views. These illustrative examples are not
intended to represent average prediction accuracy.
Across all 50 tasks, head-view AbsRel is 0.0455/0.0460 for the high-/low-noise
experts, compared with 0.1069/0.1080 for the left wrist and
0.1130/0.1154 for the right wrist. Future camera rotation errors also
increase to $3.122^\circ$/$3.475^\circ$
(\cref{tab:supp_geometry_online_camera}). The high-noise expert's
larger focal error is dominated by a single extreme prediction in
\texttt{turn\_switch}; we retain it in the average. These observations
identify moving-camera and articulated-scene prediction as remaining
limitations. The training-to-rollout gap is consistent with the change
from demonstrated trajectories to policy-induced states and with
intervening replanning, although this diagnostic does not separate
their individual effects.

\subsection{\teal{Policy-Aligned World-Action Modeling} \graytext{vs.\ Action-Conditioned Simulation}}
\label{sec:policy_aligned_wam}

A central architectural question is how E0.5 functions as a world-action model (WAM) when the geometric branch $B_{\theta_k}(O_t, \ell)$ does not take the candidate action sequence $A_t$ (or noisy action latent $U_\sigma$) and proprioceptive state $s_t$ as forward-pass inputs. Clarifying this design distinguishes two paradigms for coupling world modeling with robot control.

\paragraph{Two factorizations of world-action modeling.}
Let $(O_t, s_t, \ell)$ denote the current multi-view observations, robot state, and language instruction, $A_t = (a_t, \dots, a_{t+H_a-1})$ the target action chunk, and $G_t^+ = (D_{t_f}, P_{t_f}, C_{t_f})$ the 3D scene geometry at the action-horizon endpoint $t_f = \tau(t + H_a)$. Approaches to joint world-state and action modeling fall into two categories:
\begin{enumerate}
    \item \emph{Action-conditioned forward simulators} factorize the joint distribution as $p(G_t^+, A_t \mid O_t, s_t, \ell) = p(G_t^+ \mid O_t, s_t, A_t)\,\pi(A_t \mid O_t, s_t, \ell)$. Here, the world model acts as an explicit transition simulator over arbitrary candidate actions (e.g., for test-time search or model-predictive control). While expressive for counterfactual action evaluation, this formulation requires conditioning the world backbone on candidate actions and typically incurs repeated backbone evaluations and explicit state decoding during deployment.
    \item \emph{Policy-aligned joint WAMs}---the paradigm instantiated by E0.5, in line with test-time imagination-free frameworks such as Unified Video Action Model, FastWAM, and Image\-WAM---model the joint distribution through shared latent representations without test-time rollout search. Under the demonstration distribution induced by a target policy $\pi$, marginalizing over policy actions yields the task-conditioned future geometry distribution
    \begin{equation}
        p_\pi(G_t^+ \mid O_t, \ell) = \int p(G_t^+ \mid O_t, A_t)\,\pi(A_t \mid O_t, \ell)\,\mathrm{d}A_t,
        \label{eq:marginal_future_geo}
    \end{equation}
    where multi-view RGB observations $O_t$ (head/external plus wrist cameras) provide direct visual observability of the manipulator configuration, while exact joint-space proprioception $s_t$ conditions the action expert $\pi_{\psi_a}$ across all 30 ActionDiT blocks for fine-grained kinematic disambiguation.
\end{enumerate}

At denoising level $\sigma$ with selected expert $k = r(\sigma)$, E0.5 couples future geometry prediction and flow-matching control through the shared geometric backbone state $\mathcal{H}_t(\sigma) = B_{\theta_k}(O_t, \ell)$ (\cref{eq:native_geometric_wam,eq:geometry_mixture}): both the geometric head $Q_{\psi_g}^{(k)}$ and the shared velocity decoder $v_{\psi_a}$ read from the same backbone computation rather than cascading explicit future reconstructions into a separate policy.

\paragraph{Unidirectional forward pass and bidirectional gradient coupling.}
Omitting $U_\sigma$ from $B_{\theta_k}$ is a deliberate architectural choice that serves two purposes. First, during early denoising ($\sigma \to 1$), $U_\sigma$ approaches isotropic Gaussian noise; excluding noisy action tokens from $B_{\theta_k}$ preserves the pretrained multi-view geometric priors of VGGT-Omega under low-data adaptation. Second, because $\mathcal{H}_t(\sigma)$ is invariant to $U_\sigma$ within each routing interval, layer-wise geometric keys and values $M^{(k)}$ can be computed once per visited expert and cached across denoising steps (\cref{alg:supp_inference}), avoiding redundant 3D backbone evaluations at every Euler step and eliminating geometry-head decoding at test time.

Crucially, unidirectional forward conditioning does not reduce $B_{\theta_k}$ to a static subgoal extractor. During training, because the projected geometric K/V features $M^{(k)}$ are passed to the ActionDiT without gradient detachment, the parameters $\theta_k$ of each routed backbone receive coupled supervision from \cref{eq:joint_world_action_objective} (implemented as \cref{eq:supp_batch_loss}):
\begin{equation}
    \nabla_{\theta_k}\mathcal{L}
    = \underbrace{\frac{\partial \mathcal{L}_{\mathrm{act}}}{\partial M^{(k)}} \frac{\partial M^{(k)}}{\partial \theta_k}}_{\text{action-affordance gradient}}
    + \;\lambda_g \underbrace{\frac{\partial \mathcal{L}_{\mathrm{future}}^{(K)}}{\partial \mathcal{H}_t(\sigma)} \frac{\partial \mathcal{H}_t(\sigma)}{\partial \theta_k}}_{\text{future-geometry gradient}}.
    \label{eq:coupled_gradient}
\end{equation}
The second term forces the backbone to anticipate control-horizon spatial dynamics from $(O_t, \ell)$, while the first term shapes those same multi-scale geometric representations to directly support continuous action generation.

\paragraph{Amortizing dynamics into representations.}
This formulation also explains why E0.5 achieves strong closed-loop manipulation gains despite imperfect pixel-level future reconstructions on articulated robot arms and moving wrist views (\cref{tab:supp_geometry_online,tab:supp_geometry_online_camera}). Because E0.5 never bottlenecks control through reconstructed point clouds $\widehat{G}_t^+$ at deployment, world-dynamics reasoning is amortized during training into the intermediate geometric K/V representations $M^{(k)}$. As confirmed by the controlled ablations in \cref{tab:ablation_future}, shifting supervision from current-scene reconstruction ($G_t$) to horizon-aligned future geometry ($G_t^+$) under identical inference architectures improves RoboTwin Randomized success by 12.81 percentage points at $K=1$ and 15.44 points at $K=2$, validating policy-aligned future geometry supervision as an effective mechanism for learning dynamics-aware control representations.

\section{\teal{Evaluation Protocols} \graytext{and Detailed Results}}
\label{sec:supp_evaluation}

This section describes the evaluation settings and task-success criteria
and provides detailed results for the simulation benchmarks and real-robot
experiments. Success rates are reported in percent.

\subsection{\teal{XPolicyLab} \graytext{RoboTwin}}
\label{sec:supp_eval_robotwin}

\paragraph{Tasks and evaluation conditions.}
We evaluate all 50 bimanual manipulation tasks using the XPolicyLab
RoboTwin clean-to-random setting. Training uses 50 clean demonstrations
per task. Clean evaluation uses \texttt{demo\_clean} with seen language
instructions, and Random evaluation uses \texttt{demo\_randomized} with
unseen instructions. Random evaluation introduces randomized background
textures, table clutter, lighting, and table height. In the evaluated
configuration, the table-height offset is sampled from $[-0.03,0]$ m;
the clean-background and extreme-lighting probabilities are both 0.02.
The head-camera displacement parameter is zero. Thus, this condition
changes both scene appearance/configuration and the instruction split.

\paragraph{Rollouts.}
The policy receives the head and two wrist RGB views, robot state, and
the task instruction, and predicts 32 actions. The simulator checks
task-specific success predicates and terminates on success or at the
task's action limit, which ranges from 400 to 1,700 steps. The evaluator
first tests candidate scenes with the scripted expert and evaluates the
policy on accepted scenes. Evaluation targets 100 accepted trials per
task and condition. We use seed 42 and action-branch
language-guidance scale $g=1.0$.

\paragraph{Detailed results.}
\Cref{tab:eval_robot_tasks} reports the results for all 50 tasks.
The task-macro success rates are 89.30\% on Clean and 79.90\% on
Random, with an arithmetic mean of 84.60\%.

\begin{table}[!tp]
\centering
\footnotesize
\setlength{\tabcolsep}{3.5pt}
\renewcommand{\arraystretch}{1.05}
\caption{Per-task RoboTwin success (\%).
Results and task-macro averages reproduce those reported in \cref{tab:robotwin_c2r}.
Each task--condition pair uses 100 evaluation trials,
except for the Random condition of
\texttt{put\_object\_cabinet}
($\dagger$: 39 successes out of 83 completed trials,
yielding 46.99\%). Completing the remaining 17 trials for this task with
the same checkpoint and evaluation protocol yielded
48/100 successes (48.00\%).
This additional evaluation is not included in the
per-task results or averages shown here.}
\label{tab:eval_robot_tasks}

\begin{tabular*}{\textwidth}{
    @{\extracolsep{\fill}}
    lrr
    @{\hspace{2em}}
    lrr
    @{}
}
\toprule
Task & Clean & Random
& Task & Clean & Random \\
\cmidrule(lr){1-3}\cmidrule(lr){4-6}
adjust\_bottle & 100.00 & 98.00
& place\_can\_basket & 70.00 & 56.00 \\
beat\_block\_hammer & 98.00 & 70.00
& place\_cans\_plasticbox & 99.00 & 90.00 \\
blocks\_ranking\_rgb & 95.00 & 98.00
& place\_container\_plate & 99.00 & 93.00 \\
blocks\_ranking\_size & 81.00 & 82.00
& place\_dual\_shoes & 77.00 & 73.00 \\
click\_alarmclock & 100.00 & 94.00
& place\_empty\_cup & 99.00 & 93.00 \\
click\_bell & 100.00 & 91.00
& place\_fan & 91.00 & 80.00 \\
dump\_bin\_bigbin & 98.00 & 86.00
& place\_mouse\_pad & 75.00 & 65.00 \\
grab\_roller & 100.00 & 100.00
& place\_object\_basket & 94.00 & 89.00 \\
handover\_block & 89.00 & 90.00
& place\_object\_scale & 89.00 & 73.00 \\
handover\_mic & 98.00 & 96.00
& place\_object\_stand & 98.00 & 94.00 \\
hanging\_mug & 47.00 & 36.00
& place\_phone\_stand & 96.00 & 90.00 \\
lift\_pot & 100.00 & 72.00
& place\_shoe & 94.00 & 91.00 \\
move\_can\_pot & 90.00 & 74.00
& press\_stapler & 90.00 & 84.00 \\
move\_pillbottle\_pad & 94.00 & 90.00
& put\_bottles\_dustbin & 92.00 & 80.00 \\
move\_playingcard\_away & 99.00 & 97.00
& put\_object\_cabinet & 52.00 & 46.99$^{\dagger}$ \\
move\_stapler\_pad & 82.00 & 51.00
& rotate\_qrcode & 93.00 & 65.00 \\
open\_laptop & 89.00 & 86.00
& scan\_object & 91.00 & 64.00 \\
open\_microwave & 52.00 & 41.00
& shake\_bottle & 100.00 & 97.00 \\
pick\_diverse\_bottles & 86.00 & 70.00
& shake\_bottle\_horizontally & 100.00 & 98.00 \\
pick\_dual\_bottles & 96.00 & 81.00
& stack\_blocks\_three & 93.00 & 92.00 \\
place\_a2b\_left & 81.00 & 63.00
& stack\_blocks\_two & 100.00 & 99.00 \\
place\_a2b\_right & 88.00 & 56.00
& stack\_bowls\_three & 87.00 & 90.00 \\
place\_bread\_basket & 96.00 & 87.00
& stack\_bowls\_two & 98.00 & 92.00 \\
place\_bread\_skillet & 94.00 & 74.00
& stamp\_seal & 80.00 & 56.00 \\
place\_burger\_fries & 98.00 & 93.00
& turn\_switch & 57.00 & 68.00 \\
\midrule
\multicolumn{4}{l}{Task-macro mean (all 50 tasks)}
& 89.30 & 79.90 \\
\bottomrule
\end{tabular*}
\end{table}

\subsection{\teal{LIBERO} \graytext{Benchmark}}
\label{sec:supp_eval_libero}

\paragraph{Tasks and initialization.}
We evaluate the Spatial, Object, Goal, and Long suites, with ten tasks
per suite. Long corresponds to \texttt{libero\_10}. Each task is evaluated
once from each of its 50 stored initial states, taken in stored order,
giving 500 episodes per suite and 2,000 episodes in total. The evaluator
resets the random seed to 42 for each task. After resetting the
environment and restoring an initial state, it executes 30 dummy actions
to allow the scene to settle before policy control begins.

\paragraph{Observations, actions, and stopping rules.}
The policy receives the external and wrist RGB views, proprioception,
and the task instruction. Model preprocessing and normalization
statistics are restored from the training run. We use a ten-action
prediction horizon, ten denoising steps, and $R=10$, with action
ensembling disabled. The gripper command is converted to the simulator's
action convention and binarized. Evaluation uses the original camera
configuration. The maximum policy-control budgets are 400 steps for
Spatial, Object, and Goal and 700 for Long, in addition to the 30
settling steps. An episode succeeds when the simulator reports task
completion; otherwise it fails when the control budget is exhausted.

\paragraph{Metrics and results.}
For suite $b$, let $\mathcal I_b$ be its evaluated task--initial-state
pairs and $y_i\in\{0,1\}$ the success of pair $i$. We report
\begin{equation}
 S_b=\frac{100}{|\mathcal I_b|}\sum_{i\in\mathcal I_b}y_i,
 \qquad S_{\mathrm{overall}}=\frac{1}{4}\sum_{b=1}^{4}S_b.
 \label{eq:supp_suite_macro}
\end{equation}
The four suites have equal episode counts, so the suite mean also
equals pooled episode success. E0.5 succeeds in 497/500 Spatial,
497/500 Object, 493/500 Goal, and 491/500 Long episodes, corresponding
to 99.40\%, 99.40\%, 98.60\%, and 98.20\%. The overall result is
1,978/2,000 (98.90\%). \Cref{tab:eval_libero_tasks} provides all
40 task results. The largest remaining failure counts occur in placing
the wine bottle on the cabinet (Goal, 45/50) and the composite
mug-and-pudding task (Long, 46/50).

\begin{table}[t]
\centering
\scriptsize
\setlength{\tabcolsep}{3.5pt}
\caption{All 40 LIBERO tasks. IDs are zero-based within each suite. Each task uses 50 initial states and one attempt per state.}
\label{tab:eval_libero_tasks}
\begingroup
\setbox0=\hbox{%
\begin{tabular}{lrp{0.65\textwidth}rr}
\toprule
Suite & ID & Instruction & Successes & SR (\%) \\
\midrule
Spatial & 0 & Pick up the black bowl between the plate and the ramekin and place it on the plate. & 50/50 & 100 \\
 & 1 & Pick up the black bowl next to the ramekin and place it on the plate. & 49/50 & 98 \\
 & 2 & Pick up the black bowl from table center and place it on the plate. & 50/50 & 100 \\
 & 3 & Pick up the black bowl on the cookie box and place it on the plate. & 50/50 & 100 \\
 & 4 & Pick up the black bowl in the top drawer of the wooden cabinet and place it on the plate. & 49/50 & 98 \\
 & 5 & Pick up the black bowl on the ramekin and place it on the plate. & 50/50 & 100 \\
 & 6 & Pick up the black bowl next to the cookie box and place it on the plate. & 50/50 & 100 \\
 & 7 & Pick up the black bowl on the stove and place it on the plate. & 50/50 & 100 \\
 & 8 & Pick up the black bowl next to the plate and place it on the plate. & 49/50 & 98 \\
 & 9 & Pick up the black bowl on the wooden cabinet and place it on the plate. & 50/50 & 100 \\
\midrule
Object & 0 & Pick up the alphabet soup and place it in the basket. & 50/50 & 100 \\
 & 1 & Pick up the cream cheese and place it in the basket. & 50/50 & 100 \\
 & 2 & Pick up the salad dressing and place it in the basket. & 50/50 & 100 \\
 & 3 & Pick up the bbq sauce and place it in the basket. & 49/50 & 98 \\
 & 4 & Pick up the ketchup and place it in the basket. & 50/50 & 100 \\
 & 5 & Pick up the tomato sauce and place it in the basket. & 50/50 & 100 \\
 & 6 & Pick up the butter and place it in the basket. & 50/50 & 100 \\
 & 7 & Pick up the milk and place it in the basket. & 48/50 & 96 \\
 & 8 & Pick up the chocolate pudding and place it in the basket. & 50/50 & 100 \\
 & 9 & Pick up the orange juice and place it in the basket. & 50/50 & 100 \\
\midrule
Goal & 0 & Open the middle drawer of the cabinet. & 49/50 & 98 \\
 & 1 & Put the bowl on the stove. & 50/50 & 100 \\
 & 2 & Put the wine bottle on top of the cabinet. & 45/50 & 90 \\
 & 3 & Open the top drawer and put the bowl inside. & 50/50 & 100 \\
 & 4 & Put the bowl on top of the cabinet. & 50/50 & 100 \\
 & 5 & Push the plate to the front of the stove. & 50/50 & 100 \\
 & 6 & Put the cream cheese in the bowl. & 50/50 & 100 \\
 & 7 & Turn on the stove. & 50/50 & 100 \\
 & 8 & Put the bowl on the plate. & 50/50 & 100 \\
 & 9 & Put the wine bottle on the rack. & 49/50 & 98 \\
\midrule
Long & 0 & Put both the alphabet soup and the tomato sauce in the basket. & 50/50 & 100 \\
 & 1 & Put both the cream cheese box and the butter in the basket. & 50/50 & 100 \\
 & 2 & Turn on the stove and put the moka pot on it. & 50/50 & 100 \\
 & 3 & Put the black bowl in the bottom drawer of the cabinet and close it. & 50/50 & 100 \\
 & 4 & Put the white mug on the left plate and put the yellow and white mug on the right plate. & 47/50 & 94 \\
 & 5 & Pick up the book and place it in the back compartment of the caddy. & 50/50 & 100 \\
 & 6 & Put the white mug on the plate and put the chocolate pudding to the right of the plate. & 46/50 & 92 \\
 & 7 & Put both the alphabet soup and the cream cheese box in the basket. & 50/50 & 100 \\
 & 8 & Put both moka pots on the stove. & 48/50 & 96 \\
 & 9 & Put the yellow and white mug in the microwave and close it. & 50/50 & 100 \\
\bottomrule
\end{tabular}%
}
\leavevmode
\ifdim\wd0>\linewidth
  \resizebox{\linewidth}{!}{\box0}%
\else
  \box0
\fi
\endgroup
\end{table}

\subsection{\teal{LIBERO-Plus} \graytext{Robustness Suite}}
\label{sec:supp_eval_plus}

\paragraph{Perturbations and rollout protocol.}
We evaluate the same LIBERO-trained checkpoint without further
fine-tuning on all 10,030 perturbed task instances: 2,402 Spatial,
2,518 Object, 2,591 Goal, and 2,519 Long. The seven categories are
camera viewpoints, robot initial states, language instructions,
lighting, background textures, sensor noise, and object layout.
We use the benchmark-provided initialization for each perturbed instance. Language conditioning uses the
instruction in the actual task specification, including rewritten
instructions. We preserve the benchmark's camera perturbations and
apply no additional camera shift. The 30 settling actions, control
budgets, seed, and action processing follow \cref{sec:supp_eval_libero}.

We use ten denoising steps. The geometric branch's language
conditioning is unchanged. Task completion is checked by the simulator
using the corresponding perturbed task specification.

\paragraph{Detailed results.}
\Cref{tab:eval_plus_categories,tab:eval_plus_suites}
report results by perturbation category and base suite. The overall
result is 8,905/10,030 (88.78\%), weighted by the number of instances;
the equal-suite mean is 88.97\%. Lighting has the highest success rate
(98.51\%), while robot initial states (83.74\%) and camera viewpoints
(84.37\%) remain more challenging.

\begin{table}[t]
\centering
\footnotesize
\setlength{\tabcolsep}{3.5pt}
\caption{LIBERO-Plus results by perturbation.}
\label{tab:eval_plus_categories}
\begingroup
\setbox0=\hbox{%
\begin{tabular}{lrr}
\toprule
Perturbation & Successful cases & SR (\%) \\
\midrule
Camera Viewpoints & 1349/1599 & 84.37 \\
Robot Initial States & 1298/1550 & 83.74 \\
Language Instructions & 1353/1537 & 88.03 \\
Light Conditions & 1125/1142 & 98.51 \\
Background Textures & 1028/1076 & 95.54 \\
Sensor Noise & 1447/1601 & 90.38 \\
Objects Layout & 1305/1525 & 85.57 \\
\bottomrule
\end{tabular}%
}
\leavevmode
\ifdim\wd0>\linewidth
  \resizebox{\linewidth}{!}{\box0}%
\else
  \box0
\fi
\endgroup
\end{table}

\begin{table}[t]
\centering
\footnotesize
\setlength{\tabcolsep}{3.5pt}
\caption{LIBERO-Plus suite results. Overall pools all 10,030 instances; the suite-macro mean weights the four suites equally.}
\label{tab:eval_plus_suites}
\begingroup
\setbox0=\hbox{%
\begin{tabular}{lrr}
\toprule
Suite & Successful cases & SR (\%) \\
\midrule
Spatial & 2339/2402 & 97.38 \\
Object & 2345/2518 & 93.13 \\
Goal & 1973/2591 & 76.15 \\
Long & 2248/2519 & 89.24 \\
\midrule
Suite macro & -- & 88.97 \\
Overall & 8905/10030 & 88.78 \\
\bottomrule
\end{tabular}%
}
\leavevmode
\ifdim\wd0>\linewidth
  \resizebox{\linewidth}{!}{\box0}%
\else
  \box0
\fi
\endgroup
\end{table}

\subsection{\teal{LIBERO-PRO} \graytext{Distribution Shifts}}
\label{sec:supp_eval_pro}

\paragraph{Task variants and control budgets.}
We evaluate Object, Position, Semantic, Task, and Environment shifts
over the same four base suites. Each category contains 40 tasks with
50 initial states each, giving 2,000 evaluated task--state pairs per
category and 10,000 pairs across the five categories. Each variant uses
its own task specification, instruction, success predicates, and
initialization states. Position and Semantic correspond to
\texttt{swap} and \texttt{lan} in the evaluator. Initial states are
processed in stored order. We use seed 42 and 30 settling actions,
followed by at most 220, 280, 300, and 520 policy steps for Spatial,
Object, Goal, and Long, respectively. These PRO control budgets differ
from the LIBERO/Plus budgets above.
We reuse the LIBERO checkpoint with ten denoising steps and $g=1.0$. Each rollout is judged by the success predicates
of its perturbed task specification.

\paragraph{Detailed results.}
\Cref{tab:eval_pro_categories,tab:eval_pro_suites}
report the overall result of 5,536/10,000 (55.36\%) and its category
and suite decomposition. Because all categories and base suites
have equal numbers of cases, category-macro, suite-macro, and pooled
averages agree. Semantic changes have the highest success (94.35\%), followed
by object substitutions (83.45\%). Position changes (23.05\%) and
task changes (22.15\%) remain the principal weaknesses; the
Environment extension achieves 53.80\%.

\begin{table}[t]
\centering
\footnotesize
\setlength{\tabcolsep}{3.5pt}
\caption{LIBERO-PRO category results.}
\label{tab:eval_pro_categories}
\begingroup
\setbox0=\hbox{%
\begin{tabular}{lrr}
\toprule
Category & Successful cases & SR (\%) \\
\midrule
Object & 1669/2000 & 83.45 \\
Position & 461/2000 & 23.05 \\
Semantic & 1887/2000 & 94.35 \\
Task & 443/2000 & 22.15 \\
Environment & 1076/2000 & 53.80 \\
\midrule
Overall & 5536/10000 & 55.36 \\
\bottomrule
\end{tabular}%
}
\leavevmode
\ifdim\wd0>\linewidth
  \resizebox{\linewidth}{!}{\box0}%
\else
  \box0
\fi
\endgroup
\end{table}

\begin{table}[t]
\centering
\footnotesize
\setlength{\tabcolsep}{3.5pt}
\caption{LIBERO-PRO success (\%) by category and base suite. Each cell contains ten tasks with 50 states each (500 cases).}
\label{tab:eval_pro_suites}
\begingroup
\setbox0=\hbox{%
\begin{tabular}{lrrrrr}
\toprule
Category & Spatial & Object & Goal & Long & Overall \\
\midrule
Object & 99.60 & 92.20 & 76.40 & 65.60 & 83.45 \\
Position & 56.80 & 8.20 & 13.40 & 13.80 & 23.05 \\
Semantic & 88.40 & 100.00 & 89.40 & 99.60 & 94.35 \\
Task & 51.40 & 9.40 & 15.80 & 12.00 & 22.15 \\
Environment & 66.20 & 54.80 & 41.20 & 53.00 & 53.80 \\
\bottomrule
\end{tabular}%
}
\leavevmode
\ifdim\wd0>\linewidth
  \resizebox{\linewidth}{!}{\box0}%
\else
  \box0
\fi
\endgroup
\end{table}

\subsection{\teal{Real-World Task Suite} \graytext{and Scoring Criteria}}
\label{sec:supp_eval_real}
\label{app:real_task_protocol}

We evaluate mouse placement, charger packing, and trash disposal on Zerith H1 and AgileX Piper, with per-task and three-task mean results reported in \cref{tab:h1_real_world} (for $K=2$ without embodied pretraining) and \cref{tab:real_robot_pretrain_results} (for $K=1$ with and without AgiBotWorld-Beta pretraining). Below we detail the task objectives, ID/OOD conditions, and formal completion-score definitions.

\paragraph{Evaluation settings.}
Each task is evaluated under in-distribution (ID) and out-of-distribution (OOD) conditions, corresponding to the EASY and HARD settings. ID trials use objects and initial configurations within the training distribution. OOD trials introduce unseen mouse poses, unfamiliar chargers and cables, and larger trash collections containing unseen objects. The task objectives and scoring criteria are shared across the two settings. \Cref{tab:app_real_task_summary} summarizes the task objectives and OOD conditions.

\begin{table}[t]
    \centering
    \caption{Real-world task objectives and OOD conditions.
    EASY and HARD correspond to ID and OOD, respectively.}
    \label{tab:app_real_task_summary}
    \small
    \setlength{\tabcolsep}{4pt}
    \renewcommand{\arraystretch}{1.15}
    \begin{tabular}{@{}p{0.18\linewidth}p{0.34\linewidth}p{0.40\linewidth}@{}}
        \toprule
        Task & Task objective & OOD condition \\
        \midrule
        Mouse placement
        & Move the mouse onto the target mouse pad.
        & Unseen initial mouse poses. \\
        Charger packing
        & Place the charger body and the entire cable inside the receptacle.
        & Unfamiliar chargers and cables. \\
        Trash disposal
        & Deposit all designated objects into the bin.
        & Six objects on each platform, including unseen objects.
          ID uses four objects on Piper and three on H1. \\
        \bottomrule
    \end{tabular}
\end{table}

\subsection{\teal{Mouse} \graytext{Placement}}
\label{app:mouse_placement}

\paragraph{Task description.}
A mouse and a target mouse pad are placed on the work surface. The robot must move the mouse from its initial tabletop position onto the pad. A representative instruction is ``Place the mouse on the mouse pad.'' The pick-and-place demonstrations involve grasping the rear of the mouse, moving it above the pad, and releasing it within the pad region. In the Piper demonstrations, the right arm performs the manipulation while the left arm initially remains on standby. The task requires locating the mouse and controlling its placement relative to the target pad.

\paragraph{ID and OOD conditions.}
ID trials use mouse poses and initial configurations within the training distribution. OOD trials introduce unseen initial mouse poses. These trials evaluate whether the policy can complete the placement task when the mouse starts in an unfamiliar pose.

\paragraph{Success and completion score.}
A trial succeeds when the mouse ends on the target pad. Placement achieved by pushing also counts as success. Let $m_i \in \{0,1\}$ indicate whether this condition is satisfied in trial $i$. The completion score and success indicator are
\begin{equation}
    c_i^{\mathrm{mouse}} = m_i,
    \qquad
    y_i^{\mathrm{mouse}} = m_i.
    \label{eq:app_mouse_score}
\end{equation}
The task receives a score of 1 for successful placement and 0 otherwise. Its mean completion score equals its success rate divided by 100.

\subsection{\teal{Charger} \graytext{Packing}}
\label{app:charger_packing}

\paragraph{Task description.}
The scene contains a charger body, its cable, and a storage receptacle on the work surface. The robot must place both the charger body and the entire cable inside the receptacle. A representative instruction is ``Put the charging cable into the basket.'' In the demonstrated sequence, the robot grasps the charger assembly, moves it above the receptacle, and places it inside. Completing the task also requires moving any cable segments that remain outside into the receptacle. The task therefore involves manipulating both a rigid body and a flexible cable.

\paragraph{ID and OOD conditions.}
ID trials use chargers, cables, and initial configurations within the training distribution. OOD trials introduce unfamiliar chargers and cables. The containment objective remains the same in both settings. Successful execution requires placing the full assembly inside the receptacle.

\paragraph{Success and completion score.}
Let $b_i \in \{0,1\}$ indicate whether the charger body is inside the receptacle at the end of trial $i$. Let $l_i \in \{0,1\}$ indicate whether the entire cable is inside. We define
\begin{equation}
    c_i^{\mathrm{charger}} = 0.25 b_i + 0.75 l_i,
    \qquad
    y_i^{\mathrm{charger}} = b_i l_i.
    \label{eq:app_charger_score}
\end{equation}
The charger body contributes 0.25 to the completion score, and full cable containment contributes 0.75. Each component receives credit when its containment condition is satisfied. A trial with the body inside and part of the cable outside receives a score of 0.25. Complete success requires both the body and the entire cable to be inside, yielding a score of 1.

\subsection{\teal{Trash} \graytext{Disposal}}
\label{app:trash_disposal}

\paragraph{Task description.}
A collection of designated objects and a trash bin are placed on the work surface. The robot must transfer every designated object into the bin. A representative instruction is ``Put the trash into the bin.'' Objects in the demonstrations include crumpled paper and plastic bottles. The robot grasps an object, moves it above the bin, and opens the gripper to release it. This sequence is repeated until all designated objects have been deposited. The task requires successful execution across multiple object transfers within the same trial.

\paragraph{ID and OOD conditions.}
ID trials use objects and initial configurations within the training distribution. The number of designated objects depends on the platform and setting:
\begin{equation}
    N =
    \begin{cases}
        4, & \text{Piper, ID}, \\
        3, & \text{H1, ID}, \\
        6, & \text{either platform, OOD}.
    \end{cases}
    \label{eq:app_trash_count}
\end{equation}
OOD trials also include unseen objects. This setting evaluates performance with both a larger object collection and changes in object identity.

\paragraph{Success and completion score.}
Let $k_i$ be the number of designated objects deposited into the bin by the end of trial $i$. The completion score and success indicator are
\begin{equation}
    c_i^{\mathrm{trash}} = \frac{k_i}{N},
    \qquad
    y_i^{\mathrm{trash}} =
    \begin{cases}
        1, & k_i = N, \\
        0, & k_i < N.
    \end{cases}
    \label{eq:app_trash_score}
\end{equation}
Each object contributes equally to the completion score. For example, depositing five of six objects in an OOD trial gives a score of $5/6$. The trial is counted as successful only when all six objects have been deposited.

\subsection{\teal{Real-World Metrics} \graytext{and Aggregation}}
\label{app:real_metrics}

For each method, pretraining regime, platform, task, and setting, we report success rate (SR) and mean completion score (CS) over $n = 10$ trials:
\begin{equation}
    \mathrm{SR} = \frac{100}{n} \sum_{i=1}^{n} y_i,
    \qquad
    \mathrm{CS} = \frac{1}{n} \sum_{i=1}^{n} c_i.
    \label{eq:app_real_metrics}
\end{equation}
SR is expressed as a percentage, and CS lies in $[0,1]$. For each platform and setting, the Mean columns average the three tasks with equal weight. Platform-level results across ID and OOD average the six task--setting combinations with equal weight. Since each combination contains 10 trials, these results also equal the corresponding averages over 60 trial outcomes.

SR measures complete task success. CS additionally captures partial completion in charger packing and trash disposal. Mouse placement uses a binary completion score. The same task-specific scoring criteria apply to all methods and pretraining regimes.

\section{\teal{Large-Scale Embodied Pretraining} \graytext{Settings (Single-Backbone Variant)}}
\label{appendix:pretrain-setting}

This appendix details the dataset preprocessing, sample construction, architecture initialization, and 56-GPU distributed optimization used for the single-backbone ($K=1$, 2.32B) AgiBotWorld-Beta pretraining study in \cref{sec:embodied_pretraining}.

\subsection{\teal{Dataset} \graytext{and Preprocessing}}

We pretrain on AgiBotWorld-Beta~\citep{bu2025agibot}, a large-scale dataset of real-world robot manipulation demonstrations collected using mobile dual-arm robots. The dataset covers diverse objects and environments, with tasks ranging from fine-grained manipulation and tool use to long-horizon activities and multi-robot collaboration. The official collection spans 217 tasks and 87 skills across more than 100 scenarios in five application domains, providing a broad foundation for learning manipulation policies.

For this pretraining, after required-modality checks and filtering, the training corpus contains 177 tasks, 160,010 trajectories, and 285,728,736 synchronized time steps. At a sampling rate of 30 Hz, this corresponds to approximately 2,645.64 hours of trajectory data. Each time step provides head, left-wrist, and right-wrist views together with the corresponding robot state and camera parameters. Duration is measured on the synchronized trajectory timeline; the three views represent simultaneous observations.

Admission requires readable RGB from all three views, valid joint and gripper states, corresponding depth, and camera calibration supported by the processing pipeline. Trajectories with missing required modalities, invalid state values, or unsupported calibration are excluded, and identical duplicate annotations are collapsed. Episodes with optional end-effector-field issues may be retained when their joint-space training inputs remain valid, with the relevant validity information recorded. All admitted trajectories are used for pretraining without a separate validation split.

The original data provide sensor depth for the head camera but lack depth observations for both wrist cameras. Moreover, the available head-camera depth maps contain substantial holes and noisy measurements, including missing regions around the robot arms, leaving manipulation-relevant areas without valid geometric supervision. To address both the missing wrist-camera depth and the incomplete head-camera depth, we generate depth pseudo-labels for all three views offline over complete RGB videos using \emph{DA3METRIC-LARGE} from Depth Anything 3~\citep{lin2025depth}. This provides denser depth targets through a uniform estimation procedure across views; these targets constitute estimated geometric supervision rather than sensor-measured ground truth. Predictions are aligned with robot states using source frame indices and timestamps and stored together with camera parameters. RGB and depth undergo consistent rectification. Original camera extrinsics are inverted as complete rigid transforms to obtain robot-reference-to-camera transformations, while intrinsics are adjusted to the resized image coordinates. Depth is encoded losslessly with 16-bit values in millimeters, converted to meters during loading, and accompanied by validity masks.

The processed corpus is stored in LeRobot v3.0 format. RGB observations are stored as videos; robot states, action labels, and camera parameters are stored in Parquet, and PNG16 depth bytes are embedded in the same dataset format. Shared episode identifiers and frame indices provide aligned access across modalities. Joint and gripper values are standardized using fixed per-dimension means and standard deviations computed before training.

The depth targets are model-generated pseudo-labels, and the three-dimensional targets are derived from those predictions and dataset-provided calibration. Their accuracy therefore depends on both depth estimation and camera calibration.

\subsection{\teal{Training Samples} \graytext{and Prediction Targets}}

Samples are constructed as consecutive temporal windows within each trajectory. A trajectory of length $T$ provides $T-32$ valid windows with a stride of one. Each window spans the current state and the following 32 time steps, without crossing an episode boundary. The corpus yields 280,608,416 valid training windows, which are globally shuffled each epoch and distributed across training processes.

Model inputs comprise the three current RGB views, the current 16-dimensional proprioceptive state, and task-language features. Images are stored at $480 \times 640$ and transformed to an actual input resolution of $224 \times 288$ by VGGT-Omega~\citep{wang2026vggtomega} balanced preprocessing; dimensions are stated as height $\times$ width. Language features are extracted offline with the pretrained UMT5-XXL~\citep{chung2023unimax} encoder, using at most 128 tokens per instruction and 4096 features per token. Cached features are read during training, and the text encoder remains fixed.

Action labels are constructed from the next observed robot state. Denoting the current state by $s_t$, the action target is $a_t = s_{t+1}$, and the complete target sequence spans $s_{t+1}$ through $s_{t+32}$, with shape $32 \times 16$. These targets represent absolute future joint positions and gripper values before per-dimension normalization. Only the current state is supplied as proprioceptive input; subsequent states are used to construct supervision.
The action objective uses continuous flow matching. Normalized action sequences are mixed with Gaussian noise at randomly sampled noise levels, and the model predicts the flow from actions to noise using a timestep-weighted mean-squared error. The loss is aggregated over action dimensions and valid time steps within the 32-step horizon. The scheduler uses 1000 nominal training timesteps, with both training and inference shifts set to 5.0.

Geometric supervision is applied at the final time step $t+32$, approximately 1.067 seconds into the future. It consists of three-view depth and camera parameters, together with three-dimensional points obtained by unprojection. The model predicts this future geometry from current observations. Future RGB and target geometry are not conditioning inputs.
The geometry objective uses the official VGGT-Omega~\citep{wang2026vggtomega} loss implementation to supervise future camera parameters, depth, and three-dimensional points. Camera, depth, and point losses each have an internal weight of 5.0, with an overall geometry weight of 0.01. The depth objective includes confidence modeling and a separate regression term. Point supervision is computed by unprojecting depth with camera parameters, coupling the two predictions without an independent point-cloud prediction head.

Before loss computation, geometry targets are expressed in the first camera's reference frame and normalized by a per-sample scale derived from the mean distance of valid three-dimensional points. Point unprojection uses target intrinsics during the first 60\% of training and predicted focal lengths with target principal points thereafter. Invalid depth pixels are masked, and geometric losses are computed in FP32. These operations preserve the relationship between cameras, depth, and points within the joint learning objective.

\subsection{\teal{Architecture} \graytext{and Initialization}}

The model consists of a single language-conditioned VGGT-Omega~\citep{wang2026vggtomega} visual stream and a 30-layer ActionDiT action stream. The visual stream processes the three current images and integrates language information within cross-view attention, producing visual features for action prediction and features for future-geometry prediction. The action stream combines visual features, language, current proprioception, and a noisy action sequence to learn action generation.

The streams are connected through layer-wise attention bridges. Twenty-four VGGT layers are paired with ActionDiT layers  (from 3 to 26 using zero-based indices), and each bridge has independent projection parameters. ActionDiT uses a hidden dimension of 1024, an FFN dimension of 4096, and 24 attention heads with 128 dimensions per head. VGGT uses a feature width of 1024 and 16 attention heads.

We initialize the VGGT aggregator and camera/depth heads with pretrained weights, while ActionDiT is randomly initialized. The attention bridges, proprioceptive encoder, and language projection are jointly optimized during training. The VGGT backbone remains fully trainable, receiving gradients from both action and geometry objectives.

\subsection{\teal{Optimization} \graytext{and Distributed Training}}

Training uses 7 servers, each equipped with eight PRO6000 GPUs with 96 GB of memory, for a total of 56 GPUs. Distributed execution uses Accelerate and DeepSpeed ZeRO Stage 1~\citep{rajbhandari2020zero} with BF16 mixed precision and no parameter or optimizer CPU offloading. A per-GPU micro-batch size of eight and two gradient-accumulation steps yield an effective global batch size of 896 for regular optimizer updates.

Optimization uses AdamW~\citep{loshchilov2017decoupled} with $\beta_1=0.9, \beta_2=0.95$, weight decay of 0.01, and gradient-norm clipping at 1.0. Both the action-related and VGGT parameter groups use a peak learning rate of $1 \times 10^{-4}$. A linear warmup over the first 2000 optimizer updates is followed by cosine decay to $1 \times 10^{-6}$.

The pretraining budget is two epochs. At the current dataset size and batch configuration, each epoch contains 313,180 optimizer updates, giving a total of 626,360 updates. A step denotes an optimizer update after gradient accumulation. Distributed sampling introduces a small amount of epoch-tail padding, and the final update of an epoch may contain fewer than two micro-batches. The full budget defines both the learning-rate schedule and the geometry-loss schedule.

Each GPU process uses 12 data-loading workers with prefetching and persistent workers. The reader combines Parquet column projection with bounded caches, decoding current RGB on demand and loading the required consecutive robot labels and geometry. Full training states are saved every 2000 updates, and standalone model weights are exported every 10,000 updates. Resumption restores model and optimizer states, the learning-rate schedule, sampler position, and random-number-generator states to preserve training continuity.

\begin{figure}[t]
    \centering
    \includegraphics[width=\linewidth]{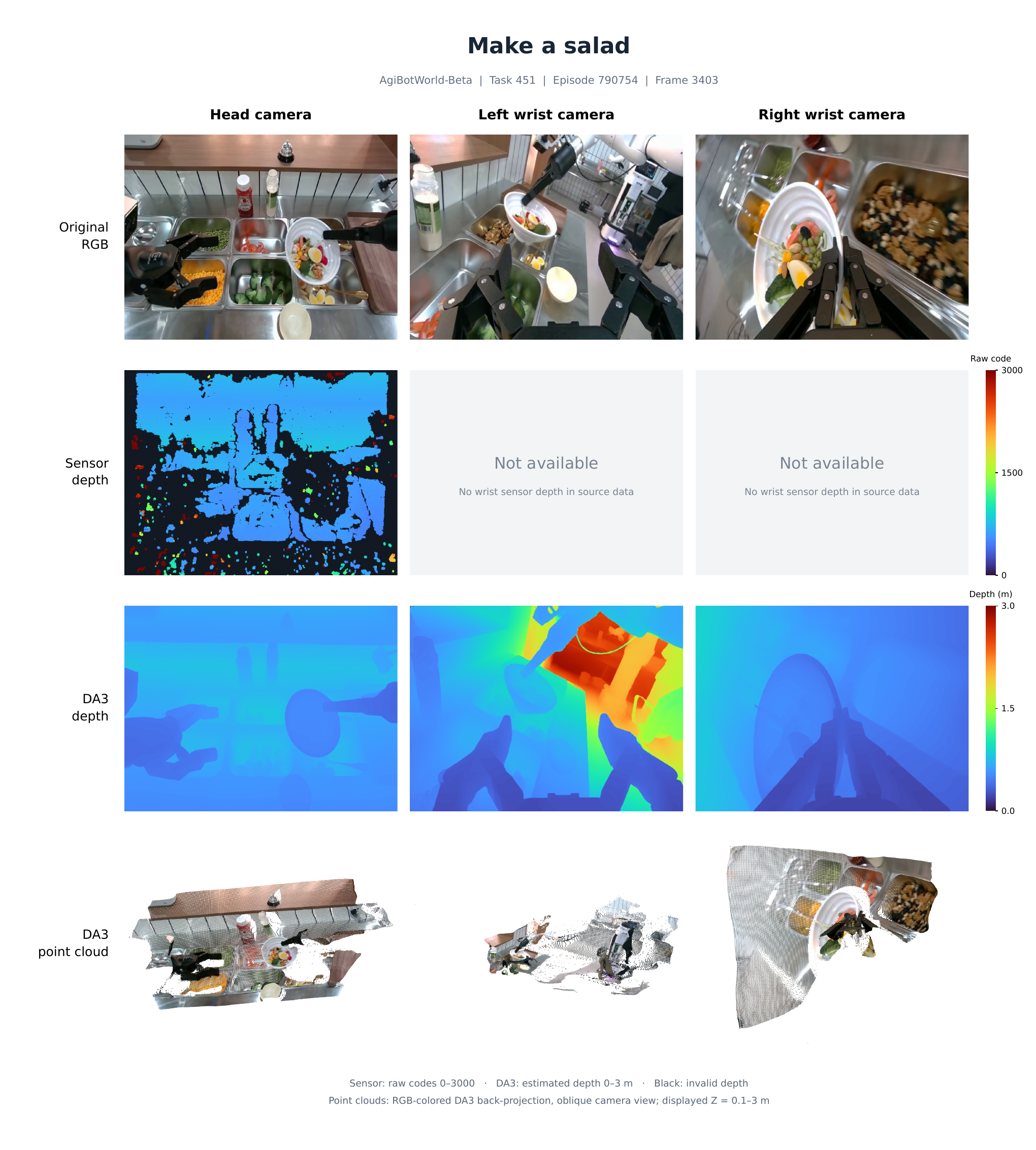}
    \caption{\textbf{Qualitative visualization of the \textit{Make a salad} task in AgiBotWorld-Beta.} Columns correspond to the head, left-wrist, and right-wrist cameras. From top to bottom, rows show the original RGB images, available sensor depth, DA3-predicted depth, and RGB-colored point clouds back-projected from the predictions. Wrist sensor depth is unavailable. Sensor depth is displayed in raw codes, while DA3 depth is displayed in estimated metres.}
    \label{fig:task-451}
\end{figure}

\begin{figure}[t]
    \centering
    \includegraphics[width=\linewidth]{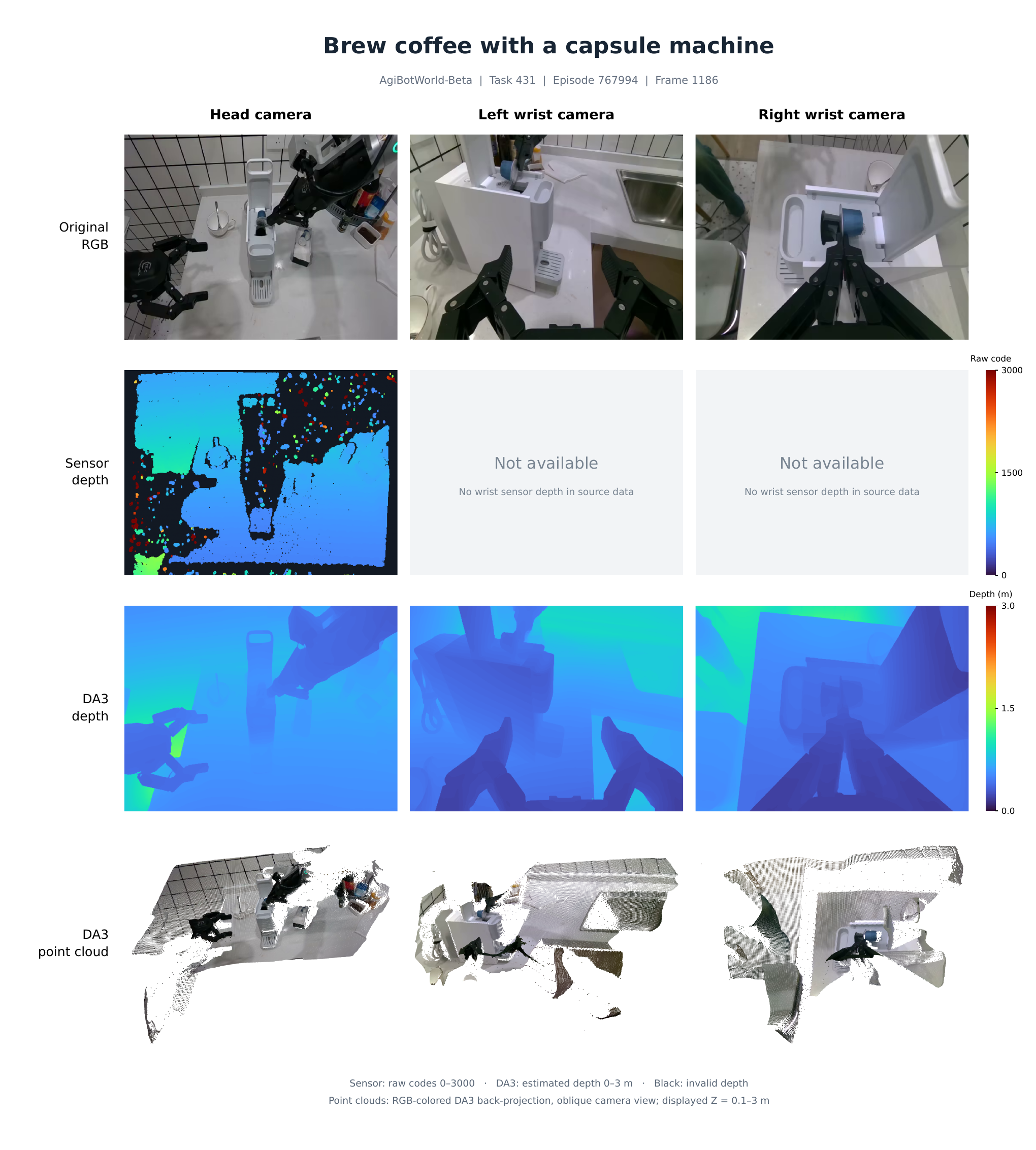}
    \caption{\textbf{Qualitative visualization of the \textit{Brew coffee with a capsule machine} task in AgiBotWorld-Beta.} Columns correspond to the head, left-wrist, and right-wrist cameras. Rows show the original RGB images, available sensor depth, DA3-predicted depth, and the corresponding RGB-colored point clouds. Wrist sensor depth is unavailable. Sensor depth and DA3 depth use raw-code and estimated-metre scales, respectively.}
    \label{fig:task-431}
\end{figure}

\begin{figure}[t]
    \centering
    \includegraphics[width=\linewidth]{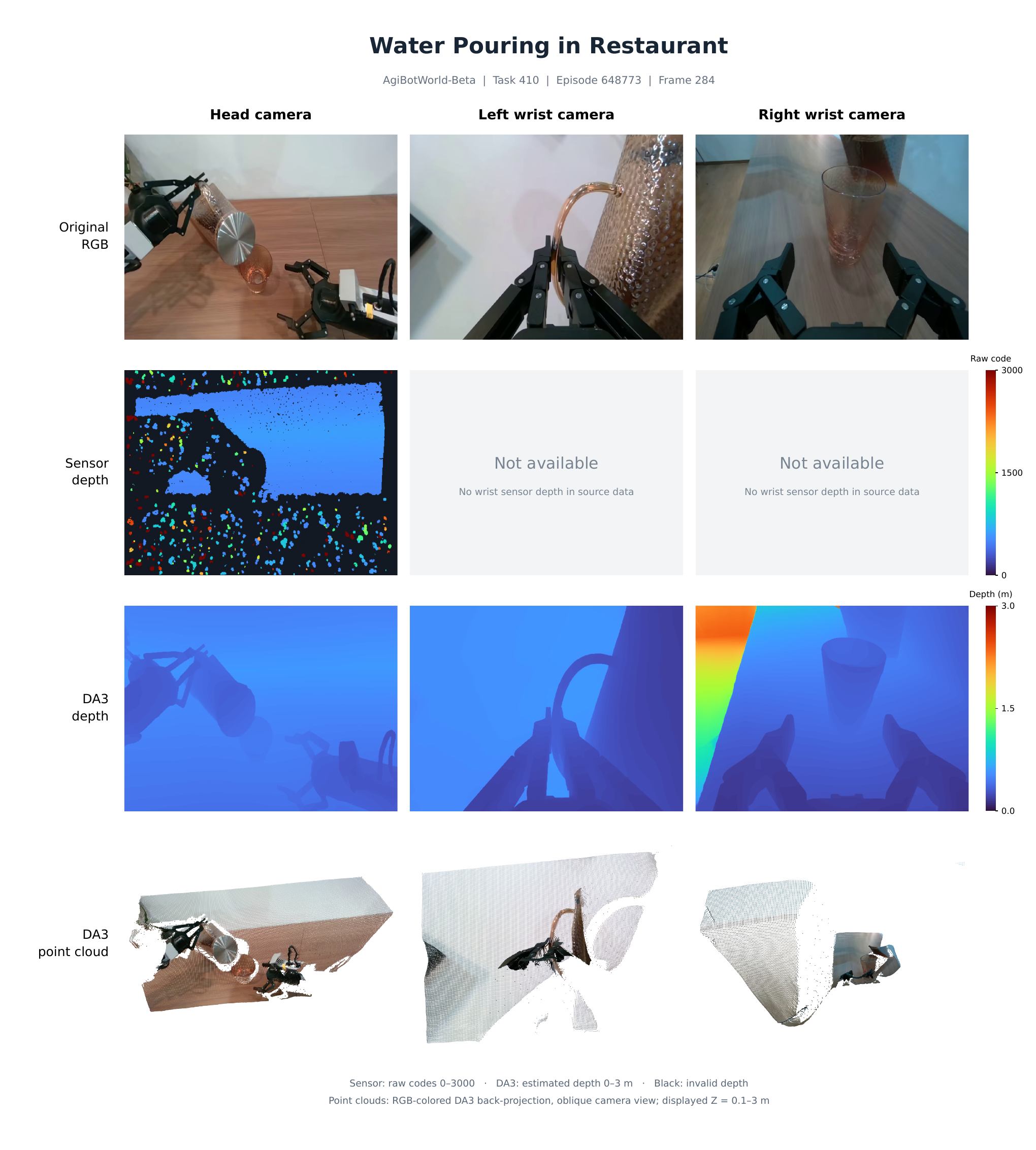}
    \caption{\textbf{Qualitative visualization of the \textit{Water Pouring in Restaurant} task in AgiBotWorld-Beta.} From left to right, the views are captured by the head, left-wrist, and right-wrist cameras. From top to bottom, the rows present RGB images, available sensor depth, DA3-predicted depth, and RGB-colored point clouds reconstructed from the predictions. Wrist sensor depth is unavailable; sensor and predicted depth are displayed in raw codes and estimated metres, respectively.}
    \label{fig:task-410}
\end{figure}

\begin{figure}[t]
    \centering
    \includegraphics[width=\linewidth]{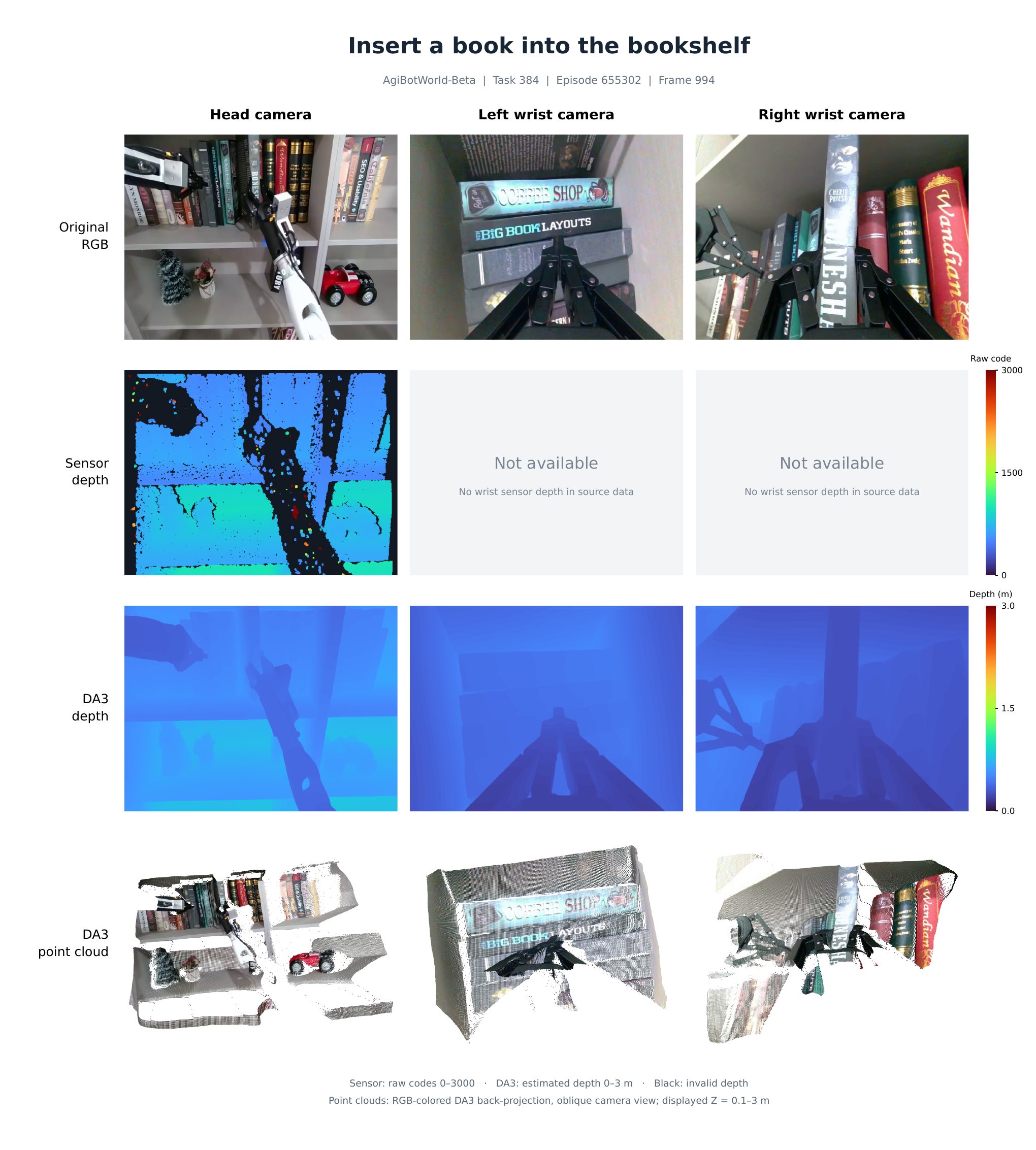}
    \caption{\textbf{Qualitative visualization of the \textit{Insert a book into the bookshelf} task in AgiBotWorld-Beta.} Columns show the head, left-wrist, and right-wrist views, and rows show RGB images, available sensor depth, DA3-predicted depth, and the corresponding RGB-colored point clouds. Wrist sensor depth is unavailable. Sensor depth is visualized in raw codes, whereas DA3 depth is visualized in estimated metres.}
    \label{fig:task-384}
\end{figure}

\clearpage
\section{\teal{Qualitative} \graytext{Demonstrations}}
\label{sec:qualitative_demonstration}

\begin{figure}[p]
    \centering
    \includegraphics[width=\textwidth,height=0.82\textheight,keepaspectratio]{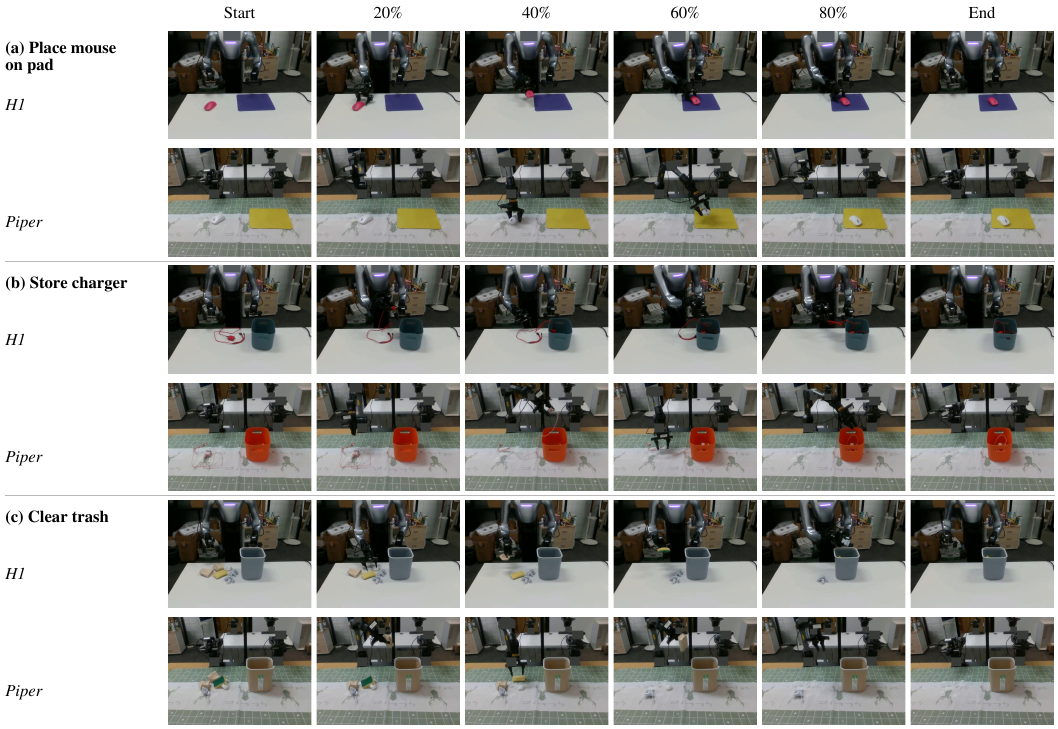}
    \caption{\textbf{Real-world demonstrations on H1 and Piper.}
    The three tasks are (a) placing a mouse on a mouse pad,
    (b) storing a charger in a container, and (c) clearing trash from a table.
    Each task occupies two consecutive rows, with H1 above Piper.
    Six external-camera observations show the progression of each execution.
        Full execution videos are available on the project page.}
    \label{fig:qualitative_real_world}
\end{figure}

\begin{figure}[p]
    \centering
    \includegraphics[width=\textwidth,height=0.82\textheight,keepaspectratio]{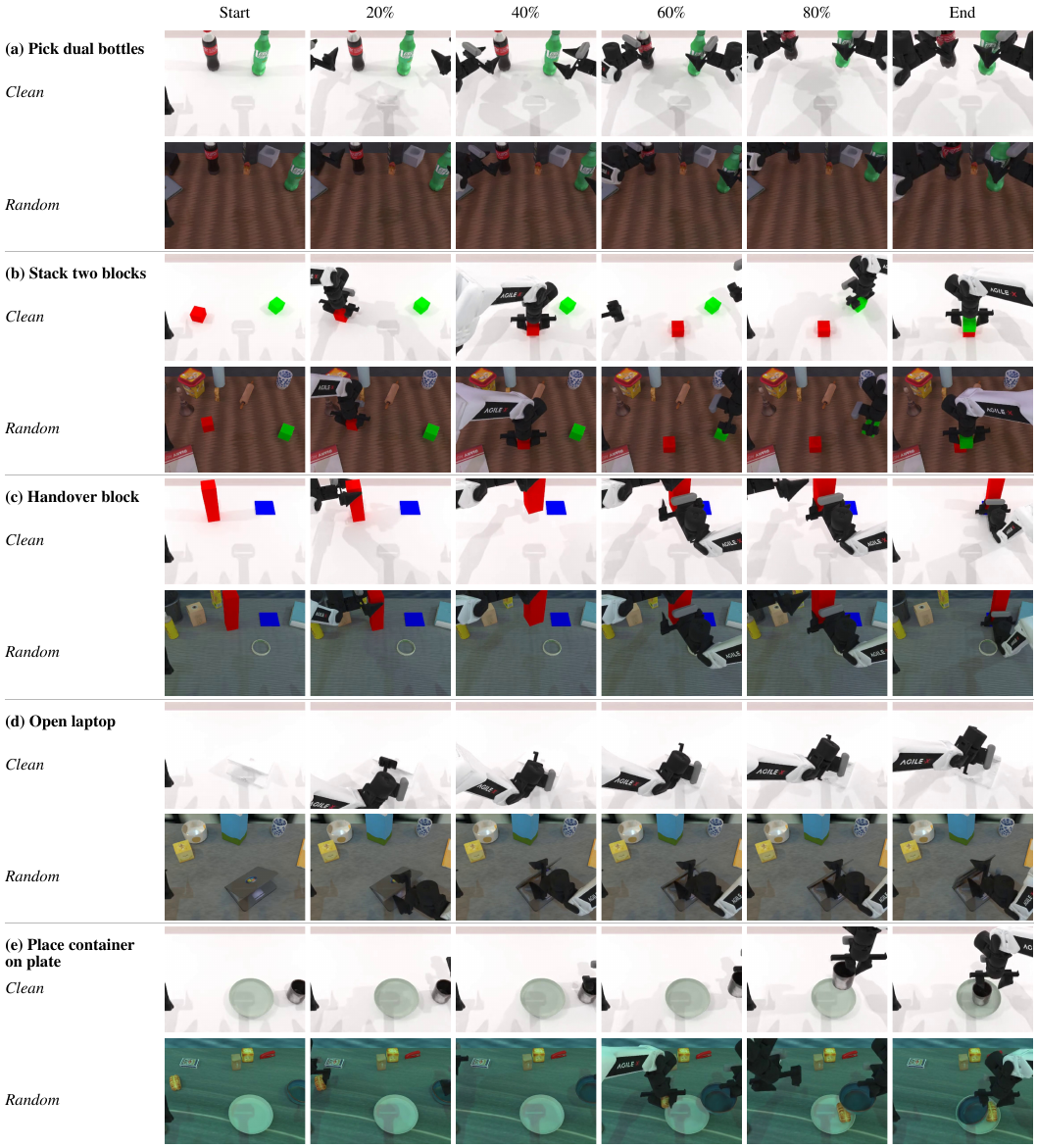}
    \caption{\textbf{Successful RoboTwin executions under Clean and Random conditions.}
    The selected tasks are (a) picking dual bottles, (b) stacking two blocks,
    (c) handing over a block, (d) opening a laptop, and
    (e) placing a container on a plate. Each task occupies two rows,
    with Clean above Random. Columns show six head-camera observations
    sampled uniformly over each recorded episode. All illustrated episodes
    satisfy the evaluator's task-success criterion. Additional multi-view task rollouts are available on the project page.}
    \label{fig:qualitative_robotwin}
\end{figure}

\begin{figure}[p]
    \centering
    \includegraphics[width=\textwidth,height=0.82\textheight,keepaspectratio]{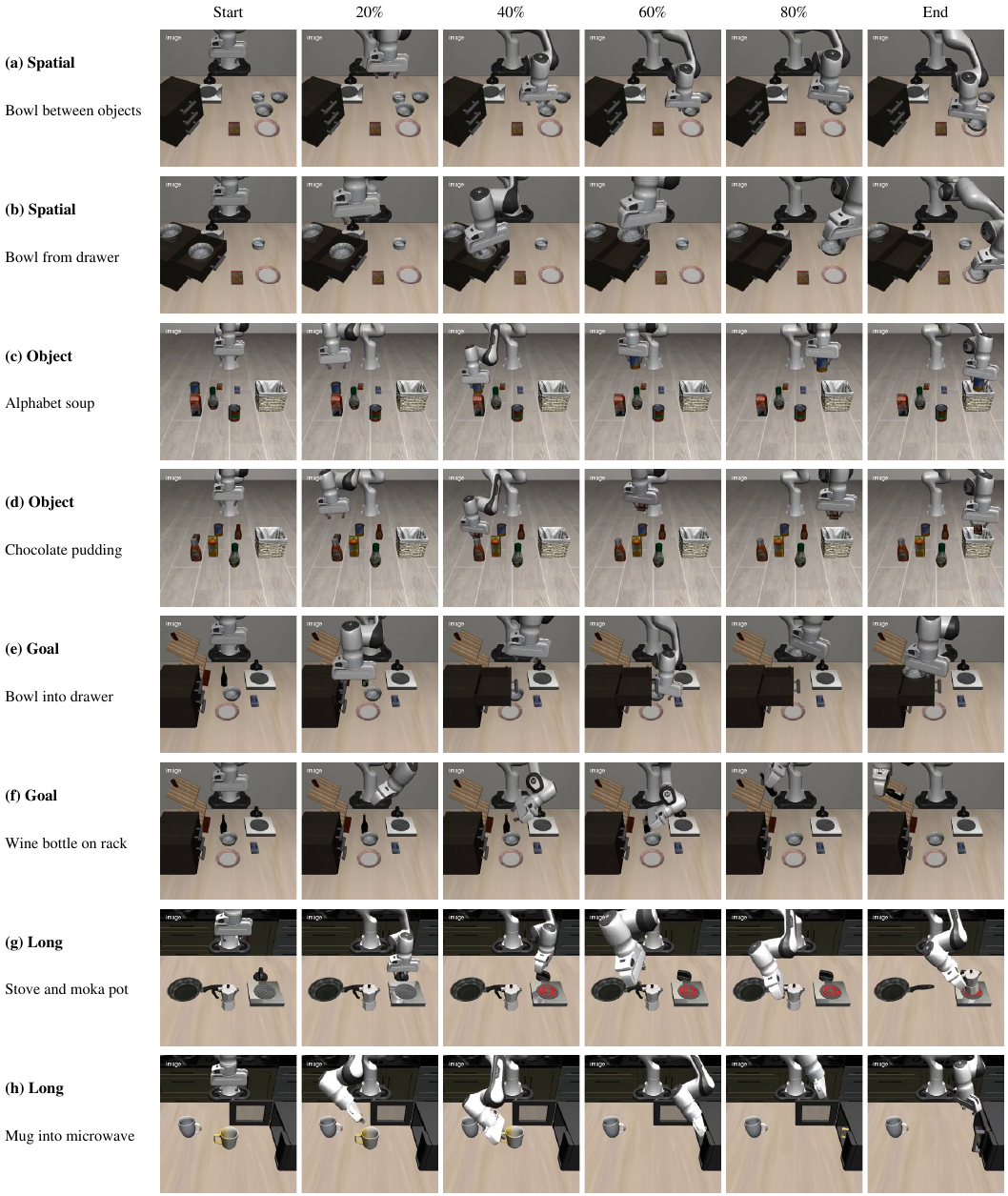}
    \caption{\textbf{Successful executions across the four LIBERO suites.}
    Rows (a)--(b), (c)--(d), (e)--(f), and (g)--(h) show two selected tasks
    from Spatial, Object, Goal, and Long, respectively. Each row contains
    six external-camera observations sampled uniformly over one episode,
    illustrating the progression from the initial scene to task completion.
    Full rollout videos are available on the project page.}
    \label{fig:qualitative_libero}
\end{figure}

\begin{figure}[p]
    \centering
    \includegraphics[width=\textwidth,height=0.82\textheight,keepaspectratio]{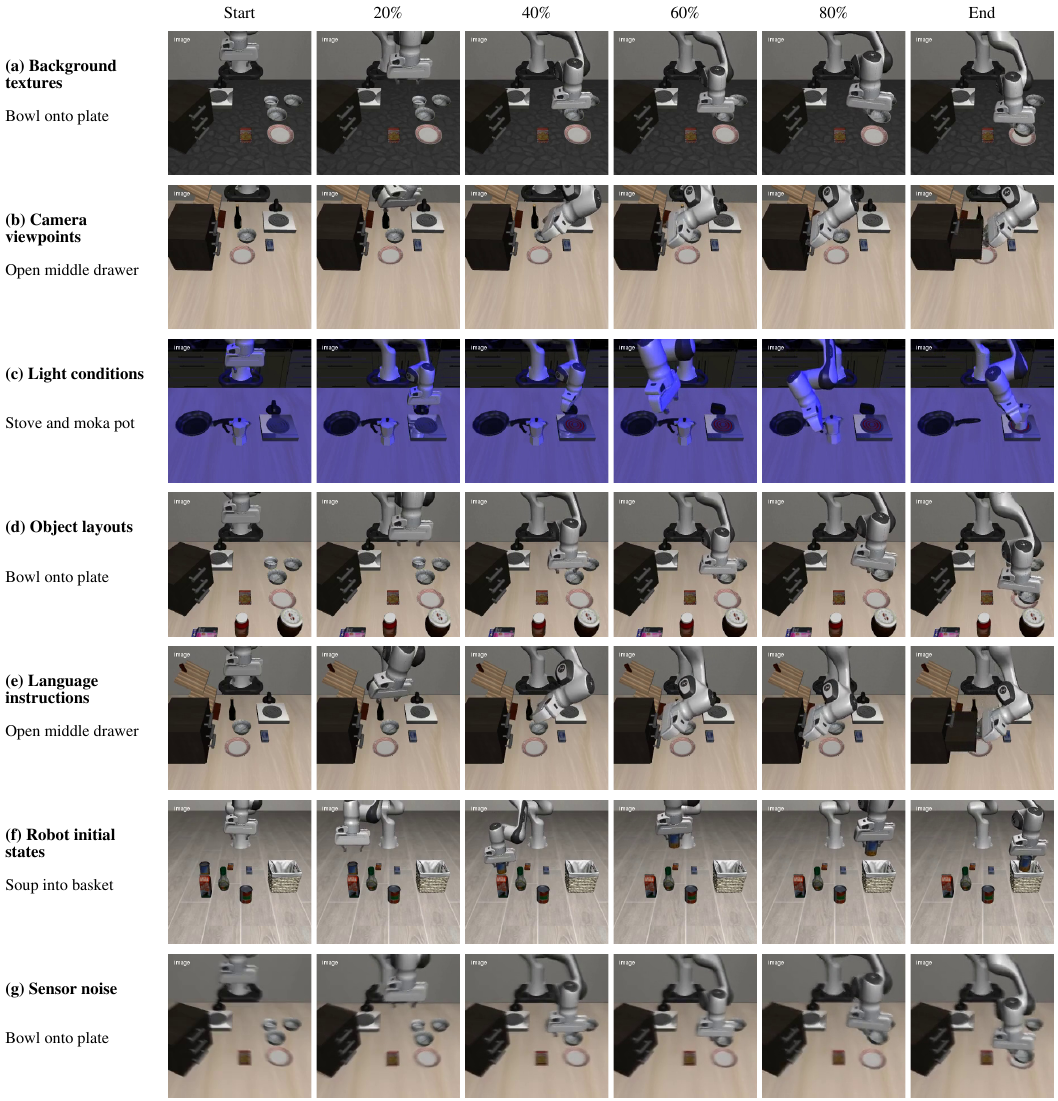}
    \caption{\textbf{Successful executions across all seven LIBERO-Plus perturbation dimensions.}
    Rows show (a) background textures, (b) camera viewpoints,
    (c) light conditions, (d) object layouts, (e) language instructions,
    (f) robot initial states, and (g) sensor noise. Each row contains
    six external-camera observations from one successful episode using
    the LIBERO-trained checkpoint without additional fine-tuning.
    Corresponding rollout videos are available on the project page.}
    \label{fig:qualitative_libero_plus}
\end{figure}

\begin{figure}[p]
    \centering
    \includegraphics[width=\textwidth,height=0.82\textheight,keepaspectratio]{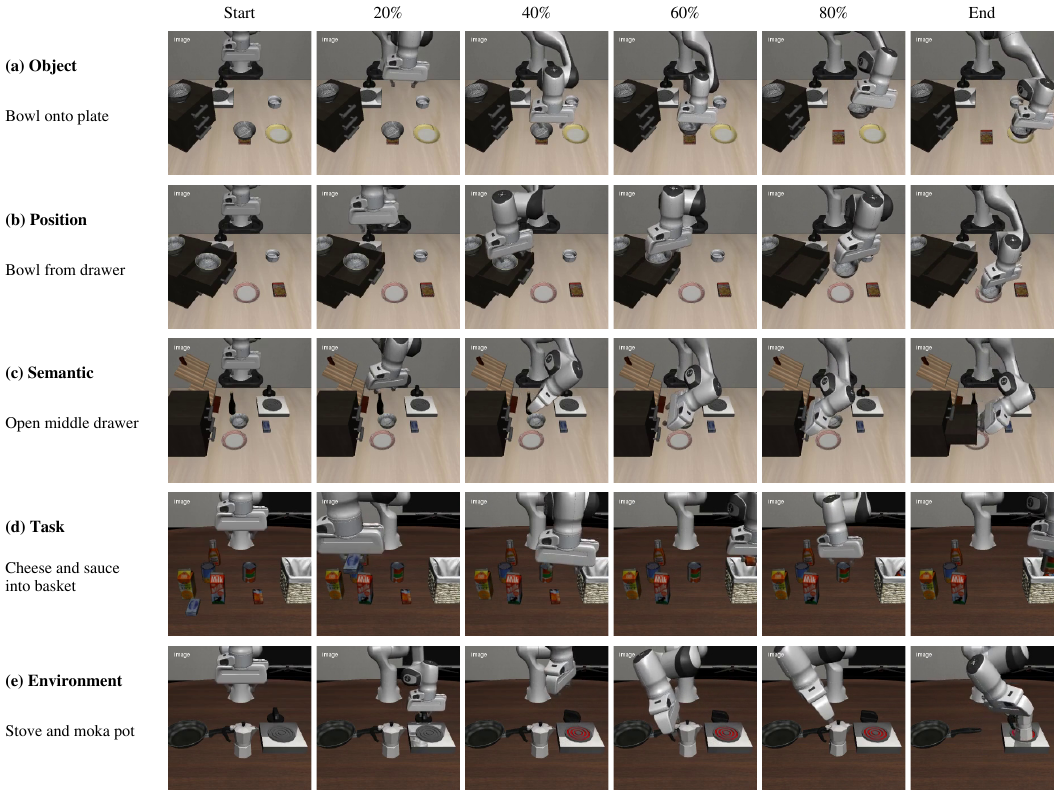}
    \caption{\textbf{Successful executions across all five LIBERO-PRO shift dimensions.}
    Rows show (a) Object, (b) Position, (c) Semantic, (d) Task, and
    (e) Environment. Each row contains six external-camera observations
    from one successful episode using the LIBERO-trained checkpoint
    without additional fine-tuning. Progress is normalized independently
    within each row. Corresponding multi-view rollout videos are available on the project page.}
    \label{fig:qualitative_libero_pro}
\end{figure}

We present qualitative demonstrations on two real-world platforms, H1 and
Piper, and four simulation benchmarks: RoboTwin, LIBERO, LIBERO-Plus, and
LIBERO-PRO. \Cref{fig:qualitative_real_world,fig:qualitative_robotwin,fig:qualitative_libero,fig:qualitative_libero_plus,fig:qualitative_libero_pro}
visualize selected executions as sequences of six observations, sampled
uniformly from the start to the end of each video. Progress is normalized
independently for each row; matching columns therefore do not imply equal
elapsed time across executions. These examples illustrate task execution
and complement the quantitative evaluation.

\paragraph{Real-world manipulation.}
\Cref{fig:qualitative_real_world} shows H1 and Piper performing three
tasks: placing a mouse on a mouse pad, storing a charger in a container,
and clearing trash from a table. For each task, paired rows show the
execution on the two platforms, from the initial scene through object
manipulation to the final configuration.

\paragraph{Simulation benchmarks.}
\Cref{fig:qualitative_robotwin} presents five RoboTwin tasks under
both Clean and Random conditions, covering dual-object grasping, stacking,
handover, articulated-object manipulation, and object placement.
\Cref{fig:qualitative_libero} shows eight LIBERO tasks, with two
examples from each of the Spatial, Object, Goal, and Long suites.
\Cref{fig:qualitative_libero_plus,fig:qualitative_libero_pro}
provide one successful execution for each of the seven LIBERO-Plus
perturbation dimensions and five LIBERO-PRO shift dimensions, respectively.
Both use the LIBERO-trained checkpoint without additional fine-tuning.


\clearpage

\clearpage

\end{document}